\documentclass[10pt,twocolumn,letterpaper]{article}

\usepackage[pagenumbers]{cvpr}

\usepackage{color, colortbl}
\usepackage{pgf} %
\definecolor{Gray}{gray}{0.9}
\usepackage{tikz} %
\usepackage{pgffor} %

\usepackage{graphicx} %
\usepackage{subcaption} %

\usepackage{tikz, pgffor} %

\usepackage{pgfplotstable}
\usepackage{graphicx} %
\usepackage{float} %
\usepackage{booktabs} %

\usepackage{amsmath}

\pgfplotstableset{
    color cells/.style={
        col sep=comma,
        string type,
        postproc cell content/.code={%
                \pgfkeysalso{@cell content=\rule{0cm}{2.4ex}\cellcolor{black!##1}\pgfmathtruncatemacro\number{##1}\ifnum\number>50\color{white}\fi##1}%
                },
        columns/x/.style={
            column name={},
            postproc cell content/.code={}
        }
    }
}

\definecolor{cvprblue}{rgb}{0.21,0.49,0.74}
\usepackage[pagebackref,breaklinks,colorlinks,allcolors=cvprblue]{hyperref}

\usepackage{tablefootnote}
\usepackage{multirow}
\usepackage{adjustbox}

\DeclareMathOperator*{\argmin}{arg\,min}

\usepackage[textsize=tiny, textwidth=2cm]{todonotes}

\def\paperID{17800}
\def\confName{CVPR}
\def\confYear{2025}

\title{\benchmarkname: Benchmarking Fine-Tuning Robustness across Multi-Modal Shifts in Visual Question Answering}

\author{Chengyue Huang\textsuperscript{*} \quad Brisa Maneechotesuwan\textsuperscript{*} \quad Shivang Chopra \quad Zsolt Kira\\
Georgia Institute of Technology\\
{\tt\small \{chuang475,bmaneech3,shivangchopra11,zkira\}@gatech.edu}
}

\def\benchmarkname{FRAMES-VQA}

\begin{document}
\maketitle 
\def\thefootnote{*}\footnotetext{Equal contribution.}

\begin{abstract}
Visual question answering (VQA) systems face significant challenges when adapting to real-world data shifts, especially in multi-modal contexts. 
While robust fine-tuning strategies are essential for maintaining performance across in-distribution (ID) and out-of-distribution (OOD) scenarios, current evaluation settings are primarily unimodal or  particular to some types of OOD, 
offering limited insight into the complexities of multi-modal contexts.
In this work, we propose a new benchmark \benchmarkname~(\textbf{F}ine-Tuning \textbf{R}obustness \textbf{A}cross \textbf{M}ulti-Modal \textbf{S}hifts in VQA) %
for evaluating robust fine-tuning for VQA tasks. We utilize ten existing VQA benchmarks, including VQAv2, IV-VQA, VQA-CP, OK-VQA and others, 
and categorize them into ID, near and far OOD datasets covering uni-modal, multi-modal and adversarial distribution shifts. We first conduct a comprehensive comparison of existing robust fine-tuning methods. We then quantify the distribution shifts by calculating the Mahalanobis distance using uni-modal and multi-modal embeddings extracted from various models. 
Further, we perform an extensive analysis to explore the interactions between uni- and multi-modal shifts 
as well as modality importance for ID and OOD samples. 
These analyses offer valuable guidance on developing more robust fine-tuning methods to handle multi-modal distribution shifts. The code is available at \url{https://github.com/chengyuehuang511/FRAMES-VQA}.
\end{abstract}

\section{Introduction}
\label{sec:intro}

Robust fine-tuning methods aim to adapt pre-trained models to downstream tasks while retaining resilience to distribution shifts~\citep{radford_learning_2021,wortsman_robust_2022}. Distribution shifts in the image modality are widely explored, with various datasets designed to evaluate a model's generalization across diverse visual conditions. For example, DomainNet~\citep{peng2019moment} spans multiple visual domains including real images, sketches, paintings, and clipart, challenging models to generalize across different styles and representations. Similarly, various ImageNet variants~\citep{deng2009imagenet, recht2019imagenet, hendrycks2021natural, wang2019learning} introduce shifts through image variations, adversarial examples, rendering transformations, and changes in texture or background. Collectively, these datasets provide a comprehensive framework for assessing how well models withstand visual distribution changes.

While robust fine-tuning algorithms are widely examined under distribution shifts in a single modality (images), few studies have explored robust fine-tuning for VQA tasks, where distribution shifts are multi-modal and models must adapt to variations across both visual and textual inputs. Apart from visual shift~\citep{agarwal_towards_2020}, there are question shifts~\citep{shah_cycle-consistency_2019, gokhale_vqa-lol_2020} involving variations in phrasing, structure, or vocabulary, as well as answer shifts~\citep{agrawal_dont_2018} with changes in answer distributions such as frequency and formatting. Beyond uni-modal shift, these variations may occur simultaneously across visual, question, and answer inputs~\citep{unni_vqa-gen_2023, dancette_beyond_2021, si_language_2022, sheng_human-adversarial_2021, li_adversarial_2021}, posing an even greater challenge as models must generalize across complex, combined shifts.

Therefore, we build upon our preliminary exploration~\citep{huang2025directionalgradientprojectionrobust} and propose a benchmark \benchmarkname~(\textbf{F}ine-Tuning \textbf{R}obustness \textbf{A}cross \textbf{M}ulti-Modal \textbf{S}hifts in VQA)~to systematically evaluate the robustness of fine-tuning in VQA task. We leverage ten existing VQA datasets and categorize distribution shifts into uni-modal and multi-modal types, quantified by Mahalanobis distance across various backbones to capture both near and far OOD scenarios. We conduct a comprehensive comparison of the existing robust fine-tuning baselines on ID and OOD performance using the benchmark. Furthermore, we analyze shift scores and modality importance across fine-tuning methods. To summarize, our contributions are:

\begin{itemize}
    \item We propose~\benchmarkname~for evaluating robust fine-tuning in VQA, including ten VQA datasets categorized by uni-modal (e.g., image, question) and multi-modal shifts. We quantify dataset shifts under different modalities using Mahalanobis distance and embeddings from different backbones. 

    \item We perform an in-depth comparison of robust fine-tuning methods using the benchmark. We find that FTP~\citep{tian_fast_2023} has the best far OOD performance while SPD~\citep{tian_rethinking_2024} outperforms others on ID, near and average OOD.

    \item We further provide insightful analyses on the shift scores and modality importance of each baseline across ID and OOD samples. Key observations include: (i) Fine-tuning amplifies question-joint shift correlation, indicating strong influence of question shifts towards multimodal representations; (ii) more robust fine-tuning methods exhibit low correlation between uni-modal and multimodal shifts; (iii) question-to-image attention rises for OOD samples, implying potential shortcuts; and (iv) robust methods emphasize intra- over inter-modality attention, underscoring intra-modality’s role in robustness. These findings offer ways to improve fine-tuning robustness under multi-modal shifts.
\end{itemize}

\section{Related Work}

\vspace{2.5pt} \noindent \textbf{Distribution Shift and OOD Robustness in VQA.} 
DomainNet~\citep{peng2019moment} and ImageNet~\citep{deng2009imagenet} along with its four variants~\citep{recht2019imagenet,hendrycks2021natural,hendrycks2021many,wang2019learning} are commonly used to assess model robustness under distribution shifts. Prior work~\citep{radford_learning_2021,wortsman_robust_2022} show that while vanilla fine-tuning enhances ID performance, it can degrade results on OOD datasets compared to the pre-trained model. Beyond traditional image classification with distribution shifts in the image modality, VQA datasets such as VQAv2~\citep{goyal_making_2017} and its variants~\citep{agarwal_towards_2020,shah_cycle-consistency_2019,gokhale_vqa-lol_2020,agrawal_dont_2018,unni_vqa-gen_2023,dancette_beyond_2021,si_language_2022,sheng_human-adversarial_2021,li_adversarial_2021} introduce both uni-modal (image, question, answer) and multi-modal shifts, posing a greater challenge for models compared to shifts in image-only classification tasks.
Various frameworks have been proposed to assess robustness in VQA. \citep{agrawal_reassessing_2023} explores cross-dataset evaluations across four VQA datasets, while \citep{ma_robust_2024,li_closer_2021} expand this scope by incorporating VQAv2 variants and distinguishing different distribution shift types.
\citep{zhang2021domainrobustvqadiversedatasets} quantifies uni-modal shifts for image and question modalities.
Building on these studies, we introduce finer categorization and distinctions between near and far OOD distributions, along with quantifying both uni- and multi-modal distribution distances. Crucially, while prior work has focused on testing different discriminative models~\citep{ma_robust_2024} or adaptation methods~\citep{chen2023benchmarkingrobustnessadaptationmethods}, our study focuses on comparing robust fine-tuning algorithms within the same backbone and extends to generative models.

\vspace{2.5pt} \noindent \textbf
{Robust Fine-Tuning of Foundation Models.} Robust fine-tuning methods aim to adapt foundation models to new tasks while retaining their pre-trained robustness. LP-FT~\citep{kumar_fine-tuning_2022} introduces a two-step approach of linear probing followed by full fine-tuning to mitigate feature distortion. WiSE-FT~\citep{wortsman_robust_2022} blends pre-trained and fine-tuned weights through interpolation, balancing the strengths of both embedding spaces. L2-SP~\citep{li_explicit_2018} and MARS-SP~\citep{gouk_distance-based_2021} add penalties on the deviation between fine-tuned and pre-trained weights, exploring different norm types. More recent methods, such as TPGM~\citep{tian_trainable_2023}, frame regularization as a constraint through bi-level optimization, learning tailored constraints for each layer. FTP~\citep{tian_fast_2023} enhances TPGM's efficiency by leveraging prior training steps, while SPD~\citep{tian_rethinking_2024} selectively regularizes layers with consistent loss reduction, projecting corresponding layers within the constraint.

\section{\benchmarkname:~Fine-Tuning Robustness across Multi-Modal Shifts in VQA}
\label{sec:robust_ft_exp}

\begin{figure*}[!h]
     \centering
     \includegraphics[width=1\textwidth]{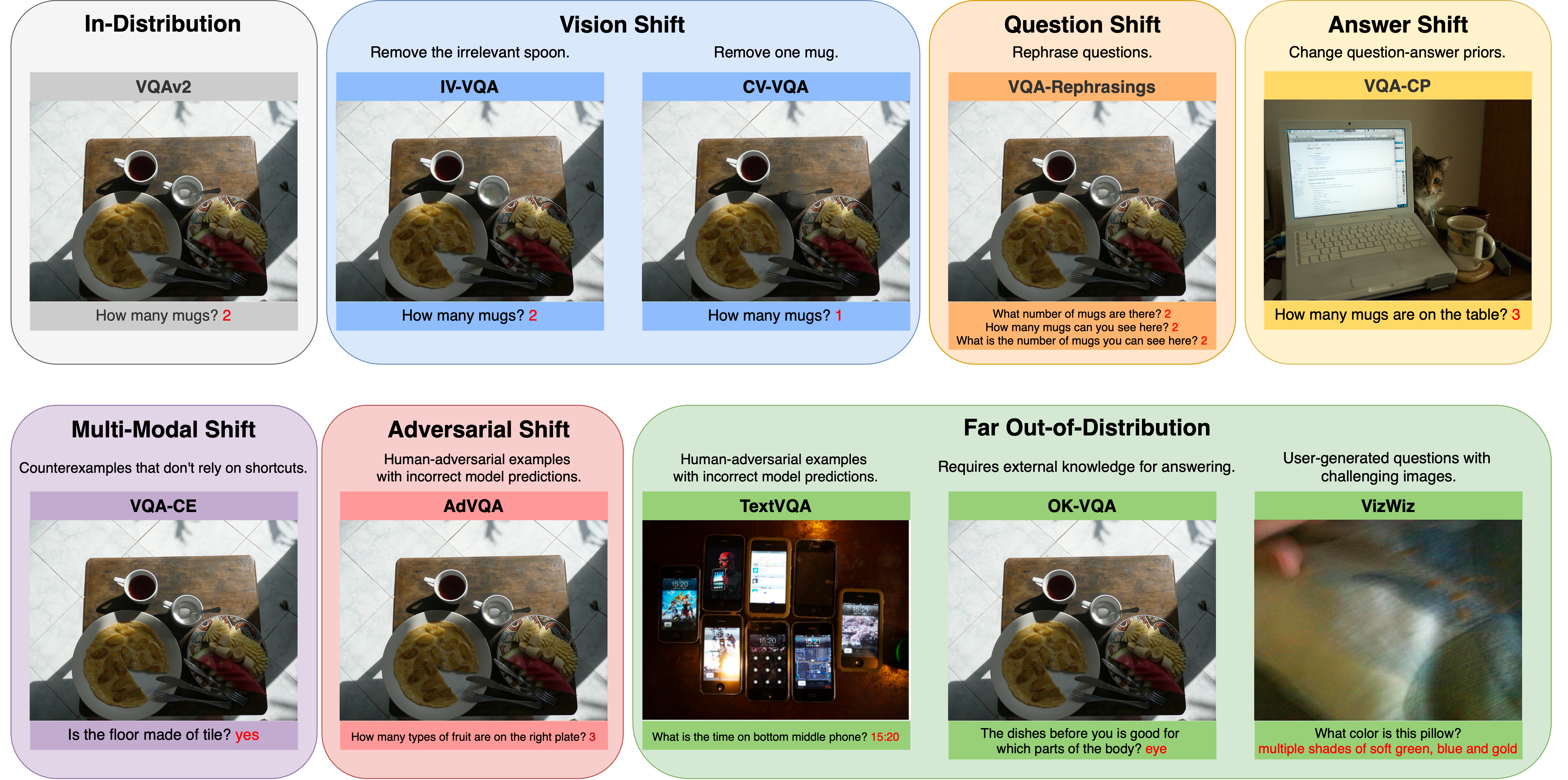}
     \caption{ID and OOD VQA datasets with Uni-modal (Vision, Question, Answer), Multi-modal, Adversarial and Far Distribution Shifts. }
      
     \label{fig:vqa_datasets}
\end{figure*}

We propose a new setting \benchmarkname~for benchmarking fine-tuning robustness across multi-modal shifts in VQA. We conduct a comprehensive categorization and experimentation on various VQA datasets, chosen to span different types of distribution shifts. We then evaluate the model on nine OOD 
datasets which are not included in the training with image, question, answer, multi-modal and adversarial distribution shift. We further distinguish between near and far OOD, concepts initially defined in OOD detection~\cite{fort2021exploringlimitsoutofdistributiondetection,winkens2020contrastivetrainingimprovedoutofdistribution}, where near OOD represents data that is perceptually similar but semantically dissimilar to the training distribution, while far OOD refers to data that is both perceptually and semantically dissimilar. In \benchmarkname, we categorize six near OOD datasets that exhibit various types of distribution shifts relative to VQAv2, and three far OOD datasets where both image and text sources differ from those in VQAv2. Additional results using GQA~\citep{hudson2019gqanewdatasetrealworld} and GQA-OOD~\citep{kervadec2021rosesredvioletsblue} can be found in Suppl.~\ref{sec:suppl_gqa}.

\subsection{Datasets}
Fig.~\ref{fig:vqa_datasets} and Tab.~\ref{tab:dataset_desc} provide an overview and statistics of all ID and OOD VQA datasets with vision, question, answer, multi-modal, adversarial and far distribution shift.

\vspace{2.5pt} \noindent \textbf
{ID Dataset.}
VQAv2~\citep{goyal_making_2017} contains open-ended questions about images with an emphasis on reducing answer biases through balanced pairs of question-image examples.  We choose this as our ID datasets as it is widely used as a benchmark for popular vision-language models (VLMs), and most other OOD datasets are derived from it.

\vspace{2.5pt} \noindent \textbf
{OOD Datasets.}
1) \textit{Distribution Shifts to Images.} IV-VQA~\citep{agarwal_towards_2020} and CV-VQA~\citep{agarwal_towards_2020} remove objects irrelevant and relevant to answering the question, resulting in unchanged and changed answers respectively.
2) \textit{Distribution Shifts to Questions.} VQA-Rephrasings~\citep{shah_cycle-consistency_2019} provides three alternative phrasings for each question. 
3) \textit{Distribution Shifts to Answers.} VQA-CP~\citep{agrawal_dont_2018} disrupts the usual correlation between question types and answers, creating a shift in answer patterns.
4) \textit{Distribution Shifts to Multi-modalities.} VQA-CE~\citep{dancette_beyond_2021} selects counterexample instances from VQAv2 validation set that highlight potential multi-modal shortcuts.
5) \textit{Adversarial Distribution Shifts.} AdVQA~\citep{sheng_human-adversarial_2021} includes human-adversarial examples where the model’s initial answer is incorrect.
6) \textit{Far OODs.} TextVQA~\citep{singh_towards_2019} requires textual comprehension within images to answer questions. VizWiz~\citep{bigham_vizwiz_nodate} features user-generated images with quality and relevance challenges. OK-VQA~\citep{marino2019okvqavisualquestionanswering} involves questions that require external knowledge beyond the image content. Tab.~\ref{tab:dataset_desc} shows the statistics of each dataset.

\begin{table}[!h]
\centering
\caption{ID and OOD VQA datasets. We use VQAv2 train for fine-tuning and VQAv2 val for model selection. For other OOD datasets, we only use the test splits for evaluation.}
\label{tab:dataset_desc}
\begin{tabular}{@{}ccc@{}}
\toprule
Shift Type & Dataset & \# Samples \\ \midrule
Source & VQAv2~\citep{goyal_making_2017} & \begin{tabular}[c]{@{}l@{}}Train: 443752\\ Val: 214354\end{tabular} \\ 
\midrule
\multirow{2}{*}{Image} & IV-VQA~\citep{agarwal_towards_2020} & 119907 \\ 
& CV-VQA~\citep{agarwal_towards_2020} & 4141 \\ 
\multirow{1}{*}{Question} & VQA-Rep.~\citep{shah_cycle-consistency_2019} & 162020 \\ 
Answer & VQA-CP~\citep{agrawal_dont_2018} & 219928 \\ 
\multirow{1}{*}{Multi-modal} 
& VQA-CE~\citep{dancette_beyond_2021} & 63298 \\ 
\multirow{1}{*}{Adversarial} & AdVQA~\citep{sheng_human-adversarial_2021} & 10000 \\ 
\midrule
\multirow{3}{*}{Far OOD} 
& TextVQA~\citep{singh_towards_2019} & 5000 \\ 
& VizWiz~\citep{bigham_vizwiz_nodate} & 3173 \\ 
& OK-VQA~\citep{marino2019okvqavisualquestionanswering} & 5046 \\ \bottomrule
\end{tabular}
\end{table}

\vspace{2.5pt} \noindent \textbf{Evaluation and Metrics.} We adopt the evaluation metric from VQAv2~\citep{goyal_making_2017}, which measures accuracy by comparing predicted answers to ground truth human-annotated answers. Each question is paired with 10 human-provided answers, and the accuracy is computed as:
$\text{Accuracy} = \min\left(\frac{\text{number of humans who gave the answer}}{3}, 1\right)
$.

\subsection{Measuring Distribution Shifts across Datasets}
We aim to quantify distribution shift to more precisely measure model robustness (e.g., relationship between performance and the degree of shift). In VQA, the presence of both uni-modal and multi-modal shifts complicates qualitative analysis of the types of shifts affecting performance. By quantifying these shifts individually, we can better understand how each modality—both independently and in combination—impacts model robustness.%

\vspace{2.5pt} \noindent \textbf{Mahalanobis Distance.} We use the Mahalanobis distance to measure the distribution shift following procedures similar to typical feature-based OOD detection methods \citep{Shi_2024_WACV}. Further analysis using Maximum Mean Discrepancy~\citep{gretton2008kernelmethodtwosampleproblem} is shown in Suppl.~\ref{sec:suppl_mmd}. Specifically, given our input training split \( X_{\text{in}}^{\text{train}} \), we compute feature representations \( z \) of the training samples to estimate the empirical mean \( \mu_\text{train} \) and sample covariance matrix \( \Sigma_\text{train} \). For each test split, we compute the test set shift relative to the training domain using the Mahalanobis distance defined in Eq.~\ref{eq:maha_distance}. The overall shift score for each test dataset, denoted as \( S_{\text{maha}} \), is calculated as the average 
\( S_{\text{Maha}} \) across all samples. 

\begin{equation}
S_{\text{Maha}}(z_{\text{test}}) = \sqrt{(z_{\text{test}} - \mu_{\text{train}})^\top \Sigma^{-1}_{\text{train}} (z_{\text{test}} - \mu_{\text{train}})}
\label{eq:maha_distance}
\end{equation}

We use the VQAv2 training set as our ID dataset. The average Mahalanobis score provides an overall measure of how distant the test set is from the ID set, with higher values indicating more shift.

\vspace{2.5pt} \noindent \textbf{Embedding Extraction.} Let \( q \) denote the question and \( v \) the image. The input features used in measuring shifts include uni-modal embeddings \( f(v) \) and \( f(q) \) and joint embeddings \( f(v,q) \).
We first leverage ViT~\cite{dosovitskiy2021imageworth16x16words} to get \( f(v) \) and BERT~\cite{devlin2019bertpretrainingdeepbidirectional} to get \( f(q) \) for pure uni-modal embeddings. Both representations are derived by mean-pooling the last layer of their respective pre-trained models. 
We use the pre-trained and fine-tuned PaliGemma from various methods to extract both uni-modal and joint embeddings, i.e., \( f(v) \), \( f(q) \), and \( f(v,q) \). For the image embedding \( f(v) \), we obtain it via masking out the question input tokens and mean-pooling the image portion from the final layer of the model before the language model head. Similarly, to get \( f(q) \), we mask out the image tokens and extract the question portion from the final layer. To obtain \(f(v,q)\), we pass in both image and text tokens as input, compute the average embedding for both modalities and then taking the overall mean. 
All embeddings are summarized in Tab.~\ref{tab:emb_sum}. 
We display the histograms for the Mahalanobis score distribution between test datasets with the ID set in Suppl.~\ref{sec:suppl_dist_hist}. Results of the distances and further analysis will be presented in the following sections.
\begin{table}[h]
    \centering
    \caption{Embedding extractions from different layers and backbone models.}
    \label{tab:emb_sum}
    \begin{tabular}{c|ccc} %
        \toprule
        Embeddings & 
        Model &
        Modality &
        PT/FT 
        \\
        \midrule
        \( f(v) \) &
        ViT &
        V &
        PT \\
        \( f(q) \) &
        BERT &
        Q &
        PT \\
        \midrule
        \( f_{\text{pt}}(v) \) &
        PaliGemma &
        V &
        PT \\
        \( f_{\text{pt}}(q) \) &
        PaliGemma &
        Q &
        PT \\
        \( f_{\text{pt}}(v,q) \) &
        PaliGemma &
        V,Q &
        PT \\
        \midrule
        \( f_{\text{ft$\_$method}}(v) \) &
        PaliGemma &
        V &
        FT \\
        \( f_{\text{ft$\_$method}}(q) \) &
        PaliGemma &
        Q &
        FT \\
        \( f_{\text{ft$\_$method}}(v,q) \) &
        PaliGemma &
        V,Q &
        FT \\
        \bottomrule
    \end{tabular}
\end{table}

\section{Robust Fine-Tuning}
In this section, we summarize several existing robust fine-tuning methods and evaluate their performance on our proposed benchmark \benchmarkname.
\subsection{Robust Fine-Tuning Baselines}
\begin{figure*}[h]
    \centering 
    \includegraphics[width=\linewidth]{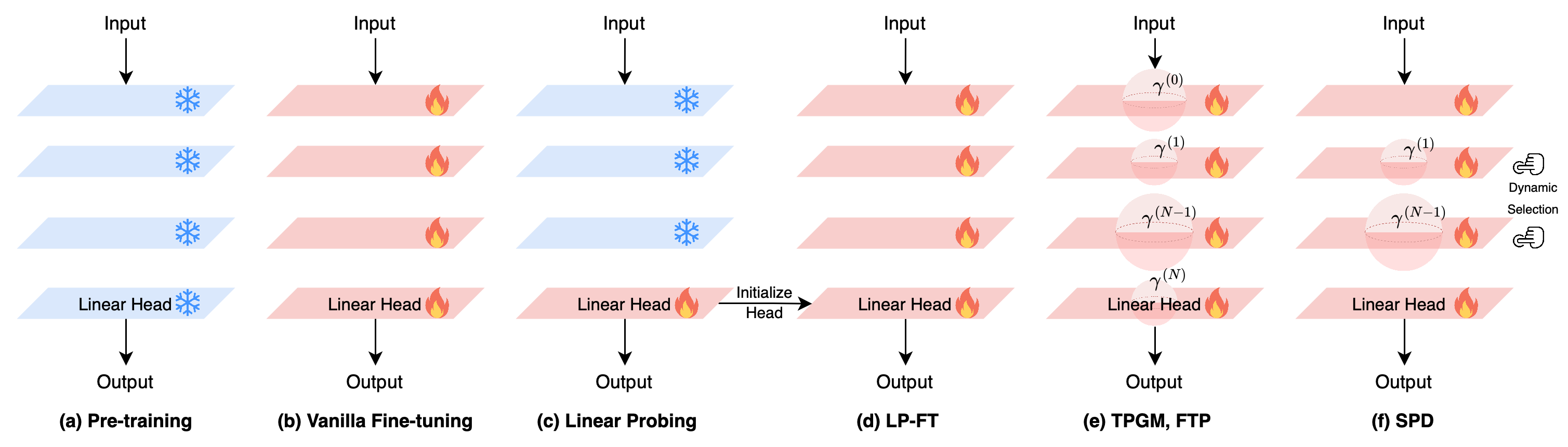} 
    \caption{Overview of Robust Fine-Tuning Methods. The flame icon represents tunable layers, while the snowflake icon represents frozen layers. $\gamma$ in TPGM, FTP and SPD is the constraint for each layer.}
    \label{fig:robust_ft_methods}
\end{figure*}

We include pre-training~\citep{huggingface_paligemma_3b_pt_224}, Vanilla Fine-Tuning, Linear Probing, LP-FT~\cite{kumar_fine-tuning_2022}, WiSE-FT~\cite{wortsman_robust_2022}, FTP~\cite{tian_fast_2023} and SPD~\cite{tian_rethinking_2024} as baselines. Fig.~\ref{fig:robust_ft_methods} provides an overview of all methods. 

Vanilla fine-tuning updates model parameters to adapt to a new task without constraint, minimizing task-specific loss 
$\mathcal{L}$ with regularization on the L2 norm of the parameters with a strength of $\lambda$:
\begin{equation}
    \min_{\lambda|(x,y)\in\mathcal{D}_{val}} \mathcal{L}(x,y;\argmin_{\theta_t|(x,y)\in\mathcal{D}_{tr}} \mathcal{L}(x,y;\theta_t) + \lambda \|\theta_t\|^2_2,\lambda)
\end{equation}

However, such free fine-tuning can cause feature distortion by excessively altering pre-trained representations. Linear Probing mitigates this by training only the final linear head while keeping other layers frozen, preserving the learned features. LP-FT~\citep{kumar_fine-tuning_2022} combines these two approaches and further proposes a two-step strategy of linear probing then full fine-tuning to achieve better adaptation while maintaining pre-trained features.

WiSE-FT~\citep{wortsman_robust_2022} introduces an interpolation technique that combines the strengths of pre-trained and fine-tuned embeddings by linearly blending their weights, with $\alpha \in [0,1]$ to control the balance:
\begin{equation}
    \tilde{\theta} = \alpha \theta_t + (1-\alpha) \theta_0, \alpha \in [0, 1]
\end{equation}

L2-SP~\citep{li_explicit_2018} applies an L2 penalty on the deviation between fine-tuned and pre-trained weights, rather than regularizing the L2 norm of the parameters themselves. This explicit penalty encourages the fine-tuned model to stay closer to the pre-trained weights:
\begin{equation}
    \min_{\lambda|(x,y)\in\mathcal{D}_{val}} \mathcal{L}(x,y;\argmin_{\theta_t|(x,y)\in\mathcal{D}_{tr}} \mathcal{L}(x,y;\theta_t) + \frac{\lambda}{2}\|\theta_t-\theta_0\|^2_2,\lambda)
\end{equation}

TPGM~\citep{tian_trainable_2023} approaches the regularization term in L2-SP as a constraint, reformulating the problem as a \textit{bi-level optimization} and enforcing the model to stay within a distance of $\gamma$ from the pre-trained model, as shown in Eq.~\ref{eq:tpgm}:
\begin{align}
\label{eq:tpgm}
     \min_{\lambda,\gamma|(x,y)\in\mathcal{D}_{val}} \mathcal{L}(x,y;\argmin_{\theta_t|(x,y)\in\mathcal{D}_{tr}}  \mathcal{L}(x,y;\theta_t,\lambda),\lambda),\\\nonumber \quad\text{s.t.}\quad \|\theta_t-\theta_0\|_2\leq\gamma,
\end{align}

To solve the constrained optimization problem, TPGM utilizes the \textit{Projected Gradient Method} (PGM) to projects the updated model weights to be within the constraint, illustrated in Eq.~\ref{eq:l2_proj}:
\begin{align}
\label{eq:l2_proj}
\Pi_{l2}(\theta_0,\theta_t,\gamma):\tilde{\theta} = \theta_0 + &\frac{1}{\text{max}\left(1,\frac{\|\theta_t-\theta_0\|_2}{\gamma}\right)}(\theta_t-\theta_0)
\end{align}

FTP~\citep{tian_fast_2023} further improves the efficiency of TPGM~\citep{tian_trainable_2023} by learning the constraint from training set of previous step instead of the current validation set, which is highlighted in blue in Eq.~\ref{eq:ftp}:
\begin{align}
\label{eq:ftp}
     \min_{\lambda,\gamma|(x,y)\in\textcolor{blue}{\mathcal{D}_{tr}}} \mathcal{L}(x,y;\argmin_{\theta_t|(x,y)\in\mathcal{D}_{tr}}  \mathcal{L}(x,y;\theta_t,\lambda),\lambda),\\\nonumber \quad\text{s.t.}\quad \|\theta_t-\theta_0\|_2\leq\gamma,
\end{align}

SPD~\citep{tian_rethinking_2024} selectively imposes a strong penalty on certain layers while allowing others to change freely, with a selecting condition expressed in Eq.~\ref{eq:spd_condition}, where $g_{t+1}=\frac{\partial \mathcal{L}(\theta_{t})}{\partial \theta_{t}}$ represents the gradient in step $t$:
\begin{align}
    \label{eq:spd_condition}
    c_t := -g_{t+1}^\intercal({\theta}_t-\theta_0)
\end{align}

Intuitively, SPD expands and contracts the parameter search space for layers with consistent and inconsistent loss reduction between the descent direction $(-g_{t+1})$ and the current progress direction $(\theta_{t-1}-\theta_0)$, respectively. For the layers that meet the condition Eq.~\ref{eq:spd_condition}, SPD projects the corresponding layers using PGM in Eq.~\ref{eq:l2_proj}.

\subsection{ID, Near \& Far OOD Performance} 
We fine-tune Google's recently released PaliGemma-3B~\citep{beyer_paligemma_2024} model on the VQAv2 dataset and evaluate on the other OOD datasets.
PaliGemma-3B is lightweight and one of the state-of-the-art models on VQAv2, making it a practical option for benchmarking.
We apply LoRA~\citep{hu_lora_2021}, a parameter-efficient fine-tuning method, to reduce the computational and memory overhead associated with fine-tuning large models, as demonstrated in prior work~\citep{hu2023llmadaptersadapterfamilyparameterefficient,zhu2023minigpt4enhancingvisionlanguageunderstanding}.
While LoRA limits excessive parameter updates to maintain robustness~\citep{biderman2024loralearnsforgets}, as shown in Tab.~\ref{tab:main_result} there is still a loss of robustness compared to vanilla fine-tuning and other robust fine-tuning methods.
The results of the robust fine-tuning methods are shown in Tab.~\ref{tab:main_result}.
Training details including configurations for different methods and additional results using full fine-tuning and LLaVA~\citep{liu2023visualinstructiontuning} can be found in Suppl.~\ref{sec:training} and~\ref{sec:suppl_llava}. 
Below we discuss our observations of the results.

\newcommand{\colormap}[1]{%
  \pgfmathsetmacro{\value}{#1}%
  \ifnum\value<25
    \cellcolor{blue!15}#1
  \else
    \ifnum\value<35
      \cellcolor{blue!35}#1
    \else
      \ifnum\value<45
        \cellcolor{blue!55}#1
      \else
        \ifnum\value<55
          \cellcolor{blue!75}#1
        \else
          \cellcolor{blue!95}#1
        \fi
      \fi
    \fi
  \fi
}

\begin{table*}[!h]
\centering
\caption{Visual Question Answering Fine-Tuning Results for ID, Near OOD and Far OOD datasets. \textbf{Bold}: best. \underline{Underline}: second best.}
\label{tab:main_result}
\resizebox{\linewidth}{!}{
\begin{tabular}{@{}c|cccccccc|cccc|c@{}}
\toprule
\multirow{2}{*}{} &
  \multicolumn{1}{c|}{ID} &
  \multicolumn{7}{c|}{Near OOD} &
  \multicolumn{4}{c|}{Far OOD} &
  \multirow{3}{*}[-1.5em]{\rotatebox{90}{OOD Avg.}}
  \\ \cmidrule{2-13} 
  &
  \multicolumn{1}{c|}{\multirow{2}{*}[-0.4em]{\rotatebox{90}{VQAv2 (val)}}} &
  \multicolumn{2}{c|}{Vis.} &
  \multicolumn{1}{c|}{Ques.} &
  \multicolumn{1}{c|}{Ans.} &
  \multicolumn{1}{c|}{M.M.} &
  \multicolumn{1}{c|}{Adv.} &
  \multirow{2}{*}[0.2em]{\rotatebox{90}{Near OOD Avg.}}&
  \multirow{2}{*}[-1.3em]{\rotatebox{90}{TextVQA}}&
  \multirow{2}{*}[-2.2em]{\rotatebox{90}{VizWiz}}&
  \multicolumn{1}{c|}{\multirow{2}{*}[-1.25em]{\rotatebox{90}{OK-VQA}}}&
  \multicolumn{1}{c|}{\multirow{2}{*}[-0.1em]{\rotatebox{90}{Far OOD Avg.}}}&
   \\ 
   \cmidrule{3-8}
 &
  \multicolumn{1}{c|}{} &
  \rotatebox{90}{IV-VQA} &
  \multicolumn{1}{c|}{\rotatebox{90}{CV-VQA}} &
  \multicolumn{1}{c|}{\rotatebox{90}{VQA-Rep.}} &
  \multicolumn{1}{c|}{\rotatebox{90}{VQA-CP}} &
  \multicolumn{1}{c|}{\rotatebox{90}{VQA-CE}} &
  \multicolumn{1}{c|}{\rotatebox{90}{AdVQA}} & 
  \multicolumn{1}{c|}{} &
   &
   &
  \multicolumn{1}{c|}{} &
  \multicolumn{1}{c|}{} &
   \\ 

\midrule

\textcolor{gray}{Zero-Shot~\cite{huggingface_paligemma_3b_pt_224}} &
  \multicolumn{1}{c|}{\textcolor{gray}{54.42}} &
  \textcolor{gray}{63.95} &
  \multicolumn{1}{c|}{\textcolor{gray}{44.72}} &
  \multicolumn{1}{c|}{\textcolor{gray}{50.10}} &
  \multicolumn{1}{c|}{\textcolor{gray}{54.29}} &
  \multicolumn{1}{c|}{\textcolor{gray}{30.68}} &
  \multicolumn{1}{c|}{\textcolor{gray}{30.46}} &
  \textcolor{gray}{45.70} &
  \textcolor{gray}{14.86} &
  \textcolor{gray}{16.84} &
  \multicolumn{1}{c|}{\textcolor{gray}{28.60}} &
  \multicolumn{1}{c|}{\textcolor{gray}{20.10}} &
  \textcolor{gray}{37.17} \\

Vanilla FT$_\text{LoRA}$~\cite{hu_lora_2021} &
  \multicolumn{1}{c|}{\underline{86.29}} &
  \underline{94.43} &
  \multicolumn{1}{c|}{\textbf{69.36}} &
  \multicolumn{1}{c|}{\underline{78.90}} &
  \multicolumn{1}{c|}{\underline{86.21}} &
  \multicolumn{1}{c|}{\underline{71.73}} &
  \multicolumn{1}{c|}{49.82} &
  \underline{75.08} &
  42.08 &
  22.92 &
  \multicolumn{1}{c|}{48.30} &
  \multicolumn{1}{c|}{37.77} &
  \underline{62.64}  \\ 
Linear Prob$_\text{LoRA}$ &
  \multicolumn{1}{c|}{78.24} &
  87.83 &
  \multicolumn{1}{c|}{63.87} &
  \multicolumn{1}{c|}{69.61} &
  \multicolumn{1}{c|}{78.48} &
  \multicolumn{1}{c|}{61.66} &
  \multicolumn{1}{c|}{42.90} &
  67.39 &
  29.61 &
  18.80 &
  \multicolumn{1}{c|}{42.27} &
  \multicolumn{1}{c|}{30.23} &
  55.00  \\
LP-FT$_\text{LoRA}$~\cite{kumar_fine-tuning_2022} &
  \multicolumn{1}{c|}{85.97} &
  93.30 &
  \multicolumn{1}{c|}{65.93} &
  \multicolumn{1}{c|}{76.49} &
  \multicolumn{1}{c|}{86.16} &
  \multicolumn{1}{c|}{72.73} &
  \multicolumn{1}{c|}{45.68} &
  73.38 &
  31.41 &
  19.01 &
  \multicolumn{1}{c|}{43.27} &
  \multicolumn{1}{c|}{31.23} &
  59.33  \\
WiSE-FT$_\text{LoRA}$~\cite{wortsman_robust_2022} &
  \multicolumn{1}{c|}{71.36} &
  85.06 &
  \multicolumn{1}{c|}{64.55} &
  \multicolumn{1}{c|}{66.42} &
  \multicolumn{1}{c|}{70.89} &
  \multicolumn{1}{c|}{48.74} &
  \multicolumn{1}{c|}{43.95} &
  63.27 &
  36.98 &
  22.41 &
  \multicolumn{1}{c|}{42.35} &
  \multicolumn{1}{c|}{33.91} &
  53.48  \\
FTP$_\text{LoRA}$~\cite{tian_fast_2023} &
  \multicolumn{1}{c|}{81.77} &
  92.61 &
  \multicolumn{1}{c|}{67.93} &
  \multicolumn{1}{c|}{76.66} &
  \multicolumn{1}{c|}{81.41} &
  \multicolumn{1}{c|}{64.14} &
  \multicolumn{1}{c|}{\textbf{50.99}} &
  72.29 &
  \textbf{49.12} &
  \textbf{25.67} &
  \multicolumn{1}{c|}{\textbf{51.07}} &
  \multicolumn{1}{c|}{\textbf{41.95}} &
  62.18\\
SPD$_\text{LoRA}$~\cite{tian_rethinking_2024} &
      \multicolumn{1}{c|}{\textbf{87.39}} &
       \textbf{95.25} &
      \multicolumn{1}{c|}{\underline{68.85}} &
      \multicolumn{1}{c|}{\textbf{79.48}} &
      \multicolumn{1}{c|}{\textbf{87.27}} &
      \multicolumn{1}{c|}{\textbf{73.52}} &
      \multicolumn{1}{c|}{\underline{50.90}} &
      \textbf{75.88} &
      \underline{43.56} &
      \underline{23.05} &
      \multicolumn{1}{c|}{\underline{50.11}} &
      \underline{38.91} &
      \textbf{63.55}\\

\bottomrule
\end{tabular}
}
\end{table*}

\vspace{2.5pt} \noindent \textbf{Vanilla fine-tuning improves zero-shot performance across ID, near OOD, and far OOD datasets.} 
As shown in Tab.~\ref{tab:main_result}, we observe no degradation in OOD performance compared to zero-shot performance following vanilla fine-tuning.
This differs from the observation in~\citep{wortsman_robust_2022} that vanilla fine-tuning for image classification degrades OOD performance compared to zero-shot. In the image classification task, fine-tuning typically involves using only the image encoder from the backbone model with a linear classification head, removing the text encoder and employing a cross-entropy loss that differs from the pre-training objective. However,~\citep{goyal2023finetune} notes that fine-tuning is more robust when retaining the same objective as pre-training, which may explain why vanilla fine-tuning performs robustly in VQA tasks, as the model structure and loss function remain consistent. Model architecture and task characteristics could also contribute to the observed robustness.

\vspace{2.5pt} \noindent \textbf{WiSE-FT~\cite{wortsman_robust_2022} decreases both ID and OOD performance in VQA.} \citep{wortsman_robust_2022} demonstrate that ensembling the weights of zero-shot and fine-tuned models can benefit from the robustness of zero-shot and the adaptability of fine-tuning in image classification tasks. However, WiSE-FT is highly dependent on a reduction in robustness after vanilla fine-tuning and the linear connectivity of the model, which does not occur in our VQA setting.
In Tab.~\ref{tab:main_result}, we see a significant gap between zero-shot and fine-tuned VQA performance, which limits the effectiveness of WiSE-FT.
As a result, combining pre-trained and fine-tuned weights reduces the model’s robustness across all shifts, making WiSE less effective than vanilla fine-tuning for VQA.

\vspace{2.5pt} \noindent \textbf{SPD~\cite{tian_rethinking_2024} achieves the highest performance on ID, near and overall OOD.}
As shown in Tab.~\ref{tab:main_result}, SPD achieves the highest scores on ID, near OOD, and overall OOD, demonstrating robustness across various types of distribution shifts, including vision, question, answer, and multi-modal combinations, as well as adversarial shifts.

\vspace{2.5pt} \noindent \textbf{FTP~\cite{tian_fast_2023} underfits ID data but excels on far OOD datasets.}
FTP~\citep{tian_trainable_2023} underfits the ID dataset, showing significantly lower ID performance than vanilla fine-tuning, even with a minimal positive gradient annealing factor ($\kappa=0$), indicating the weakest regularization. However, FTP performs exceptionally well on far OOD tasks, achieving the highest scores across the three far OOD datasets. This outcome may be due to FTP’s stronger regularization, as the projection constraints are non-decreasing throughout training, providing consistent regularization. The FTP authors also suggest that $\kappa=0$ is necessary for optimal performance when underfitting is observed. Compared to SPD, which applies a weaker regularization and performs better on ID and near OOD, FTP’s strict regularization might uniquely favor far OOD performance. The substantial improvement in far OOD, despite a weaker zero-shot baseline, presents an interesting area for future investigation.
\section{Analysis on the Shift Scores}
\label{sec:analysis_shift_score}
In this section, we aim to deepen our understanding of how multi-modal distribution shifts impact model robustness by analyzing 1) shift distances across different embeddings, datasets and fine-tuning methods, 2) interactions between uni- and multi-modal shifts.

\begin{figure*}[!h]
    \centering 
    \includegraphics[width=\linewidth]{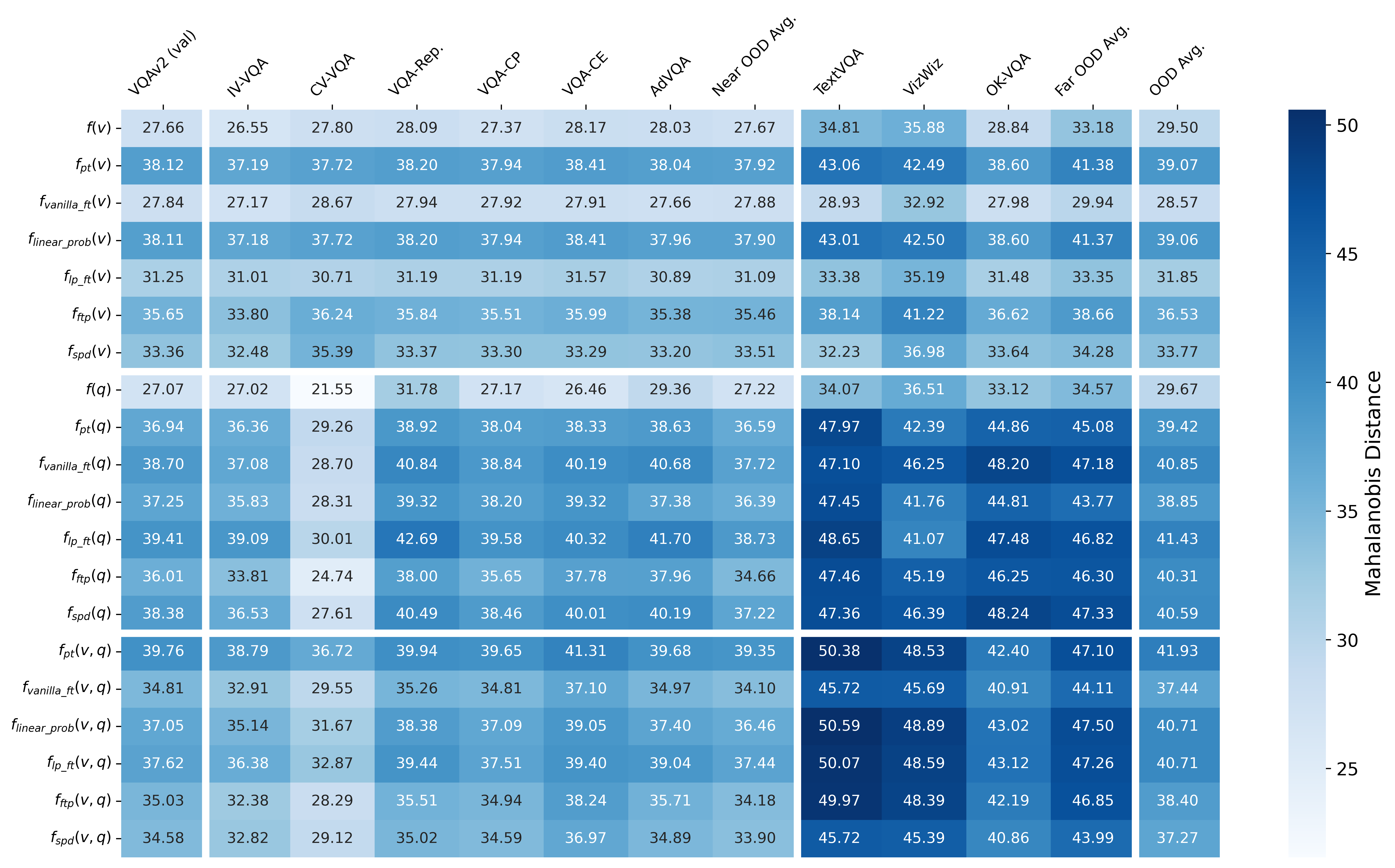} 
    \caption{Mahalanobis Distance Heatmap of Different Datasets, Embeddings and Fine-Tuning Methods. Darker blue represents higher distance score. From top to bottom the three groups are image, question and joint shift scores respectively.
    }
    \label{fig:dist_heatmap}
\end{figure*}

We display the Mahalanobis distance of different datasets, embeddings and fine-tuning methods and its heatmap in Fig.~\ref{fig:dist_heatmap}. The darker blue represents higher distance score which indicates larger distribution shifts.

\subsection{Shift Distance across Embeddings, Datasets \& Fine-Tuning Techniques}

As shown in Fig.~\ref{fig:dist_heatmap}, shift scores increase from left to right, with a corresponding darkening in color, which reflects our intuitive understanding of relative dataset shifts. An outlier is CV-VQA, which has the lowest shift scores, likely due to its small sample size or inherent dataset issues. Additionally, question, image, and joint shifts exhibit high negative correlations with performance, with values of -0.63, -0.74, -0.78 under Vanilla-FT, respectively. Full details on shift-performance correlations for various fine-tuning methods are provided in Suppl.~\ref{sec:suppl_dist_perf}, supporting the intuition that larger shifts in all modalities degrade VQA performance.

As we move from near to far OOD datasets, question shifts increase significantly, while image shifts increase at a lower rate, suggesting strong adaptivity to visual changes. This indicates that far OOD datasets introduce stronger task-guided and contextual shifts, and VLM tends to capture shifts more strongly in question representations. 

Additionally, PaliGemma’s visual, question and joint embeddings show similar variability on near OODs, whilst far OODs reveal greater sensitivity in question and joint embeddings, evidenced by higher Mahalanobis distances. In comparison, pure uni-modal embeddings from ViT and BERT remain more stable. This suggests that question and joint embeddings from multi-modal models are more sensitive to significant distribution shifts, likely due to their dependence on contextual and multi-modal interactions.

\subsection{Correlation between Uni- \& Multi-Modal Shifts}
The degree of joint shift can be influenced by both visual and question shifts. We aim to reveal how much direct influence each modality has towards the overall joint shift by computing the Pearson correlation coefficient between \( \{ f(v), f(v, q) \} \) and \( \{ f(q), f(v, q) \} \) for each test set. Results are shown in Tab.~\ref{tab:concept_corr_final}. Full correlation breakdown for each dataset can be viewed in Suppl.~\ref{sec:suppl_xconcept}.  
\begin{table}[h]
    \centering
    \caption{Uni- and multi-modal shifts correlation averaged across datasets for different fine-tuning methods.}
    \label{tab:concept_corr_final}
    \small %
    \begin{tabular}{c|c|c}
        \toprule
         & \( \{ f(v), f(v, q) \} \) & \( \{ f(q), f(v, q) \} \) \\
        \midrule
        Pre-Train~\citep{huggingface_paligemma_3b_pt_224} & 0.32 & 0.34 \\
        Vanilla FT$_\text{LoRA}$~\citep{hu_lora_2021} & 0.29 & 0.48  \\
        Linear Prob$_\text{LoRA}$ & 0.26 & 0.61  \\
        LP-FT$_\text{LoRA}$~\citep{kumar_fine-tuning_2022} & 0.34 & 0.63\\
        FTP$_\text{LoRA}$~\cite{tian_fast_2023} & 0.34 & 0.58  \\
        SPD$_\text{LoRA}$~\cite{tian_rethinking_2024} & 0.17 & 0.51  \\
        \bottomrule
    \end{tabular}
\end{table}

All fine-tuning methods has a significantly higher question-joint correlation than image-joint correlation, making the joint modality more sensitive to question shifts. However, pre-trained method maintains similar levels of correlation. Further, more robust fine-tuning methods show smaller image-joint and question-joint correlations, indicating that robust methods learn representations less sensitive to specific image-question pairings, making them less impacted by uni-modal shifts.

\section{Analysis on the Modality Importance}
\label{sec:analysis_mi}
We further explore how modality importance impacts model robustness by analyzing intra- versus inter-modality attention, shifts in modality focus between ID and OOD settings, and differences across fine-tuning methods. This analysis reveals how attention to specific modalities influences the model’s ability to generalize under distribution shifts.

\subsection{Intra- \& Inter-Modality Attention}

We introduce the following metrics to quantitatively analyze the modality importance, inspired by \citep{cao2020scenerevealingsecretspretrained}. Denote $v_1, v_2, \dots, v_N$ and $q_1, q_2, \dots, q_M$ as the image and question tokens respectively. The \textit{Modality Importance} (MI) of a token $tkn$ is defined as the ratio of its total attention to all question tokens to its total attention to all image tokens,

\begin{equation}
    \text{MI}(tkn) = \frac{\sum_{j=1}^M \text{Attn}(tkn, q_j)}{\sum_{i=1}^N \text{Attn}(tkn, v_i)}
\end{equation}

\noindent where $\text{Attn}(tkn_1, tkn_2)$ represents the average attention weights of all heads from $tkn_1$ to $tkn_2$. Note that the sum of the attention weights that a token attends to all the tokens in both modalities equals one.

Further, we want to explore the modality importance of the tokens in different modalities. We take the average MI of the tokens in each modality, which is expressed in Eq.~\ref{eq:mi_v_q},

\begin{equation}
\label{eq:mi_v_q}
    \text{MI}_v = \frac{1}{N} \sum_{i=1}^N \text{MI}(v_i),
    \text{MI}_q = \frac{1}{M} \sum_{j=1}^M \text{MI}(q_j)
\end{equation}.

MI$_m > 1$ indicates that the text modality is more influential than the image modality for tokens in modality $m$.

\begin{figure}[h!]
    \centering
    \begin{subfigure}{0.33\linewidth}
        \centering
        \includegraphics[width=\linewidth]{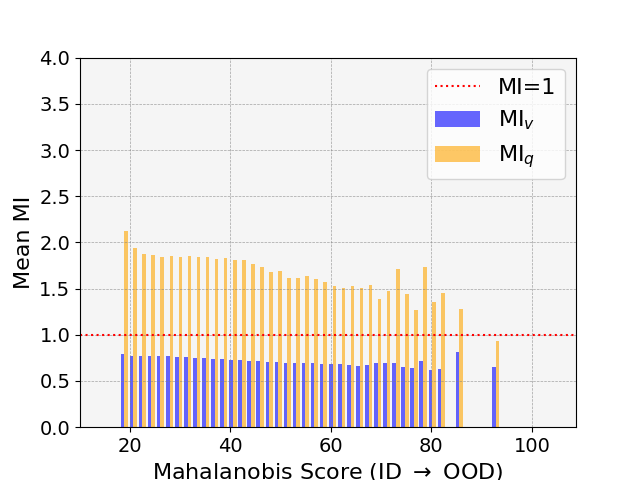}
        \caption{VQAv2, Vanilla FT}
    \end{subfigure}%
    \begin{subfigure}{0.33\linewidth}
        \centering
        \includegraphics[width=\linewidth]{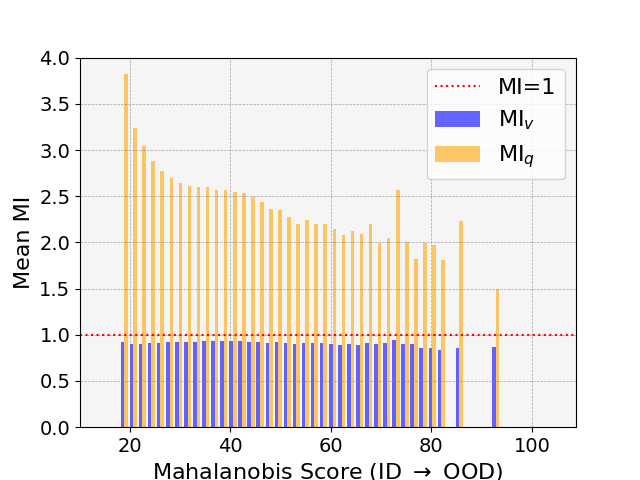}
        \caption{VQAv2, FTP}
    \end{subfigure}%
    \begin{subfigure}{0.33\linewidth}
        \centering
        \includegraphics[width=\linewidth]{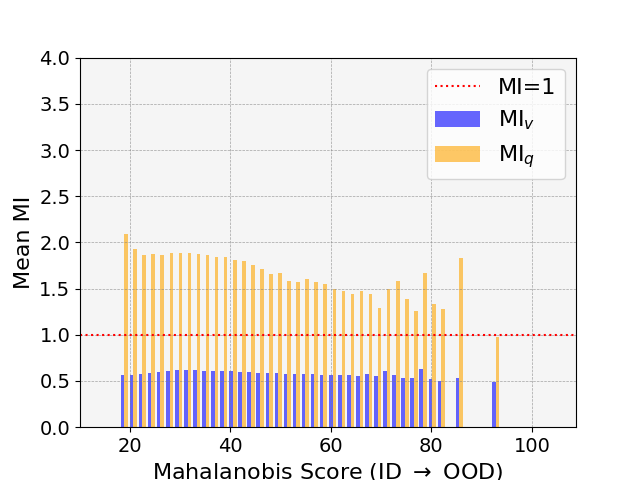}
        \caption{VQAv2, SPD}
    \end{subfigure}

    \vskip\baselineskip
    \begin{subfigure}{0.33\linewidth}
        \centering
        \includegraphics[width=\linewidth]{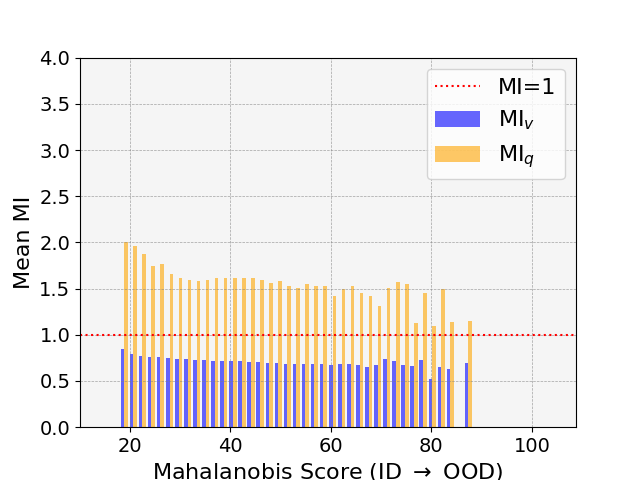}
        \caption{VQA-CE, Vanilla FT}
    \end{subfigure}%
    \begin{subfigure}{0.33\linewidth}
        \centering
        \includegraphics[width=\linewidth]{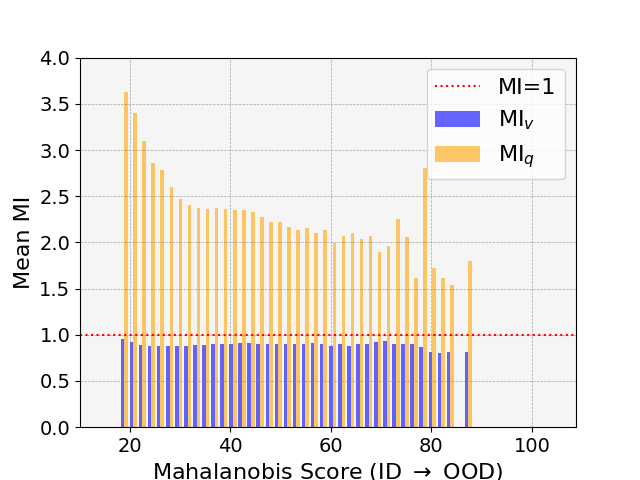}
        \caption{VQA-CE, FTP}
    \end{subfigure}%
    \begin{subfigure}{0.33\linewidth}
        \centering
        \includegraphics[width=\linewidth]{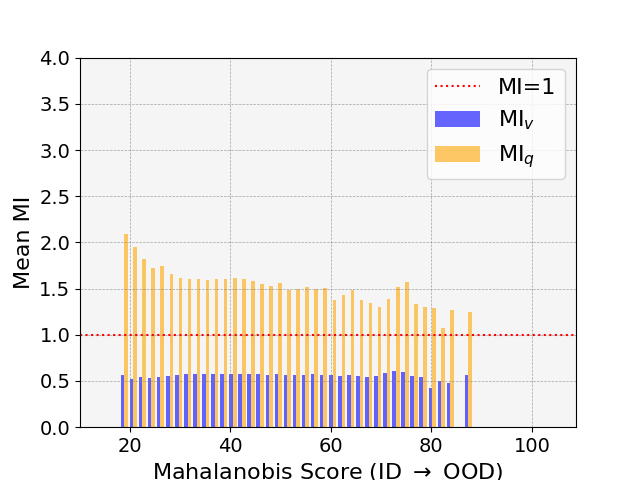}
        \caption{VQA-CE, SPD}
    \end{subfigure}

    \caption{Variation of MI$_v$ and MI$_q$ w.r.t. shift score under vanilla FT, FTP and SPD across VQAv2 and VQA-CE. The blue and orange bars represent  MI$_v$ and MI$_q$ respectively. The red dotted line marks a reference MI of 1.
    }

    \label{fig:modality_importance_hist}
\end{figure}

In Fig.~\ref{fig:modality_importance_hist}, we present the variation of MI$_v$ and MI$_q$ w.r.t. shift score under vanilla FT, FTP and SPD across VQAv2 and VQA-CE. Other datasets and baselines can be found in Suppl.~\ref{sec:suppl_xttn}. In Tab.~\ref{tab:attn_tab}, we further separate ID and OOD samples from all datasets with a Mahalanobis Distance of 60, chosen as the relative median of shift scores to illustrate the distinction between closer and more distant samples,
and show MI$_v$, MI$_q$ for different fine-tune methods.

\begin{table}[h]
    \centering
    \caption{MI$_v$ and MI$_q$ of ID and OOD samples for different fine-tuning methods. We use the Mahalanobis Distance of 60 to distinguish between ID and OOD samples.}
    \label{tab:attn_tab}
    \resizebox{1\linewidth}{!}{
    \begin{tabular}{c|cc|cc|cc} %
        \toprule
        &
        \multicolumn{2}{c|}{ID MI}& \multicolumn{2}{c|}{OOD MI}& \multicolumn{2}{c}{Overall MI}\\ 
        \cmidrule{2-7}
 & MI$_v$& MI$_q$& MI$_v$& MI$_q$& MI$_v$&MI$_q$\\
        \midrule
        Pre-Train~\cite{huggingface_paligemma_3b_pt_224}& 0.90& 2.37& 0.85& 1.93& 0.89&2.37\\
 Vanilla FT$_\text{LoRA}$~\cite{hu_lora_2021}& 0.75& 1.90& 0.66& 1.42& 0.75&1.89\\
 Linear Prob$_\text{LoRA}$& 0.71& 2.10& 0.66& 1.60& 0.71&2.10\\
 LP-FT$_\text{LoRA}$~\cite{kumar_fine-tuning_2022}& 0.94& 2.40& 0.78& 1.51& 0.94&2.39\\
 FTP$_\text{LoRA}$~\cite{tian_fast_2023}& 0.93& 2.72& 0.89& 1.97& 0.92&2.72\\
        SPD$_\text{LoRA}$~\cite{tian_rethinking_2024}& 0.61& 1.93& 0.57& 1.42& 0.61&1.93\\
        \bottomrule
    \end{tabular}}
\end{table}

\subsection{Comparison between Intra- \& Inter-Modality Attention}

According to Tab.~\ref{tab:attn_tab} and Fig.~\ref{fig:modality_importance_hist}, MI$_v < 1$ while MI$_q > 1$, implying dominant image and question attention for image and tokens respectively. The dominance of intra-modality attention is expected, as attention mechanisms tend to focus within the same modality to reinforce contextual coherence. However, Intra-modality attention of text tokens is greater than that of image tokens, since MI$_v$ is close to 1 while MI$_q$ is significantly greater than 1. This suggests that the ratio of question-to-question attention relative to question-to-image attention is higher than that of image-to-image attention relative to image-to-question, indicating stronger intra-modality attention for text tokens.

\subsection{Comparison of MIs of ID \& OOD}

As illustrated in Fig.~\ref{fig:modality_importance_hist} and Tab.~\ref{tab:attn_tab}, across each dataset and fine-tuning method, MI$_v$ remains relatively stable under distribution shift, whereas MI$_q$ shows a marked decrease as the distribution shift intensifies. Such pattern suggest that, in OOD samples, text tokens increasingly attend to image tokens. Such a shift could indicate a model bias, where ID samples rely heavily on intra-modality shortcuts for text, potentially at the expense of robust cross-modal integration. This reliance on shortcuts and lack of image grounding may reduce the model's ability to generalize effectively.

\subsection{Comparison of Fine-Tuning Methods}

According to Tab.~\ref{tab:attn_tab}, SPD, which performs best on ID and near-OOD datasets, has the lowest MI$_v$, while FTP, which excels on far-OOD datasets, shows the highest MI$_q$. One possible hypothesis is that for optimal performance across ID, near-OOD, and far-OOD settings, a model might benefit from a high MI$_q$ and low MI$_v$, suggesting that each modality should prioritize intra-modality attention over cross-modality attention. Stronger intra-modality focus—where each modality focuses on internal coherence before cross-modal integration—could potentially yield improved robustness across distribution shifts.

Analyses in Sec.~\ref{sec:analysis_shift_score} and~\ref{sec:analysis_mi} suggest that future work should aim to enhance robustness under language shifts and implement ways to dynamically handle modality importance. Our findings show that more robust methods exhibit higher intra-modality attention, highlighting the potential of adaptive attention mechanisms and modality-specific regularization to better balance modality contributions.

\section {Conclusion}

In conclusion, our paper introduces FRAMES-VQA, a new VQA benchmark to evaluate fine-tuning robustness under various distribution shifts. We categorize the shift types and quantify the degree, and conduct comparisons of robust fine-tuning baselines on the proposed benchmark. Our comprehensive analysis reveals the interactions between uni- and multi-modal shifts and how VLMs perform cross-modal learning in ID/OOD scenarios using the modality importance metric.
We observe that fine-tuning increases the influence of question shifts on multi-modal representations, with more robust methods showing a lower correlation between uni- and multi-modal shifts. Question-to-image attention rises for OOD samples, while robust methods prioritize intra-modality attention.
Our findings lay the foundation and highlight promising directions for future work: (1) Developing training methods to mitigate uni- and multi-modal correlations, and (2) Enhancing techniques for detecting and adapting to different types of distribution shifts, both of which proved crucial in our analysis.

\newpage
{
    \small
    \bibliographystyle{ieeenat_fullname}
    \bibliography{main}
}

\clearpage
\setcounter{page}{1}
\maketitlesupplementary

\section{Training Details}
\label{sec:training}

We use the model pretrained with $224*224$ input images and $128$ token input/output text sequences and fine-tune with the precision of bfloat16. We use the LAVIS~\citep{li2022lavislibrarylanguagevisionintelligence} public repository to fine-tune all methods. Standard hyper-parameters are used for all: learning rate ($1e-3$), weight-decay ($1e-4$), optimizer (AdamW), scheduler (Linear Warmup With Cosine Annealing), warm-up learning rate ($1e-4$), minimum learning rate ($1e-4$), accumulation steps ($2$), beam size (5). The model is trained for $10$ epochs with a batch size of $128$ for Tab.~\ref{tab:main_result}. For LoRA~\citep{hu_lora_2021}, we limit our study to only adapting the attention weights and freeze the MLP modules for parameter-efficiency, specifically apply LoRA to $W_q, W_k, W_v, W_o$ with $r=8$ in Tab.~\ref{tab:main_result}. The regularization hyper-parameter is found through cross-validation, and the model with the best ID validation accuracy is taken. We use 8 A40 GPU for each experiment. The best training configurations for different methods are listed in Tab.~\ref{tab:hyper_param}.

\begin{table}[htb]
        \centering
          \caption{\textbf{Best Training Configurations for Robust Fine-Tuning Methods.} 
          lr and wd stands for learning rate and weight decay.}
          \label{tab:hyper_param}
        {
       \begin{tabular}{@{}c|c|c|c}
        \toprule

        & lr& wd& others\\
        
        \midrule
          
        Vanilla FT &
          \multicolumn{1}{c|}{$1e-3$} &
          $1e-4$&
          \multicolumn{1}{c}{-} \\ 

        Linear Prob &
          \multicolumn{1}{c|}{$1e-3$} &
          $1e-4$&
          \multicolumn{1}{c}{-} \\ 

        LP-FT &
          \multicolumn{1}{c|}{$1e-3$}  &
          $1e-4$&
          \multicolumn{1}{c}{-} \\

        WiSE-FT &
          \multicolumn{1}{c}{-} &
          -&
          \multicolumn{1}{c}{$\alpha=0.5$} \\

        FTP &
          \multicolumn{1}{c|}{$1e-3$} &
          $1e-4$&
          \multicolumn{1}{c}{$\kappa=0$} \\

       SPD  &
          \multicolumn{1}{c|}{$1e-3$} &
          $0.5$&
          \multicolumn{1}{c}{-} \\

        \bottomrule
        \end{tabular}
        }
      \end{table}
 \section{Histograms of Shift Scores}
\label{sec:suppl_dist_hist}

\begin{figure*}[!h]
    \centering
    
    \begin{subfigure}[b]{0.3\linewidth}
        \centering
        \includegraphics[width=\linewidth]{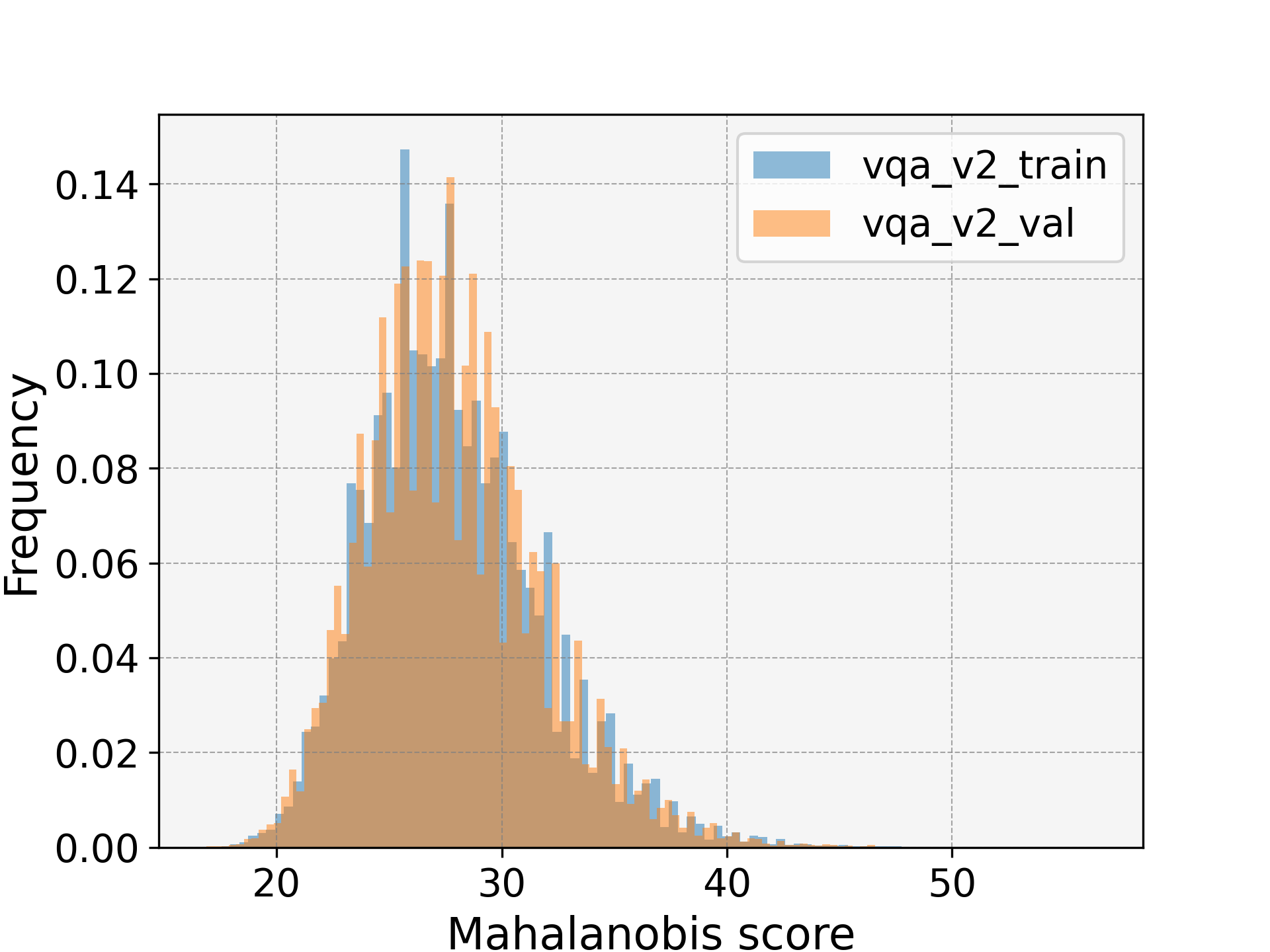}
        \caption{VQAv2 Val}
        \label{fig:VQAv2_val}
    \end{subfigure}
    \hfill
    \begin{subfigure}[b]{0.3\linewidth}
        \centering
        \includegraphics[width=\linewidth]{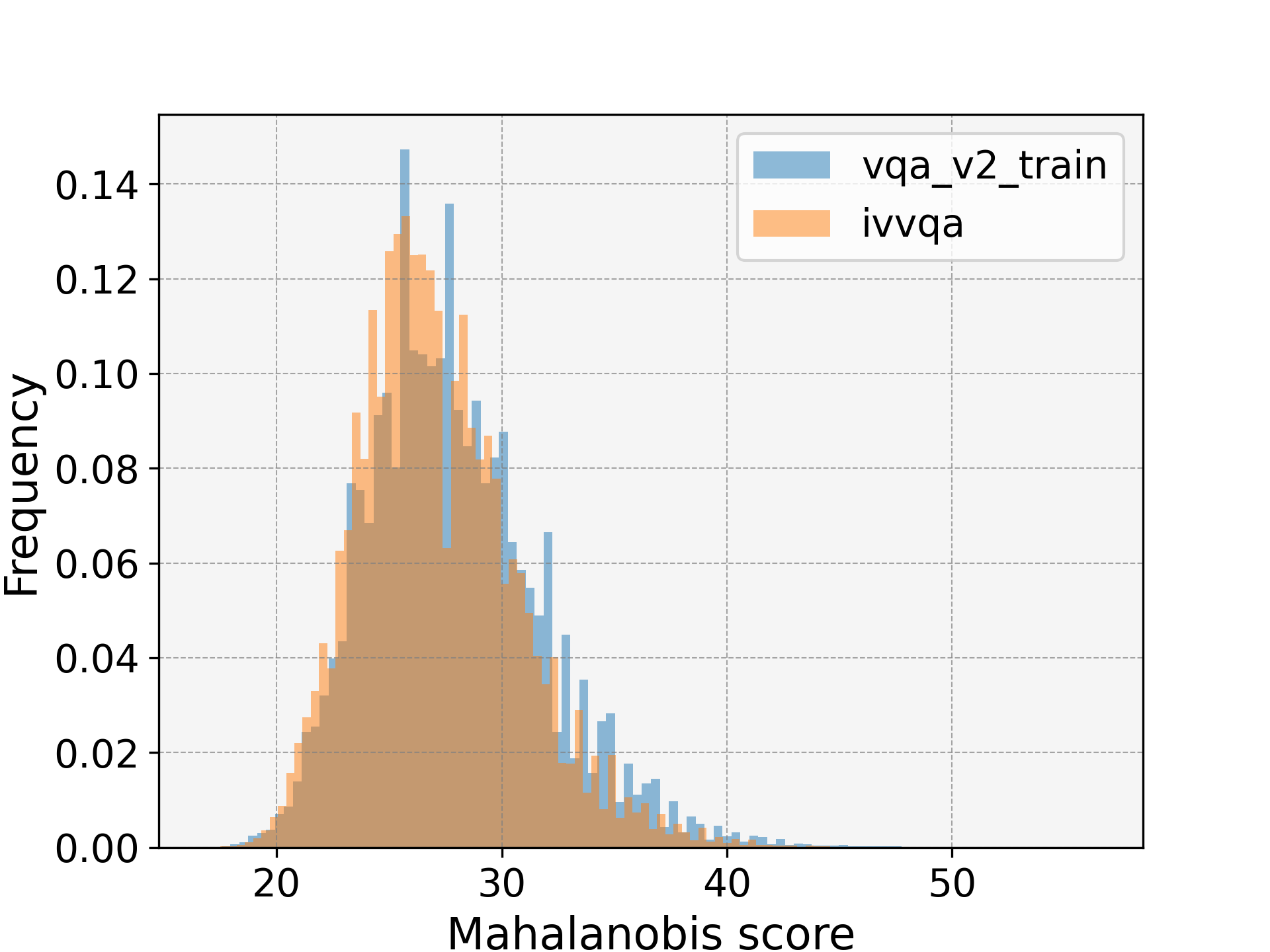}
        \caption{IV VQA}
        \label{fig:ivvqa}
    \end{subfigure}
    \hfill
    \begin{subfigure}[b]{0.3\linewidth}
        \centering
        \includegraphics[width=\linewidth]{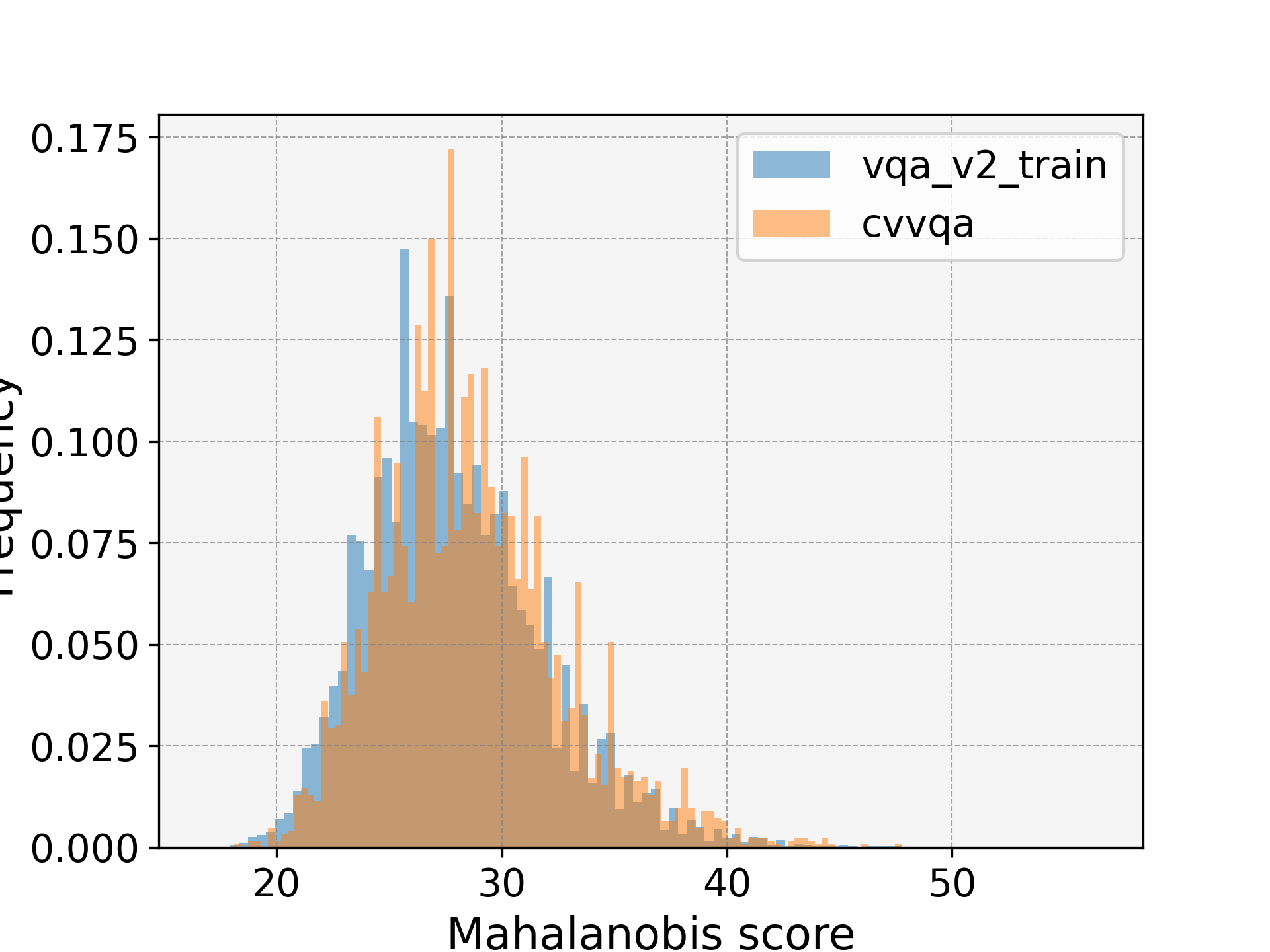}
        \caption{CV VQA}
        \label{fig:vcvvqa}
    \end{subfigure}

    \vspace{0.5cm} %

    \begin{subfigure}[b]{0.3\linewidth}
        \centering
        \includegraphics[width=\linewidth]{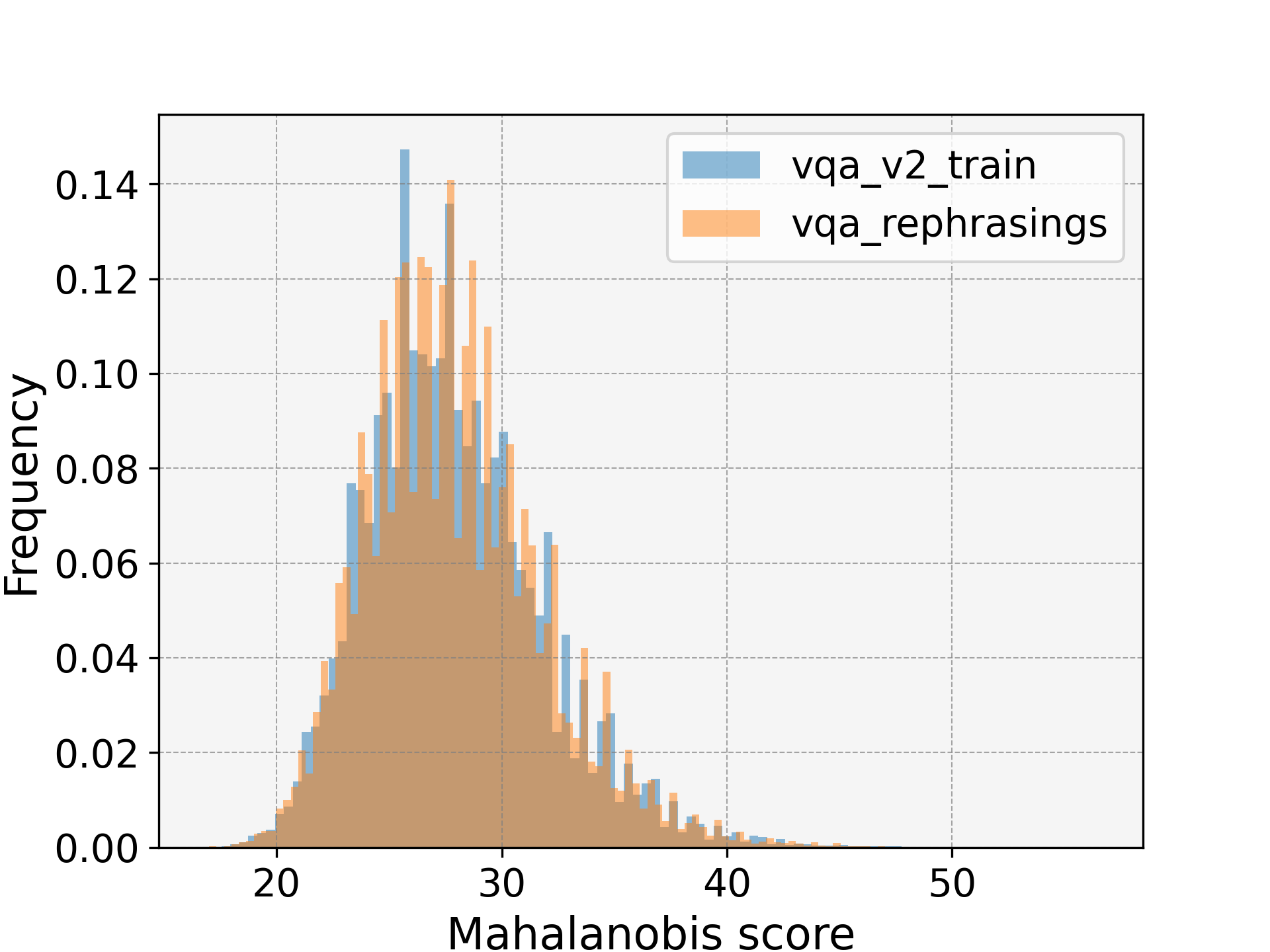}
        \caption{VQA Rephrasings}
        \label{fig:vvqa_rephrasings}
    \end{subfigure}
    \hfill
    \begin{subfigure}[b]{0.3\linewidth}
        \centering
        \includegraphics[width=\linewidth]{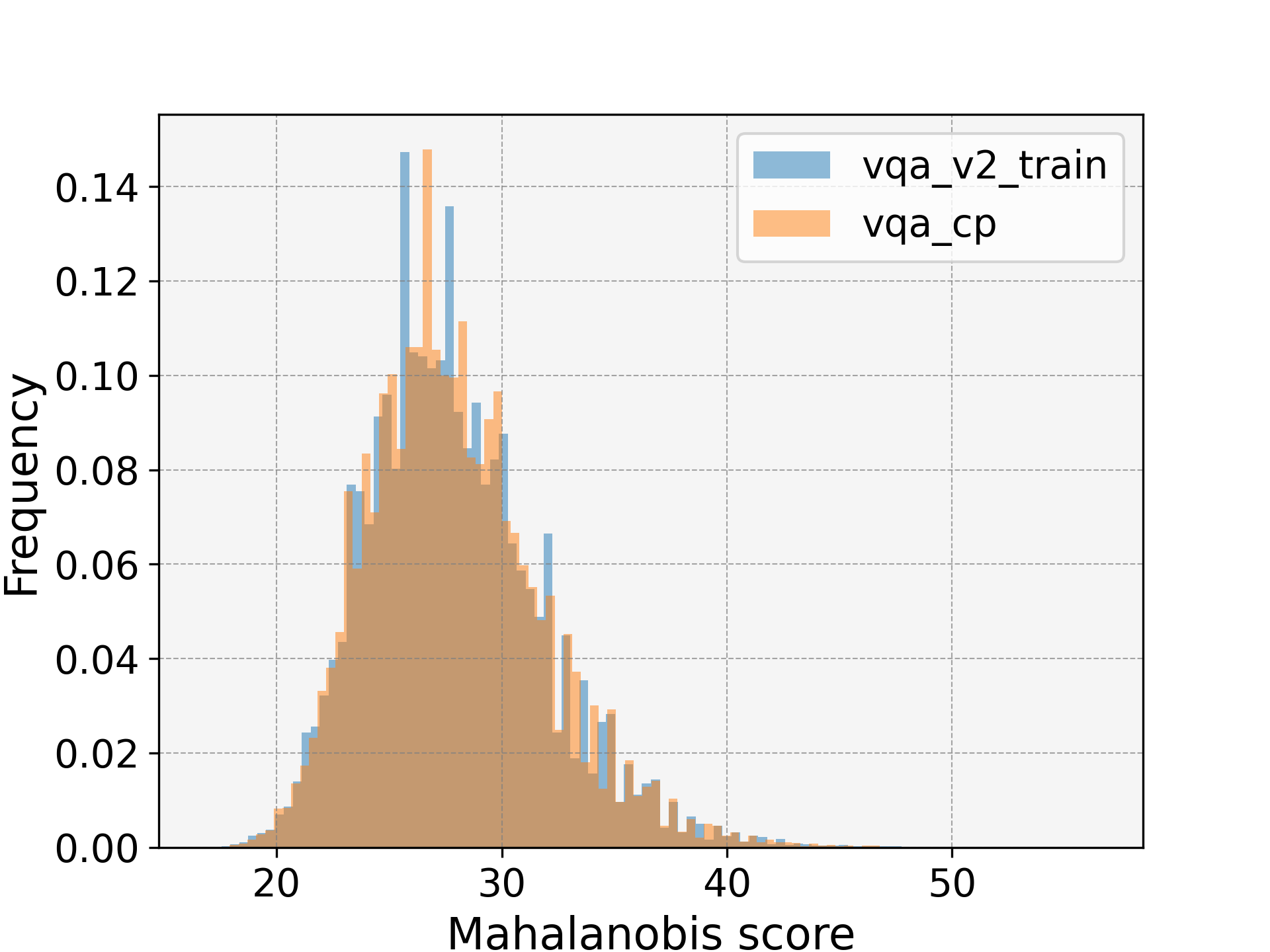}
        \caption{VQA CP v2}
        \label{fig:vvqa_cp_v2}
    \end{subfigure}
    \hfill
    \begin{subfigure}[b]{0.3\linewidth}
        \centering
        \includegraphics[width=\linewidth]{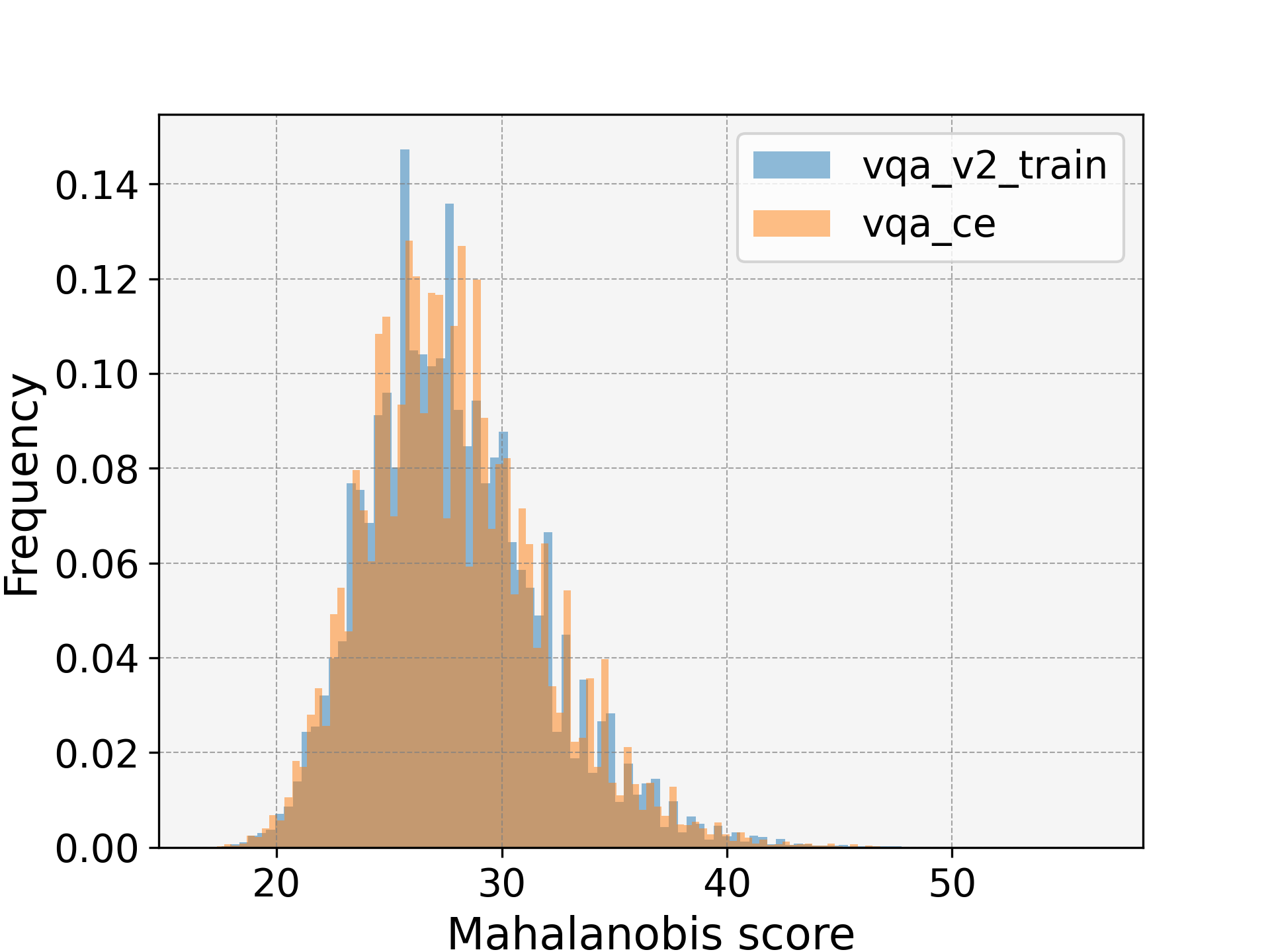}
        \caption{VQA CE}
        \label{fig:vvqa_ce}
    \end{subfigure}

    \vspace{0.5cm} %

    \begin{subfigure}[b]{0.3\linewidth}
        \centering
        \includegraphics[width=\linewidth]{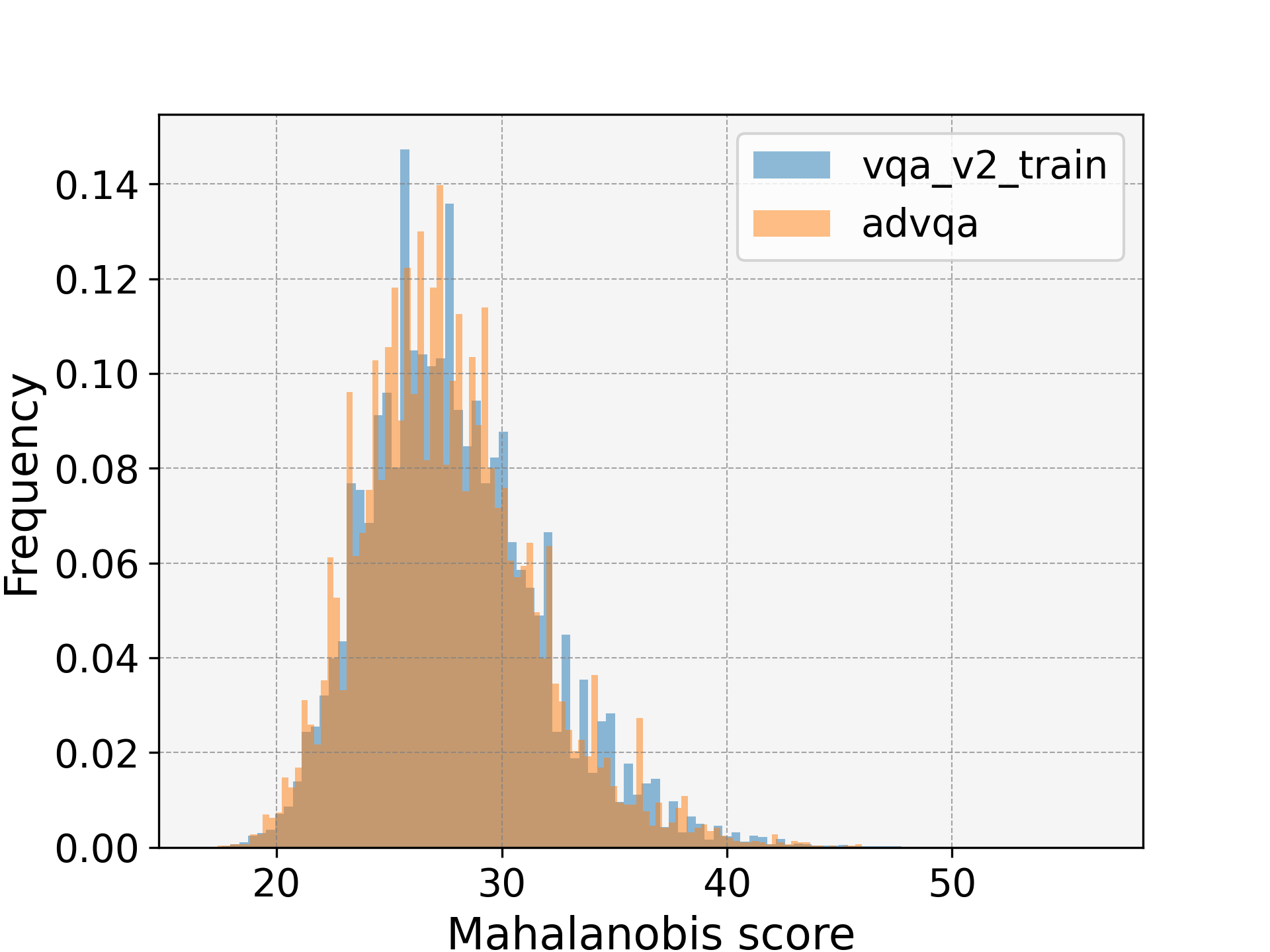}
        \caption{ADVQA}
        \label{fig:vadvqa}
    \end{subfigure}
    \hfill
    \begin{subfigure}[b]{0.3\linewidth}
        \centering
        \includegraphics[width=\linewidth]{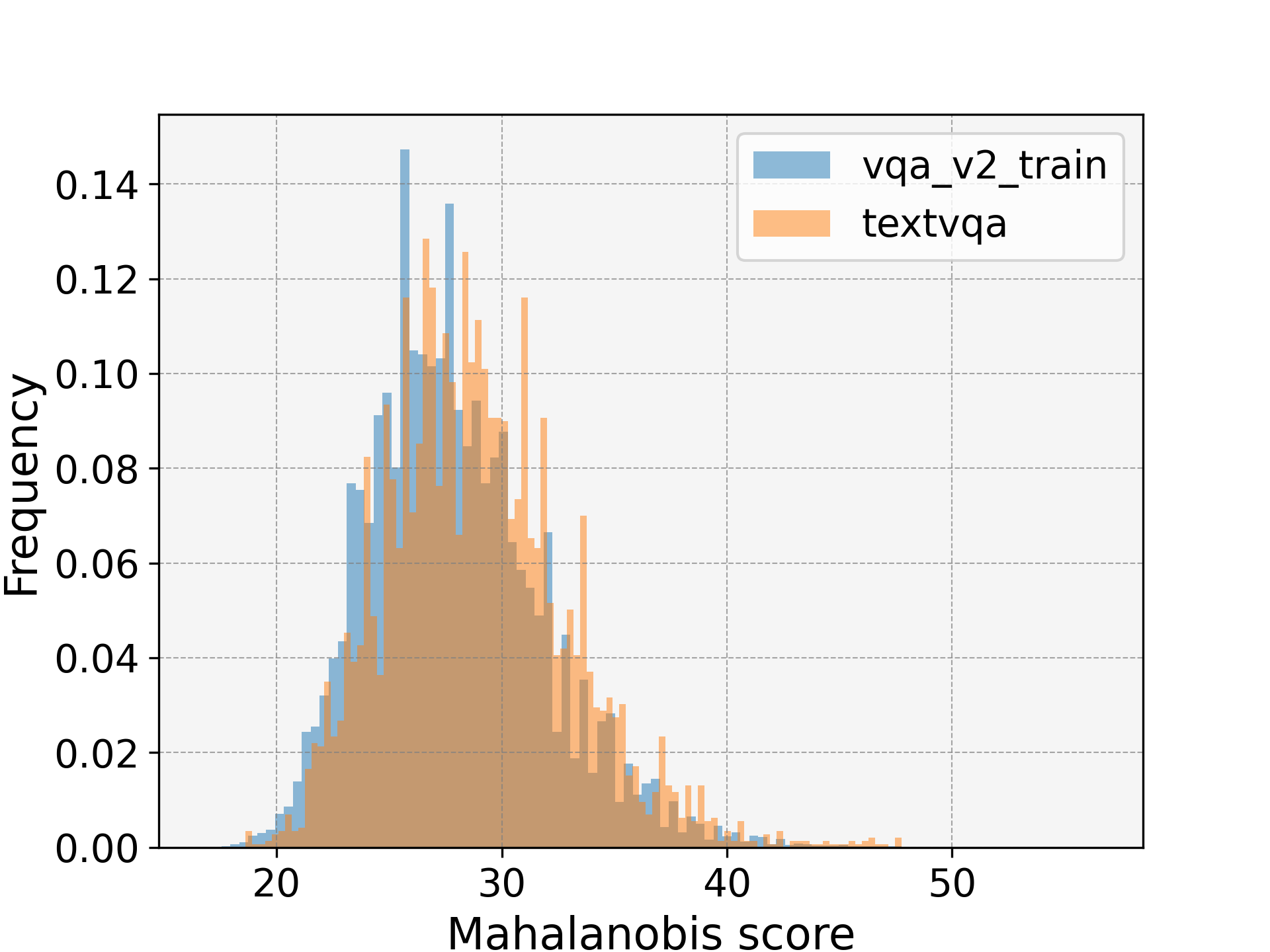}
        \caption{Text VQA}
        \label{fig:vtextvqa}
    \end{subfigure}
    \hfill
    \begin{subfigure}[b]{0.3\linewidth}
        \centering
        \includegraphics[width=\linewidth]{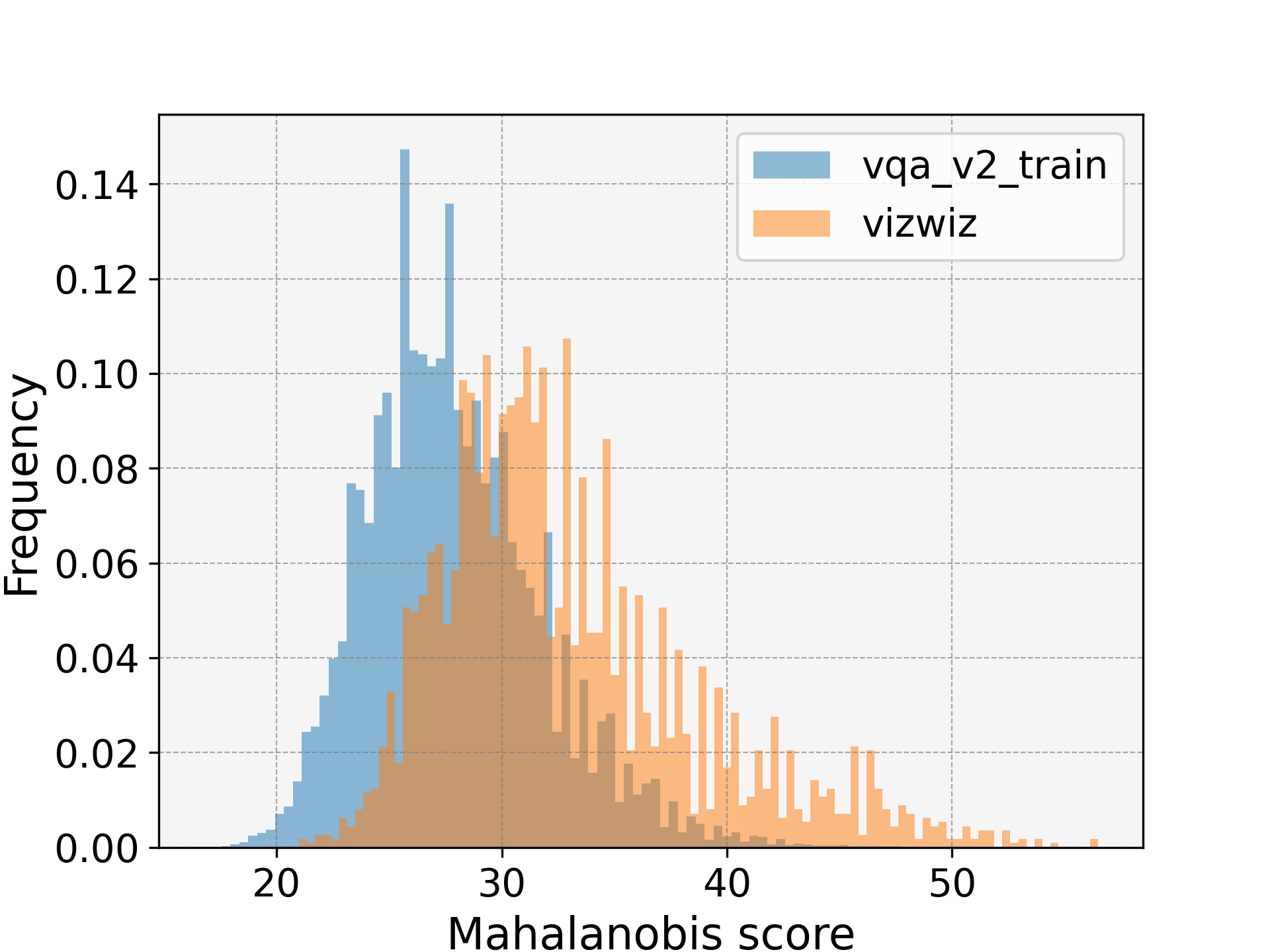}
        \caption{VizWiz}
        \label{fig:vvizwiz}
    \end{subfigure}

    \vspace{0.5cm} %

    \begin{subfigure}[b]{0.3\linewidth}
        \centering
        \includegraphics[width=\linewidth]{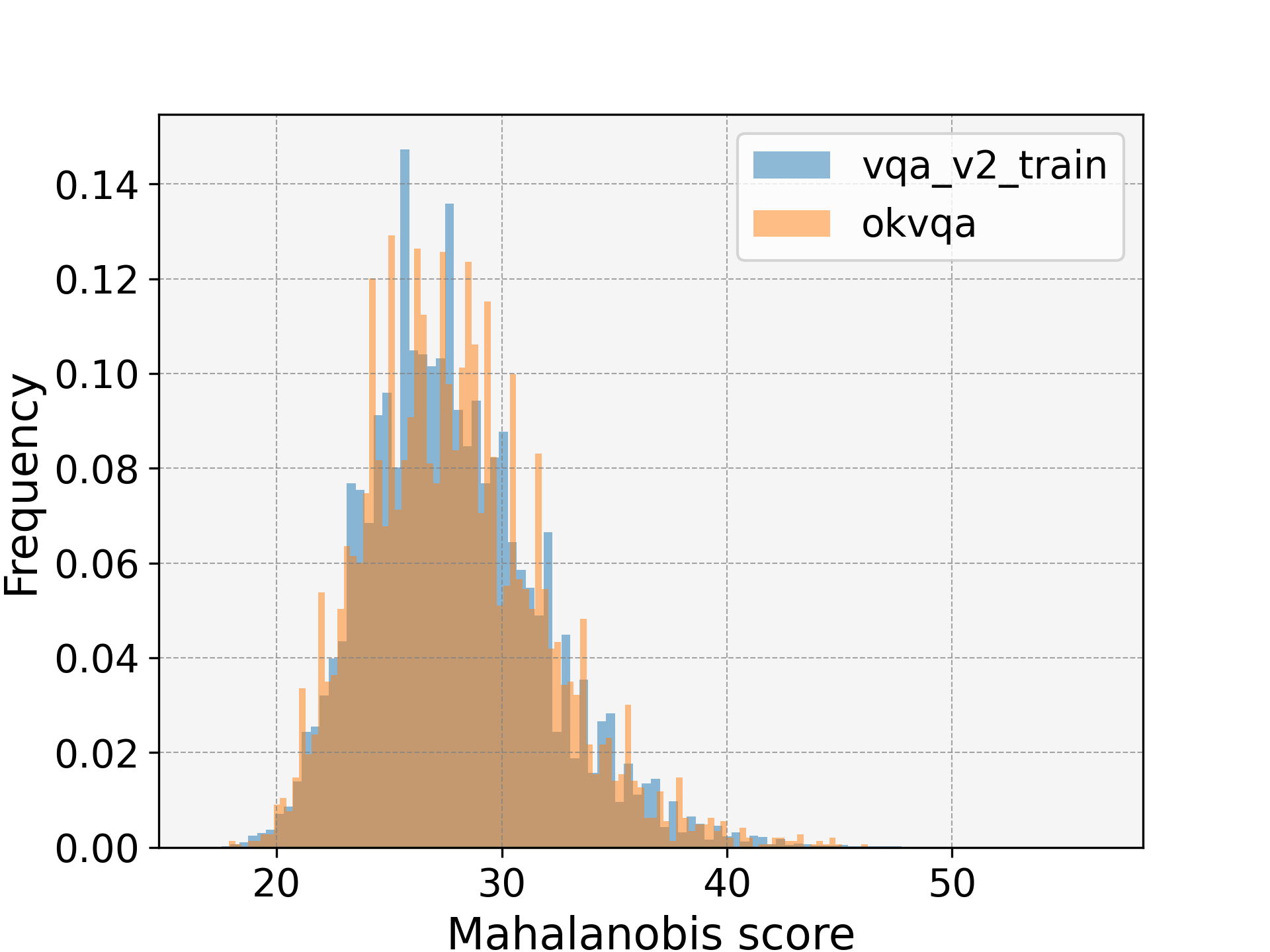}
        \caption{OK VQA}
        \label{fig:vokvqa}
    \end{subfigure}
    \caption{Histogram for Vanilla FT Visual Shifts: We depict the \( S_{\text{Maha}} \) score on the visual modality for each sample in the VQAv2 train split in blue and the corresponding test samples in orange. There's minimal visual shifts for all VQA datasets from the VQAv2 train, except for Figure \subref{fig:vvizwiz} which shows evidence of greater shifts between the orange distribution and the blue distribution. }
    \label{fig:v_histograms}
\end{figure*}

\begin{figure*}[!h]
    \centering

    \begin{subfigure}[b]{0.3\linewidth}
        \centering
        \includegraphics[width=\linewidth]{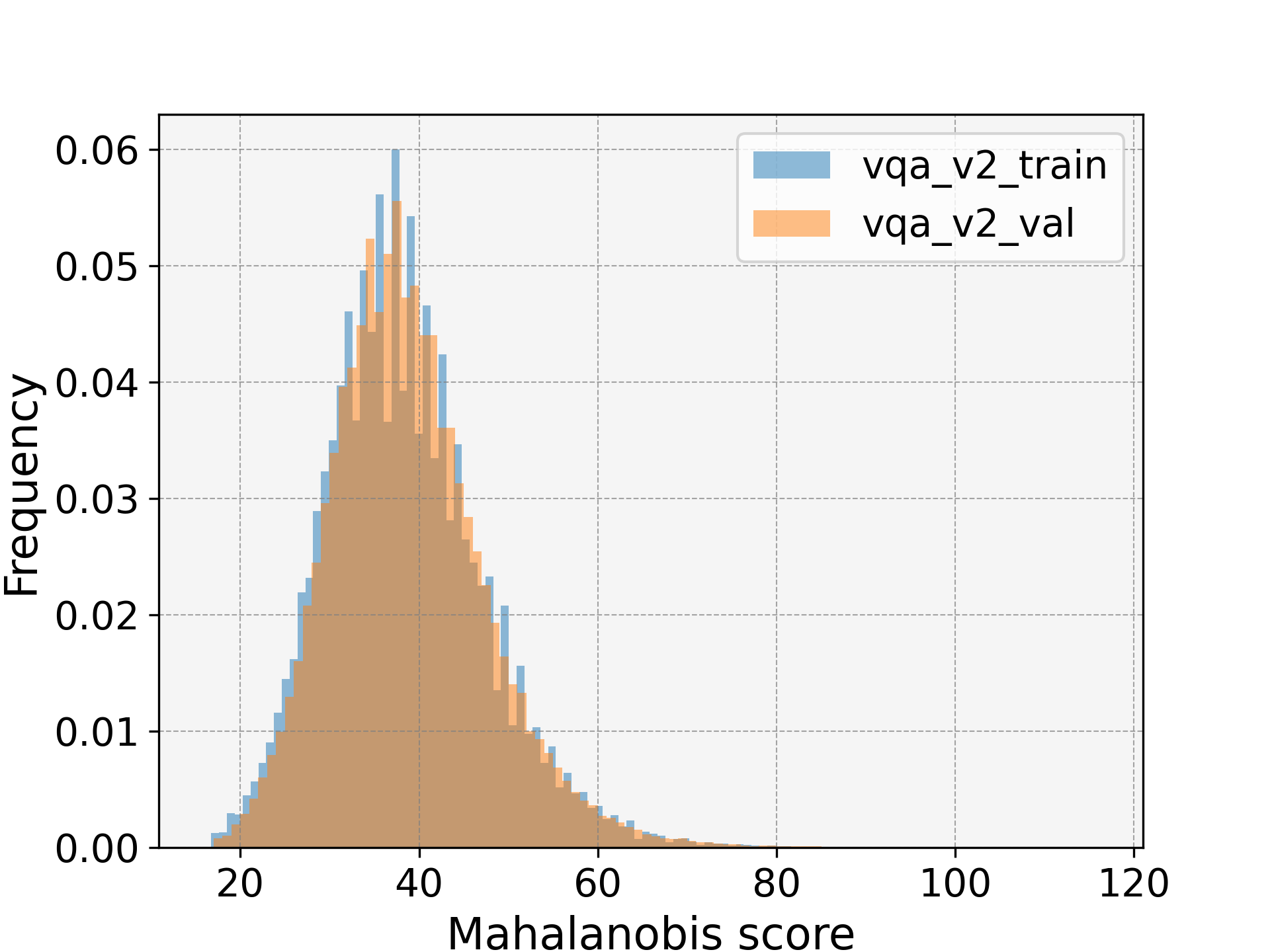}
        \caption{VQAv2 Val}
        \label{fig:vqvqav2_val}
    \end{subfigure}
    \hfill
    \begin{subfigure}[b]{0.3\linewidth}
        \centering
        \includegraphics[width=\linewidth]{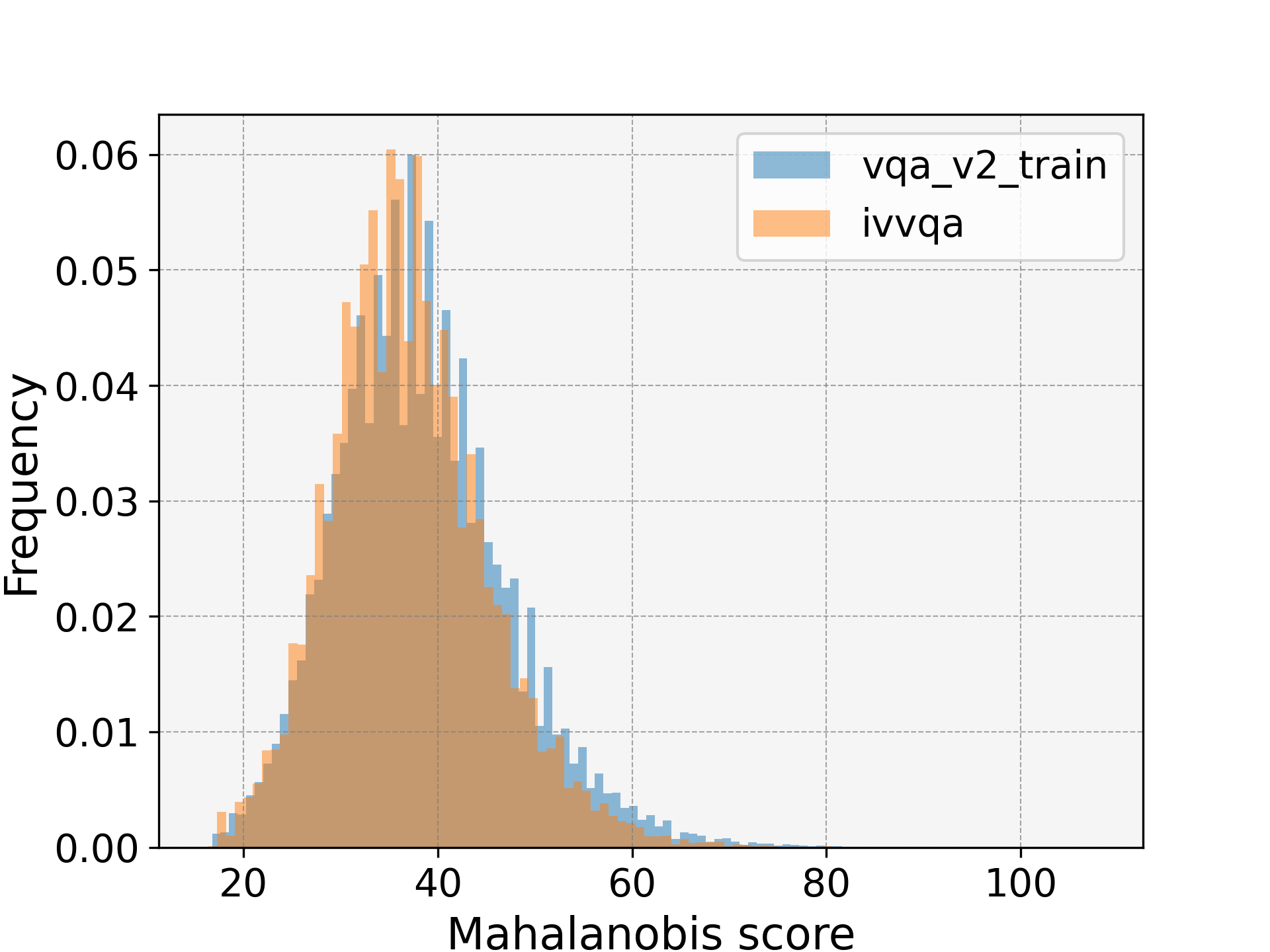}
        \caption{IV VQA}
        \label{fig:vqivvqa}
    \end{subfigure}
    \hfill
    \begin{subfigure}[b]{0.3\linewidth}
        \centering
        \includegraphics[width=\linewidth]{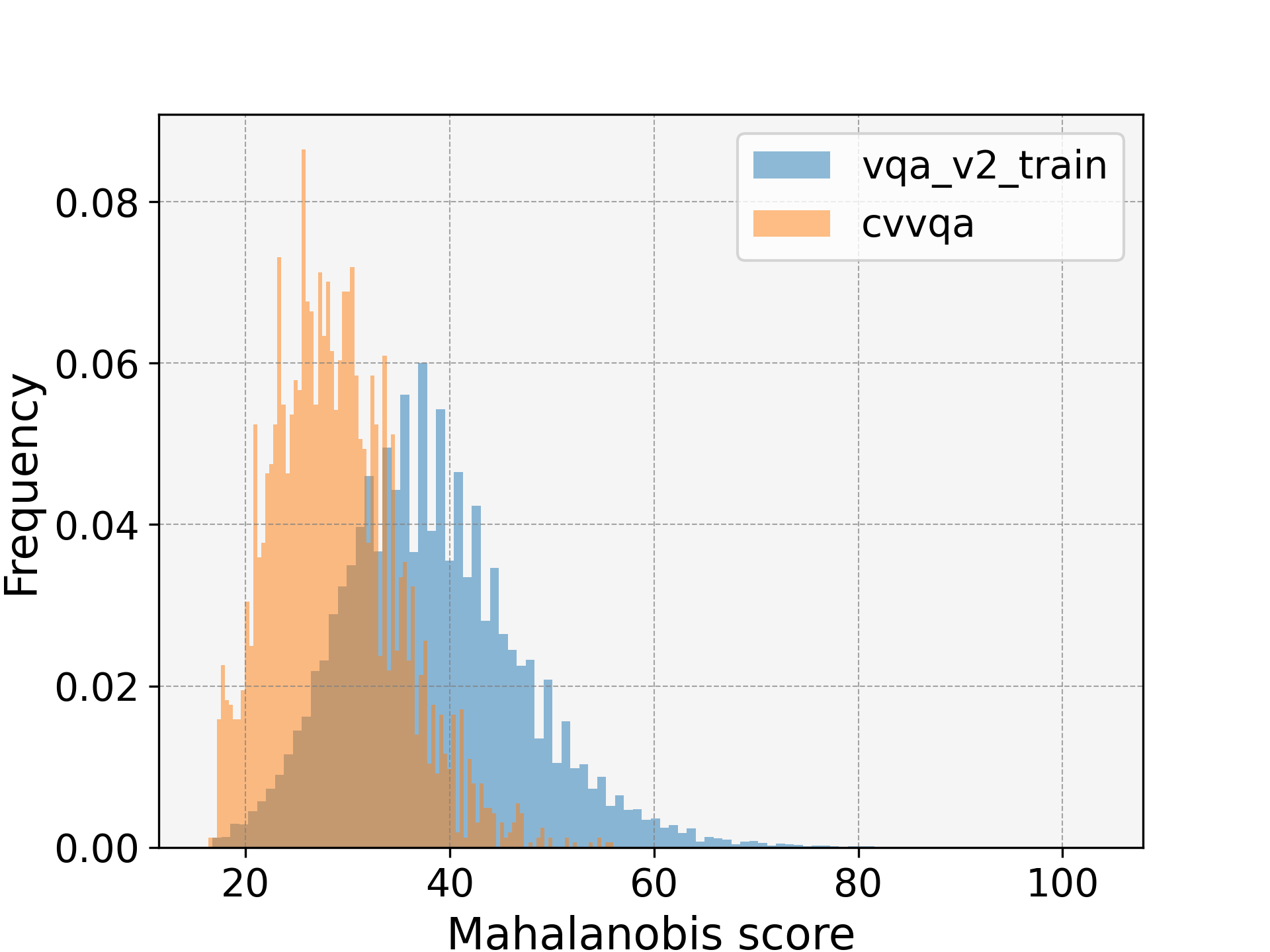}
        \caption{CV VQA}
        \label{fig:vqcvvqa}
    \end{subfigure}

    \vspace{0.5cm} %

    \begin{subfigure}[b]{0.3\linewidth}
        \centering
        \includegraphics[width=\linewidth]{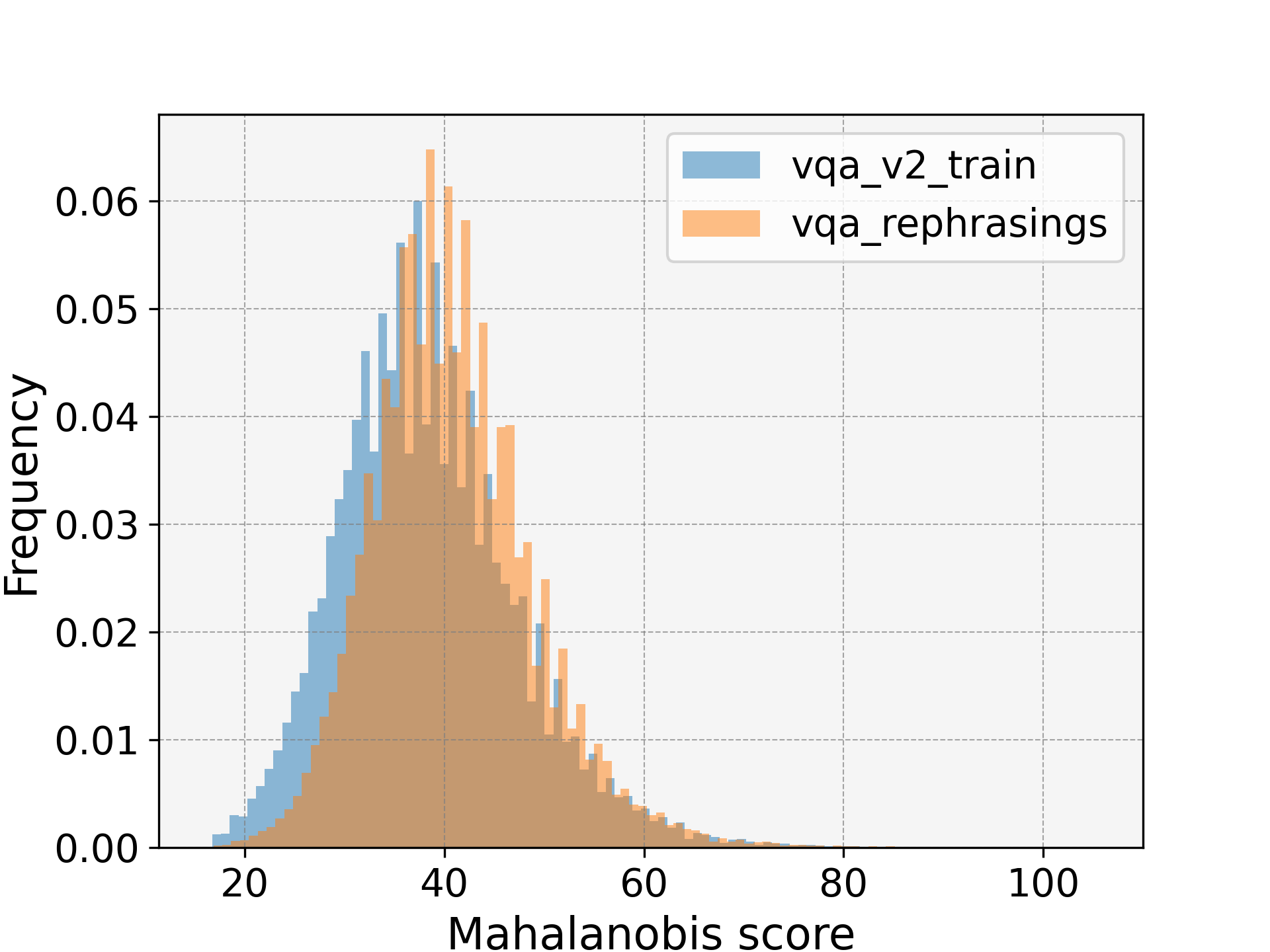}
        \caption{VQA Rephrasings}
        \label{fig:vqvqa_rephrasings}
    \end{subfigure}
    \hfill
    \begin{subfigure}[b]{0.3\linewidth}
        \centering
        \includegraphics[width=\linewidth]{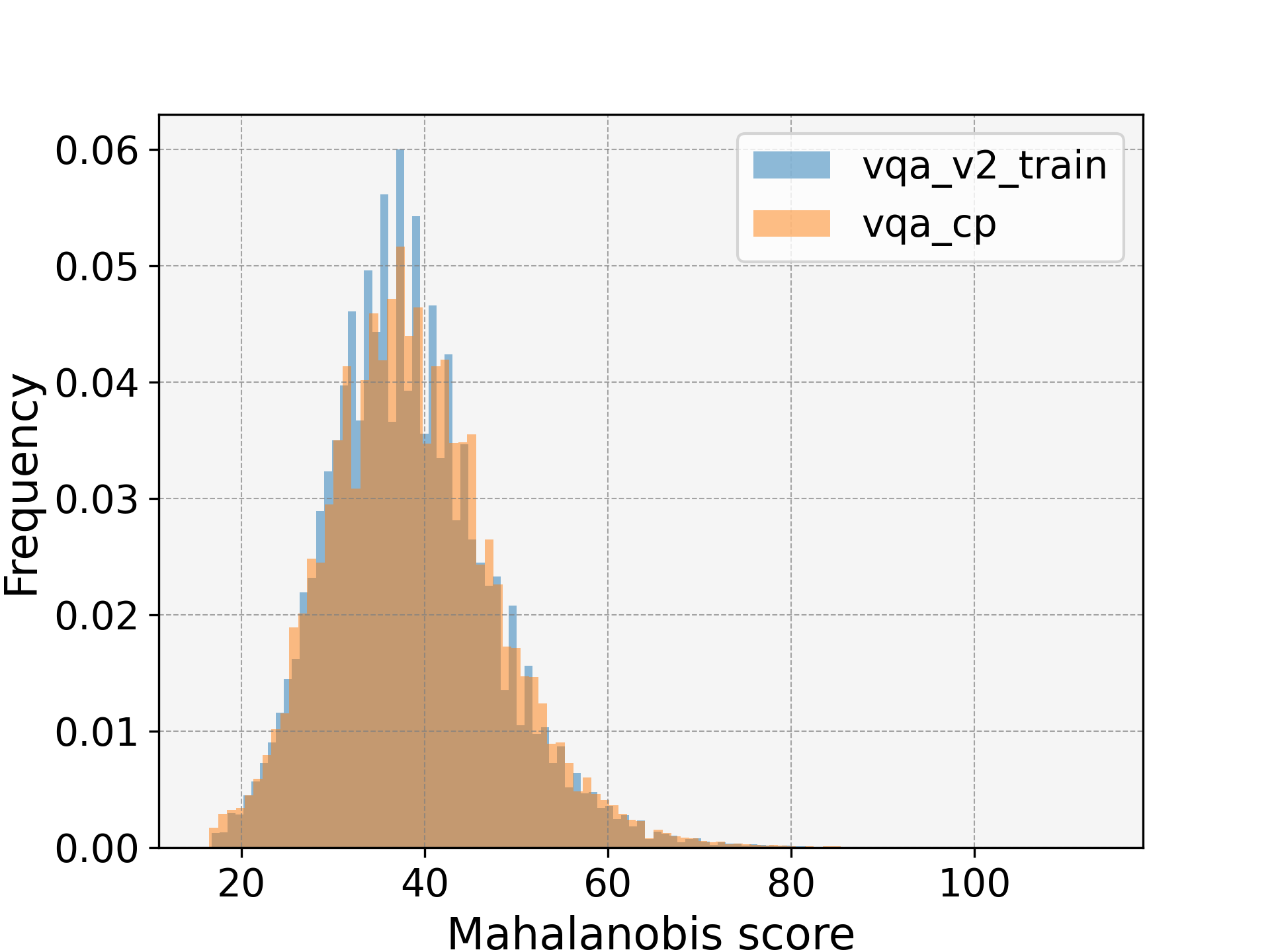}
        \caption{VQA CP v2}
        \label{fig:vqvqa_cp_v2}
    \end{subfigure}
    \hfill
    \begin{subfigure}[b]{0.3\linewidth}
        \centering
        \includegraphics[width=\linewidth]{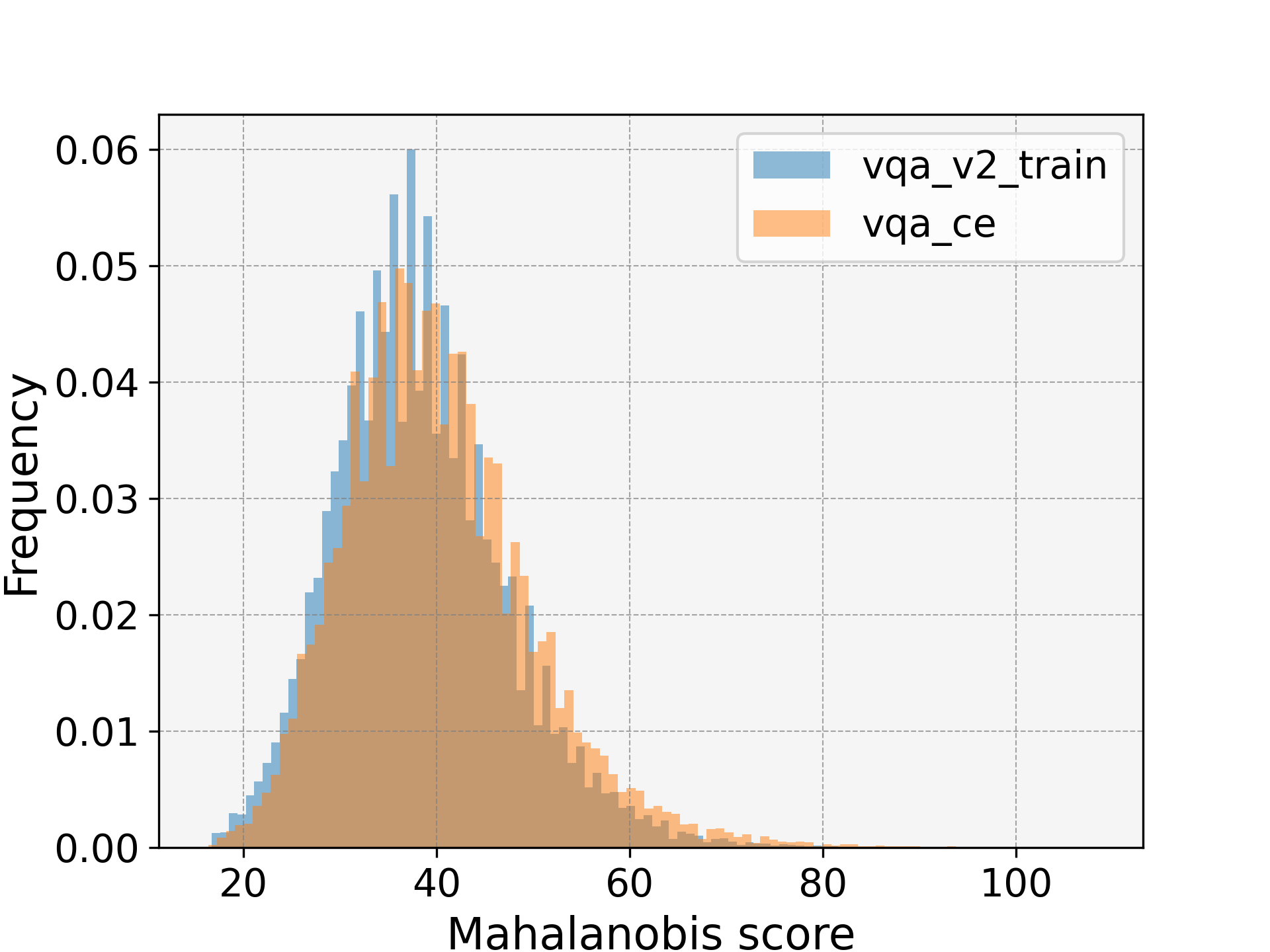}
        \caption{VQA CE}
        \label{fig:vqvqa_ce}
    \end{subfigure}

    \vspace{0.5cm} %

    \begin{subfigure}[b]{0.3\linewidth}
        \centering
        \includegraphics[width=\linewidth]{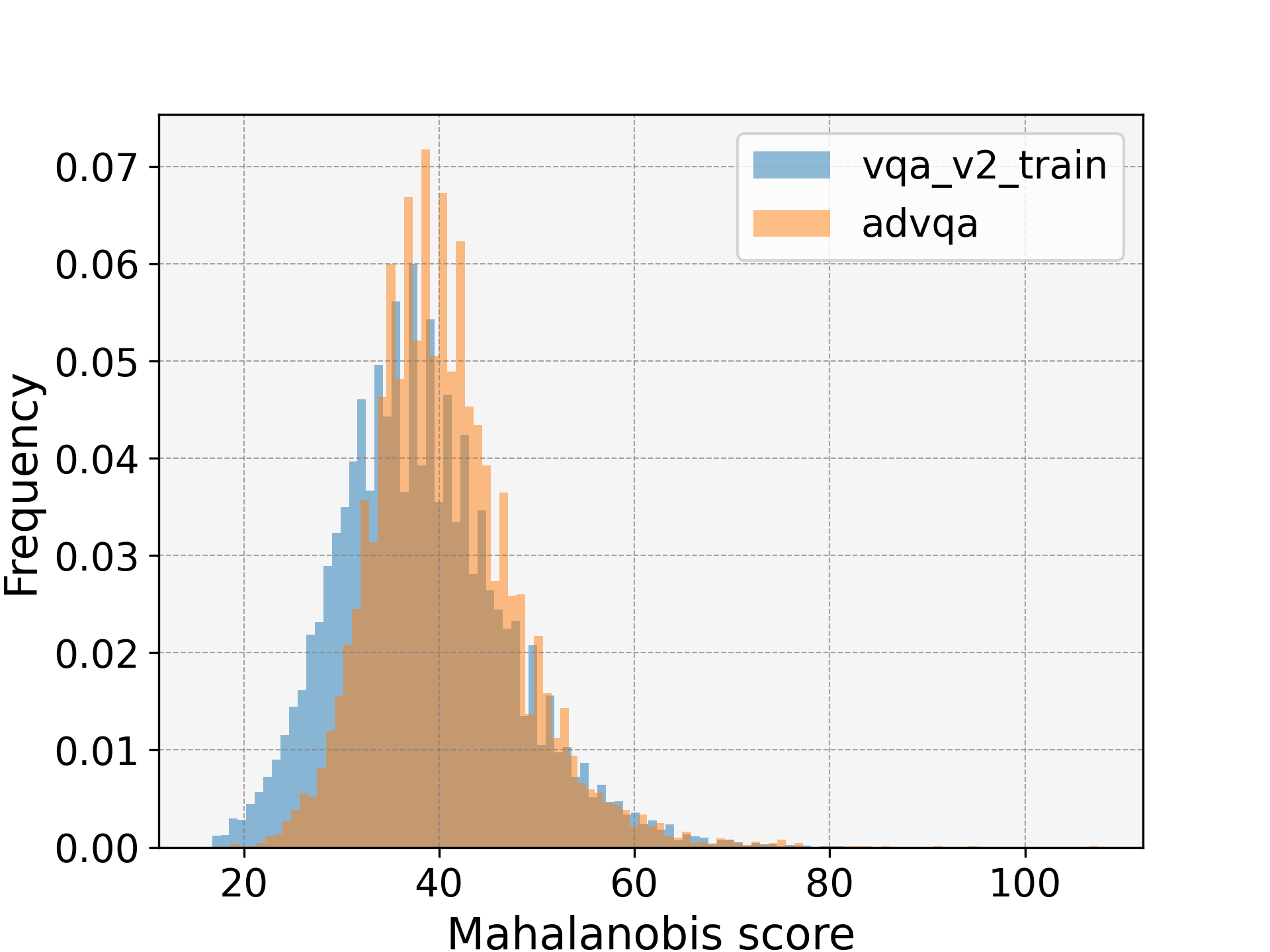}
        \caption{ADVQA}
        \label{fig:vqadvqa}
    \end{subfigure}
    \hfill
    \begin{subfigure}[b]{0.3\linewidth}
        \centering
        \includegraphics[width=\linewidth]{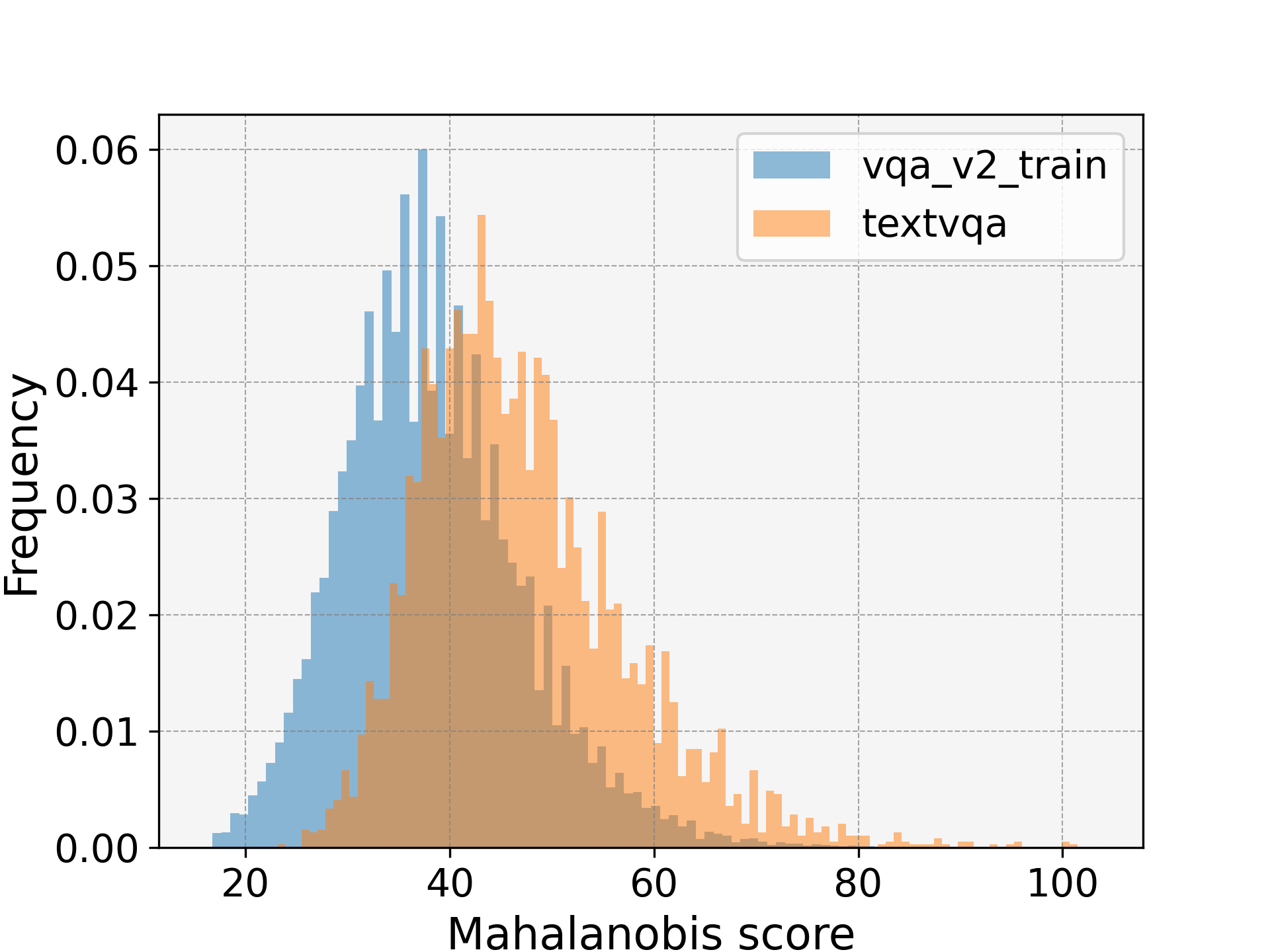}
        \caption{Text VQA}
        \label{fig:vqtextvqa}
    \end{subfigure}
    \hfill
    \begin{subfigure}[b]{0.3\linewidth}
        \centering
        \includegraphics[width=\linewidth]{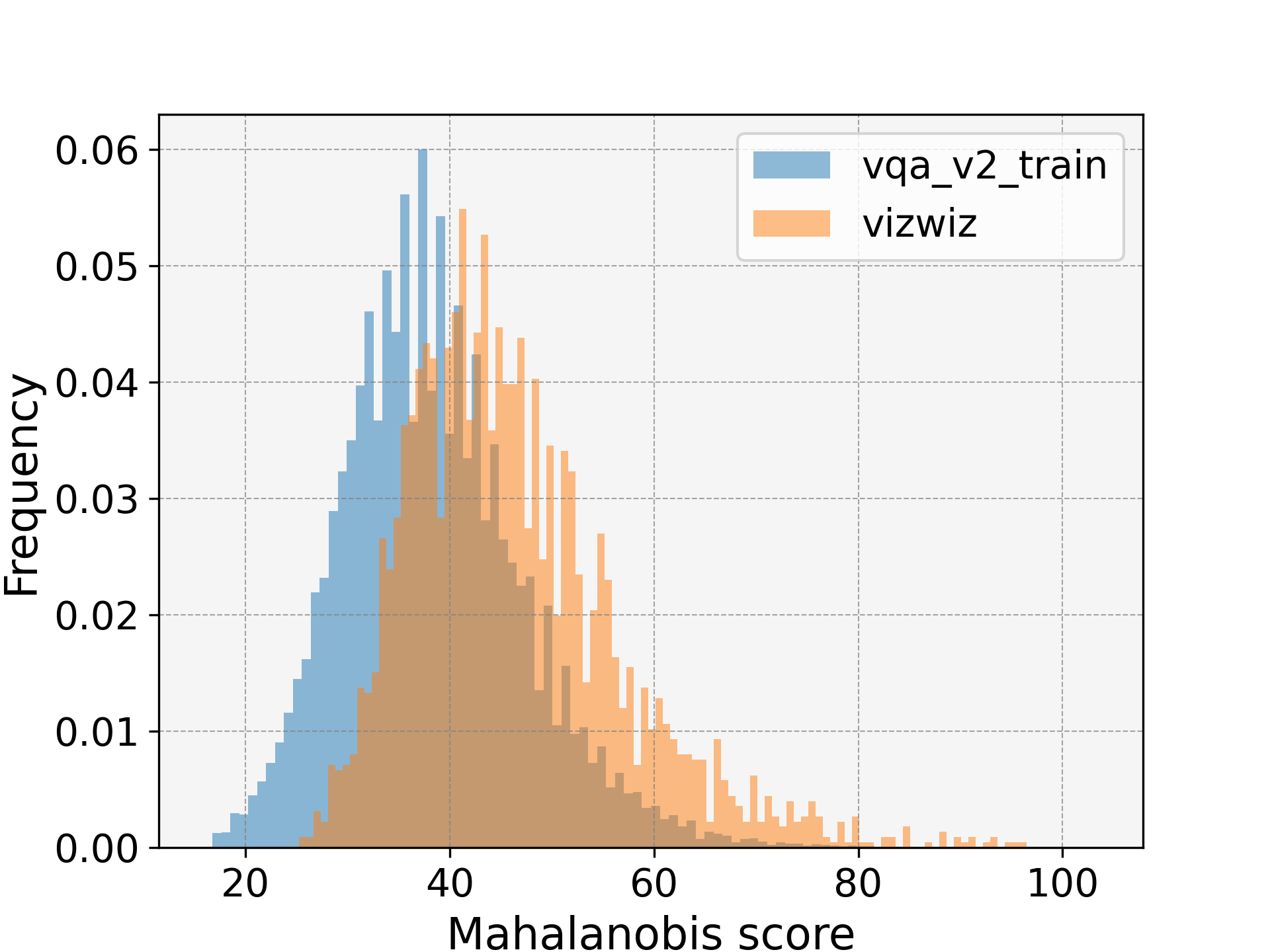}
        \caption{VizWiz}
        \label{fig:vqvizwiz}
    \end{subfigure}

    \vspace{0.5cm} %

    \begin{subfigure}[b]{0.3\linewidth}
        \centering
        \includegraphics[width=\linewidth]{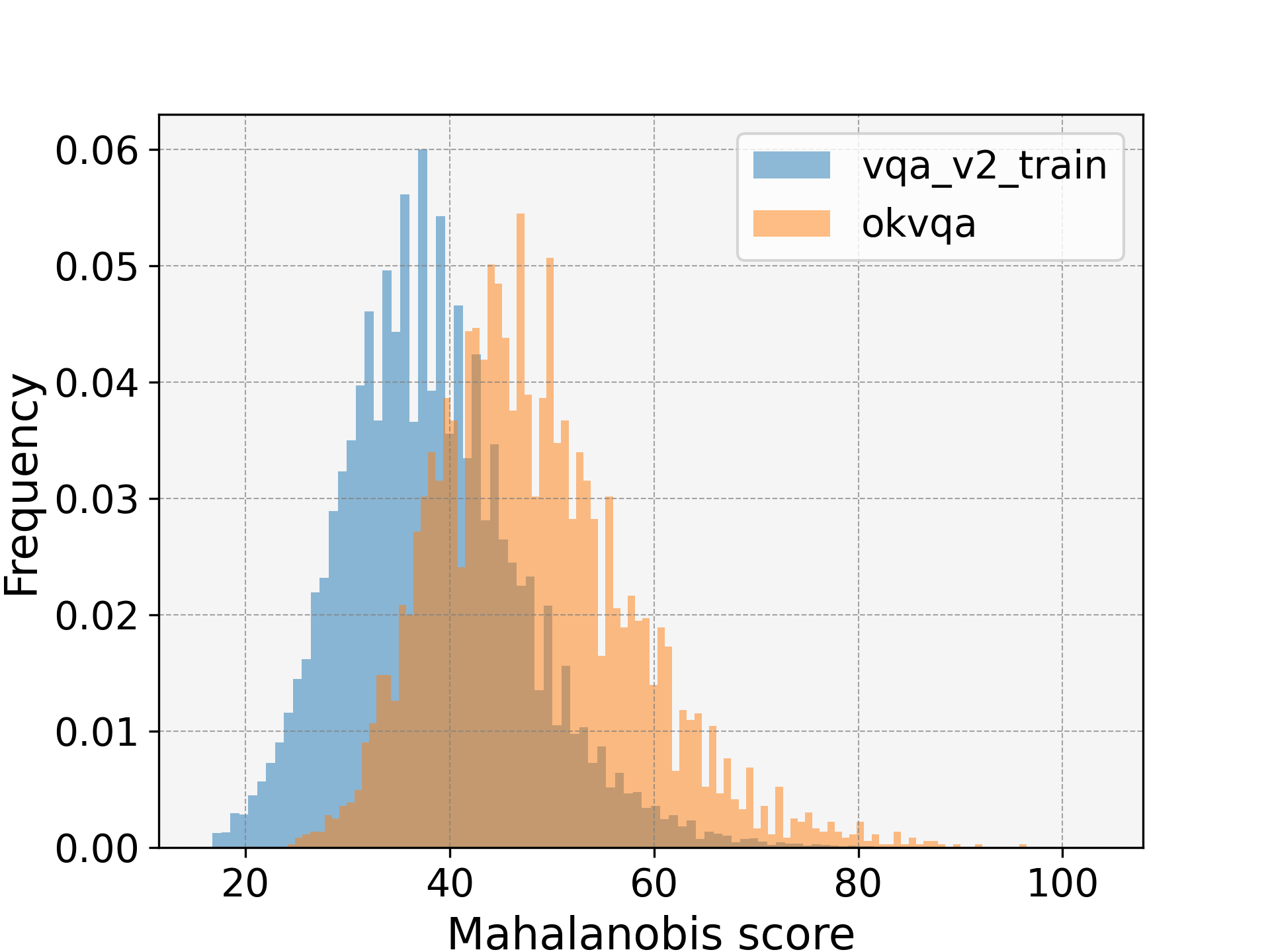}
        \caption{OK VQA}
        \label{fig:vqokvqa}
    \end{subfigure}

    \caption{Histogram for Vanilla FT Question Shifts: We depict the \( S_{\text{Maha}} \) score on the question modality for each sample in the VQAv2 train split in blue and the corresponding test samples in orange. Similar to the visual shift histograms, far OODs (Figures \subref{fig:vqtextvqa}, \subref{fig:vqvizwiz}, \subref{fig:vqokvqa}) also show evidence of greater shifts between the orange distribution and the blue distribution than near OODs.}
    \label{fig:vq_histograms}
\end{figure*}

\begin{figure*}[!h]
    \centering

    \begin{subfigure}[b]{0.3\linewidth}
        \centering
        \includegraphics[width=\linewidth]{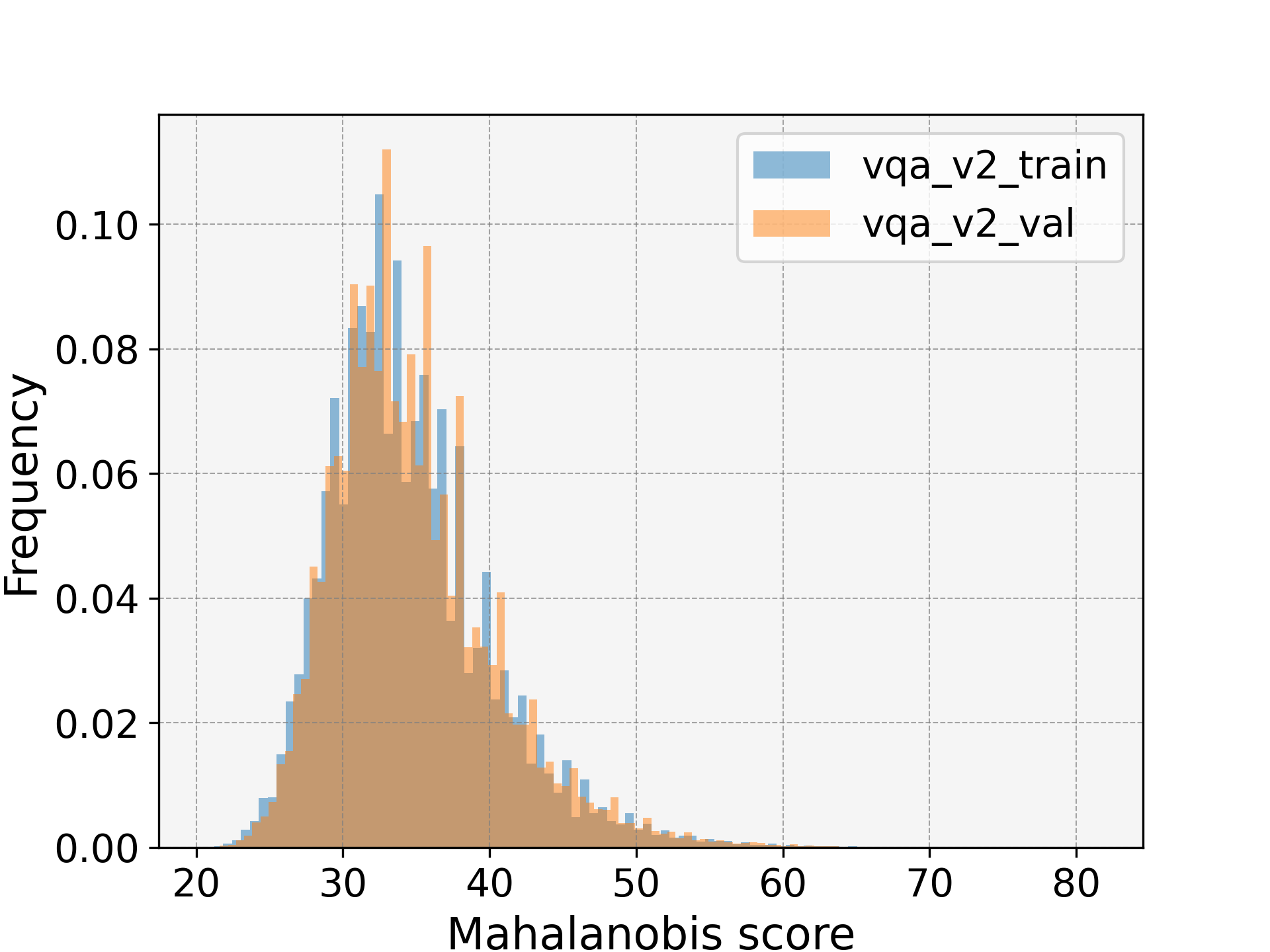}
        \caption{VQAv2 Val}
        \label{fig:vqaVQAv2_val}
    \end{subfigure}
    \hfill
    \begin{subfigure}[b]{0.3\linewidth}
        \centering
        \includegraphics[width=\linewidth]{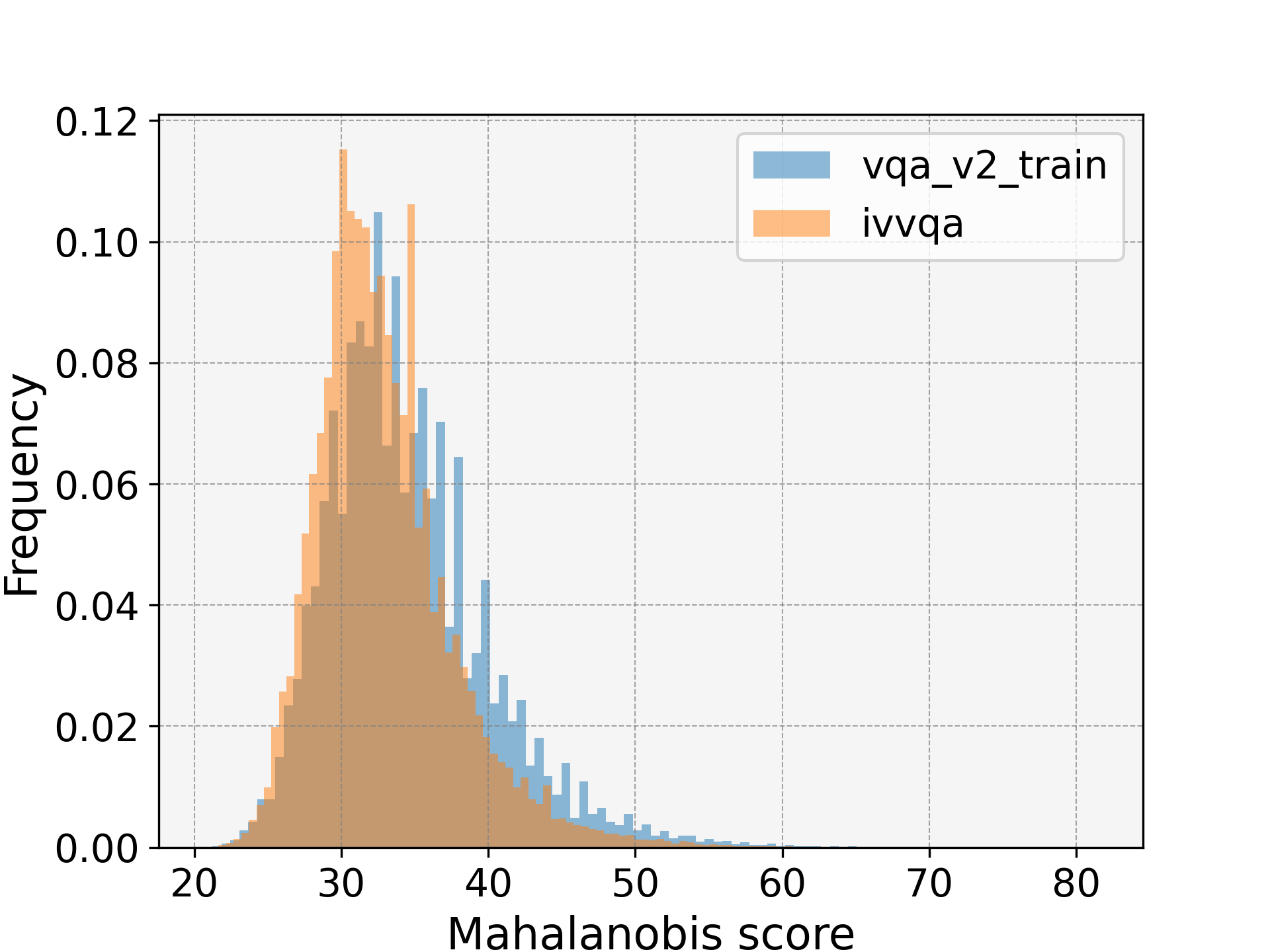}
        \caption{IV VQA}
        \label{fig:vqaivvqa}
    \end{subfigure}
    \hfill
    \begin{subfigure}[b]{0.3\linewidth}
        \centering
        \includegraphics[width=\linewidth]{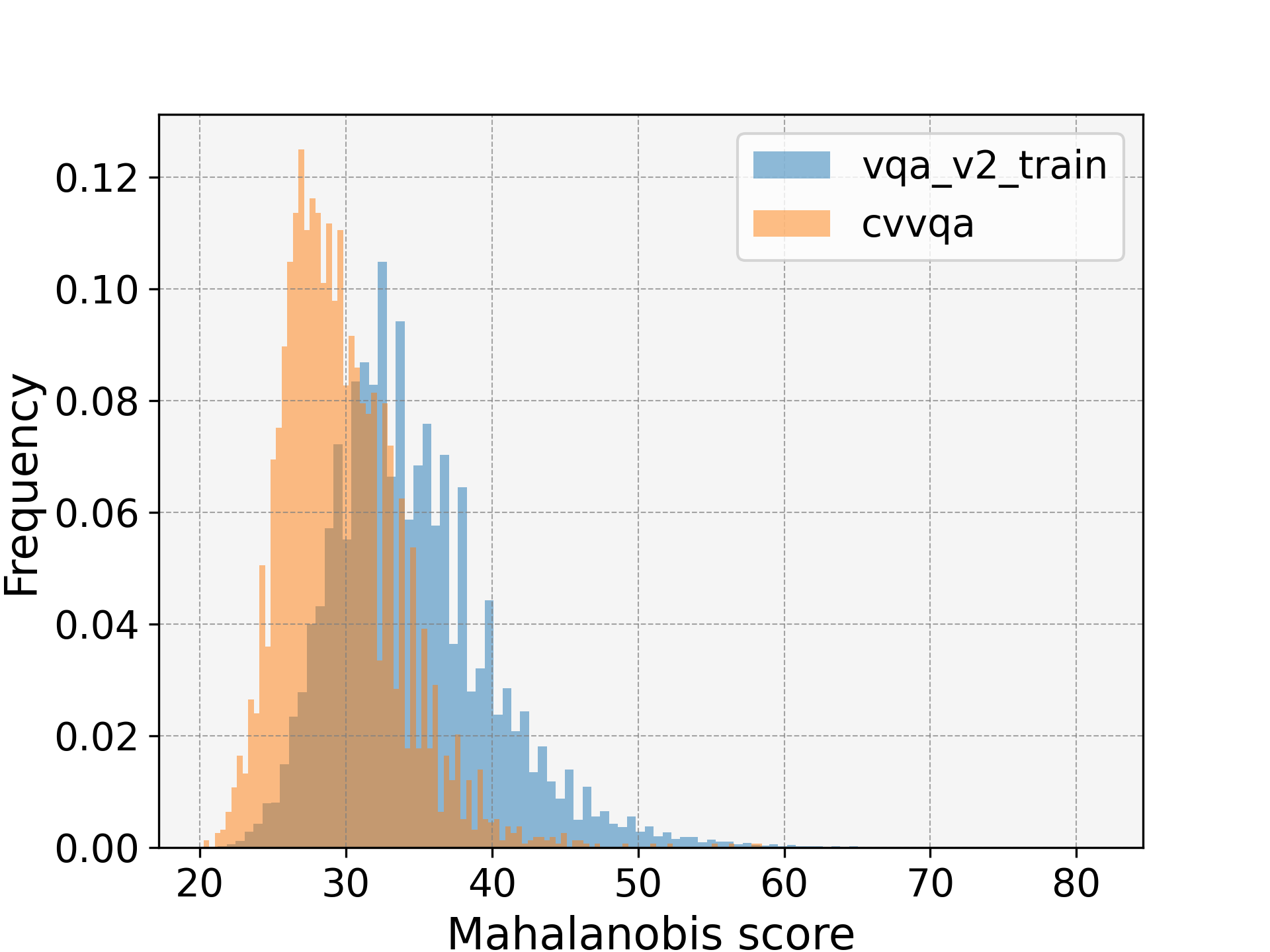}
        \caption{CV VQA}
        \label{fig:vqacvvqa}
    \end{subfigure}

    \vspace{0.5cm} %

    \begin{subfigure}[b]{0.3\linewidth}
        \centering
        \includegraphics[width=\linewidth]{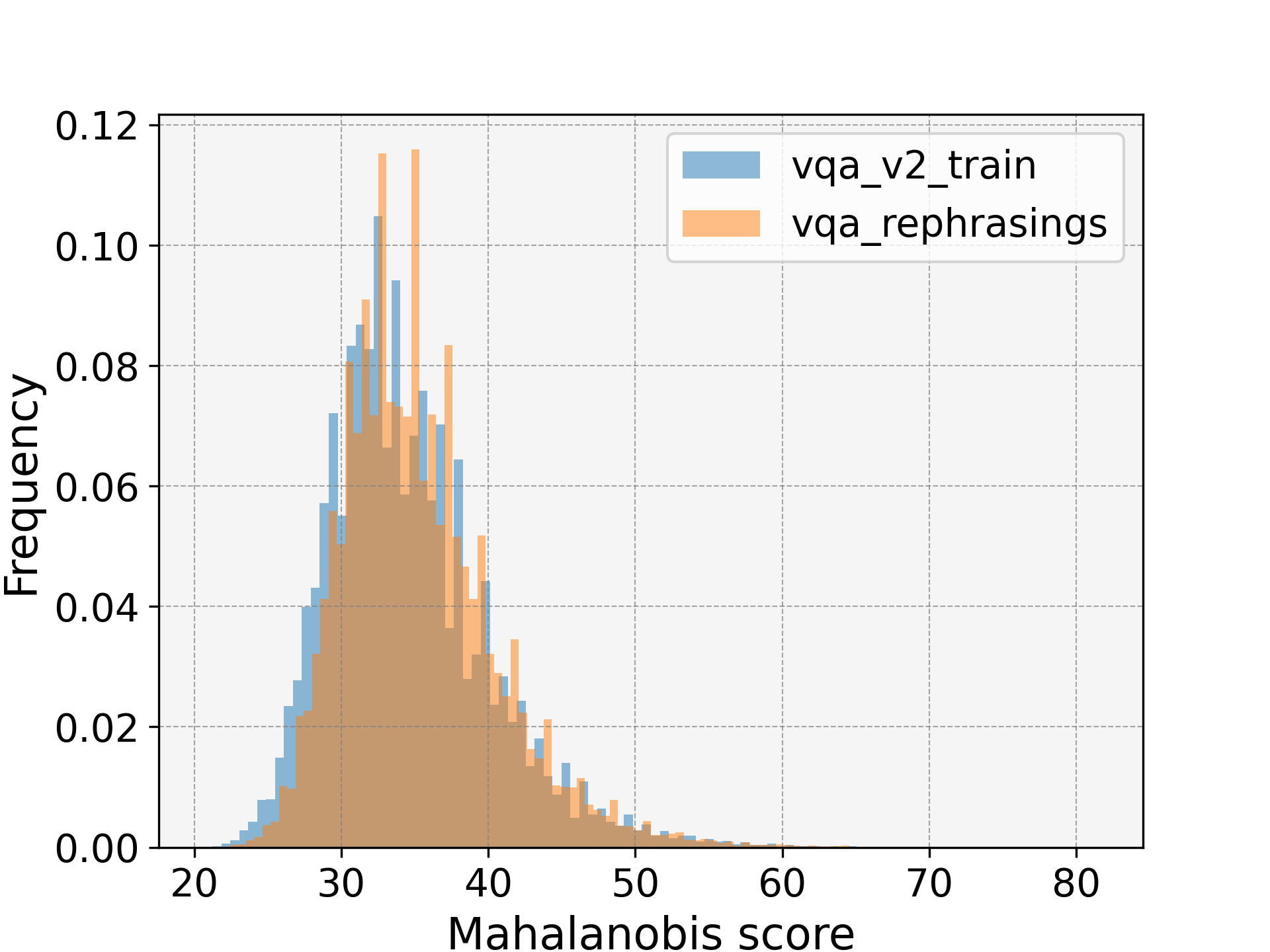}
        \caption{VQA Rephrasings}
        \label{fig:vqavqa_rephrasings}
    \end{subfigure}
    \hfill
    \begin{subfigure}[b]{0.3\linewidth}
        \centering
        \includegraphics[width=\linewidth]{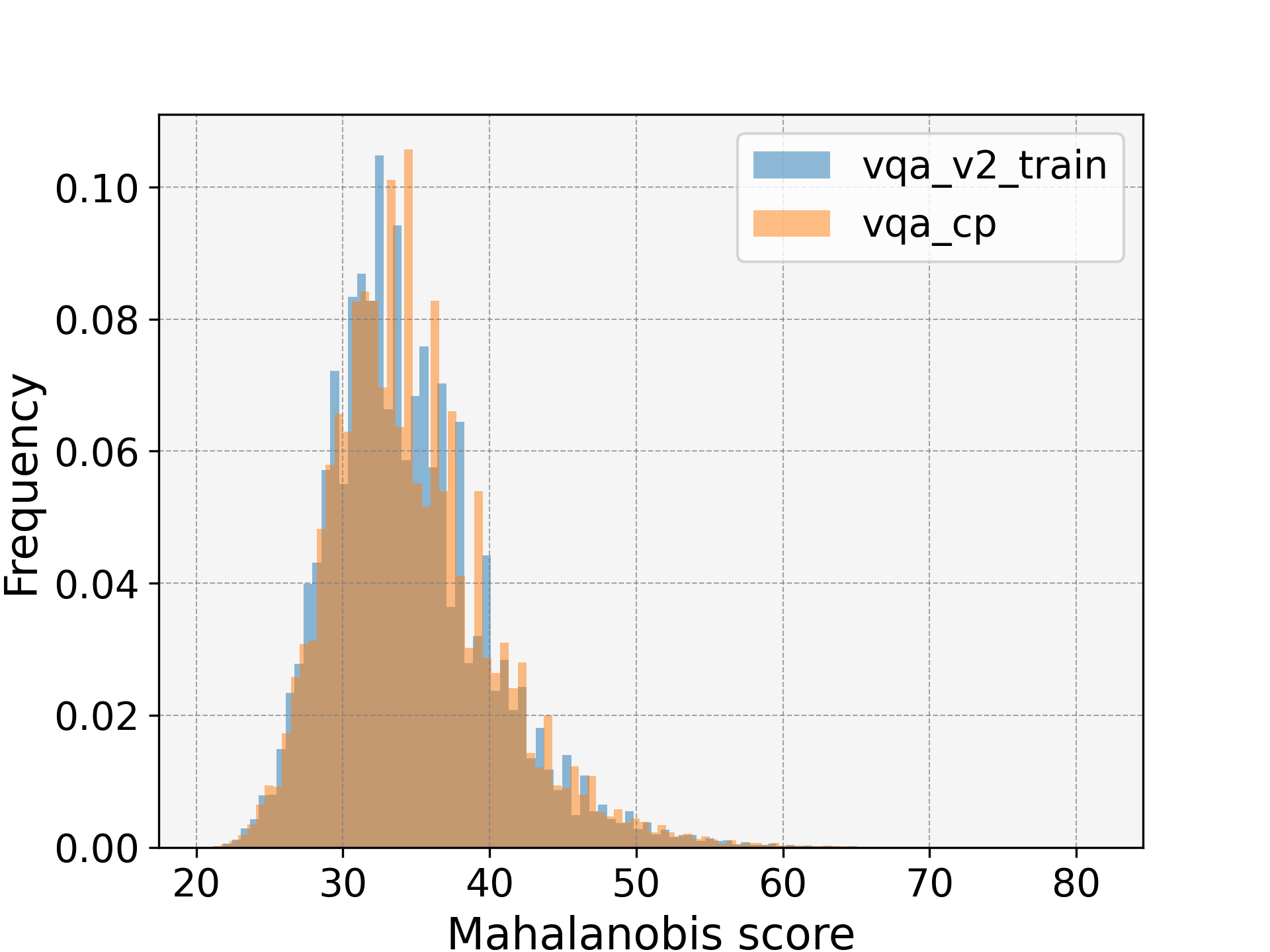}
        \caption{VQA CP v2}
        \label{fig:vqavqa_cp_v2}
    \end{subfigure}
    \hfill
    \begin{subfigure}[b]{0.3\linewidth}
        \centering
        \includegraphics[width=\linewidth]{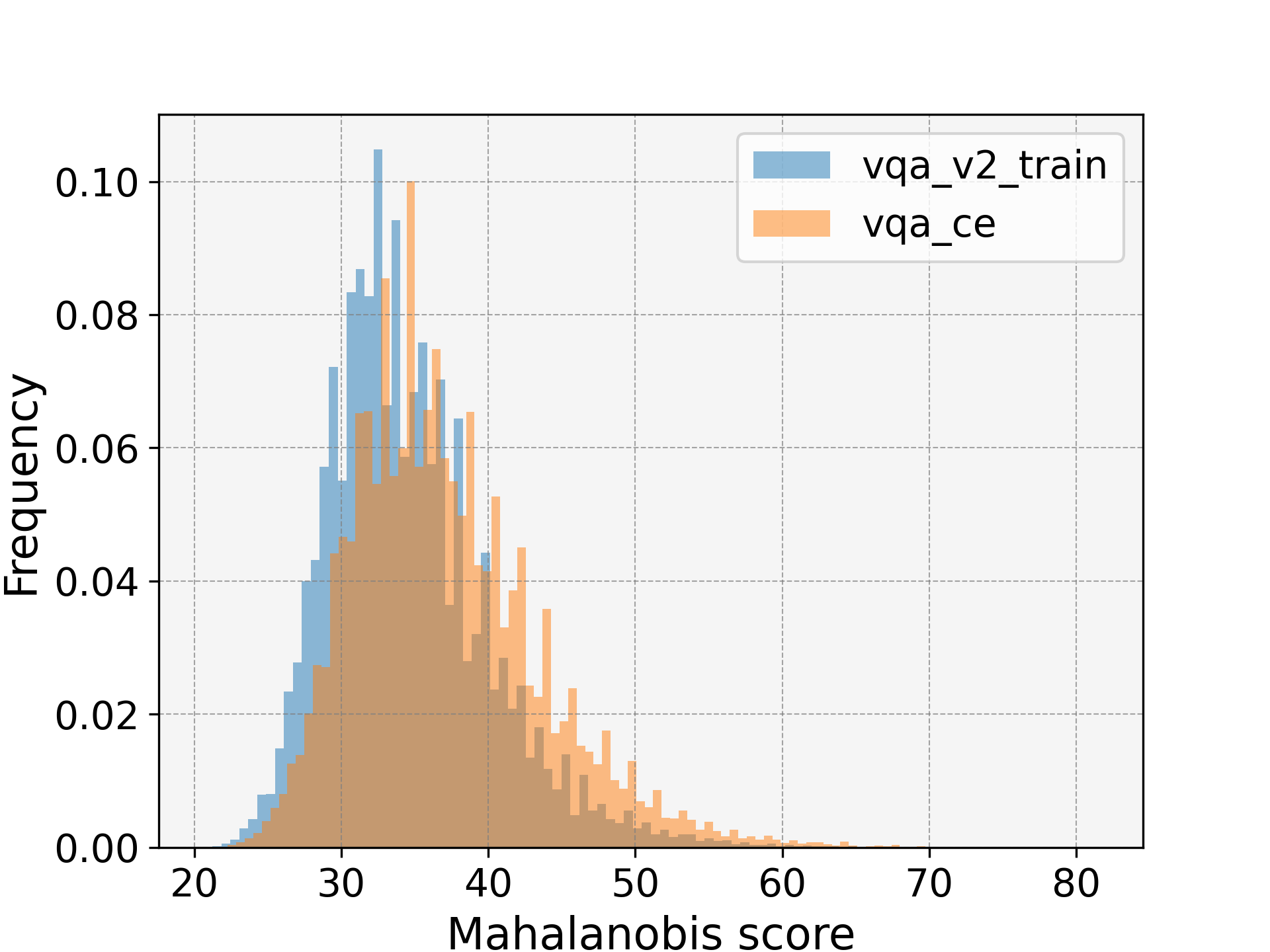}
        \caption{VQA CE}
        \label{fig:vqavqa_ce}
    \end{subfigure}

    \vspace{0.5cm} %

    \begin{subfigure}[b]{0.3\linewidth}
        \centering
        \includegraphics[width=\linewidth]{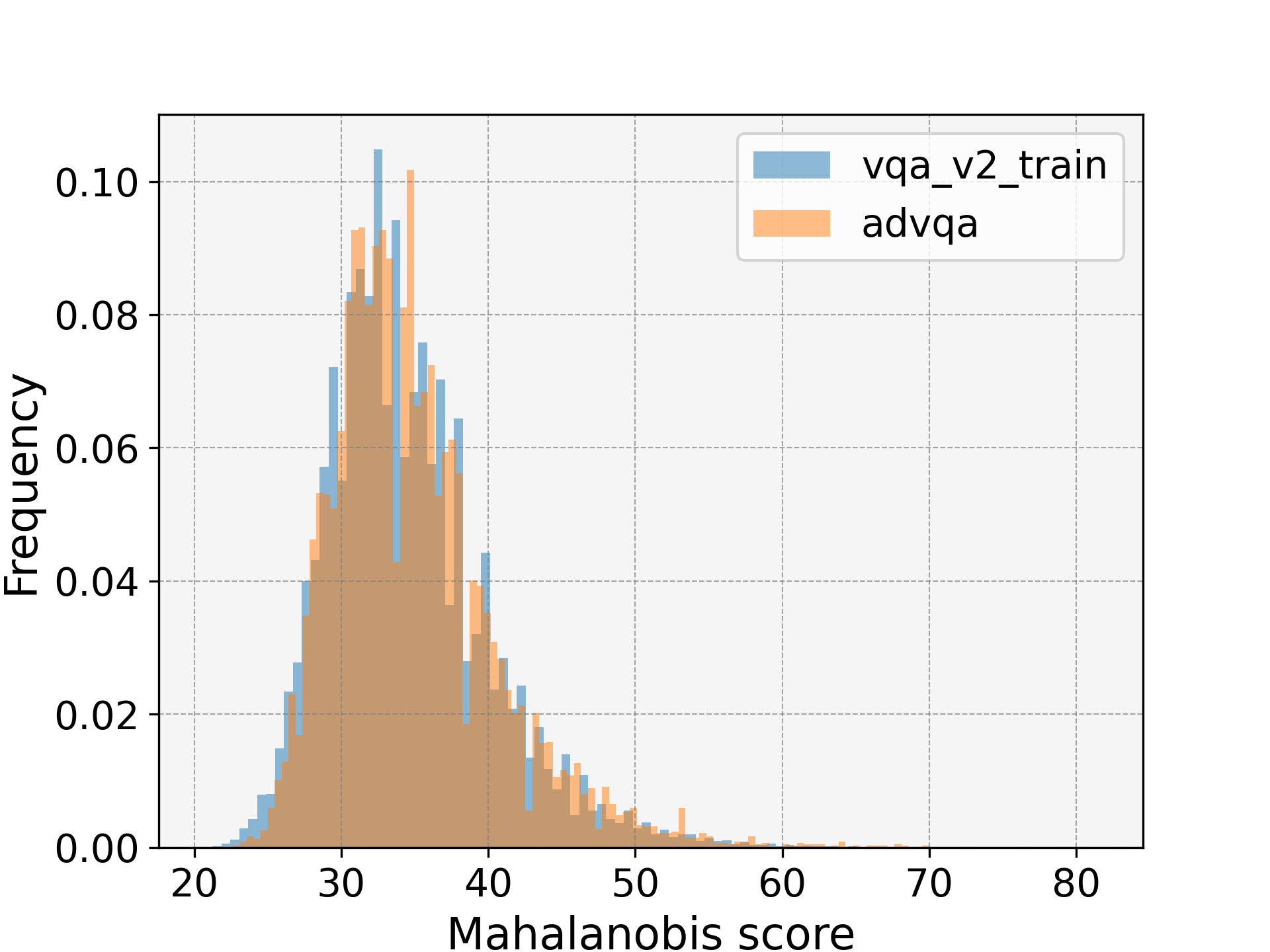}
        \caption{ADVQA}
        \label{fig:vqaadvqa}
    \end{subfigure}
    \hfill
    \begin{subfigure}[b]{0.3\linewidth}
        \centering
        \includegraphics[width=\linewidth]{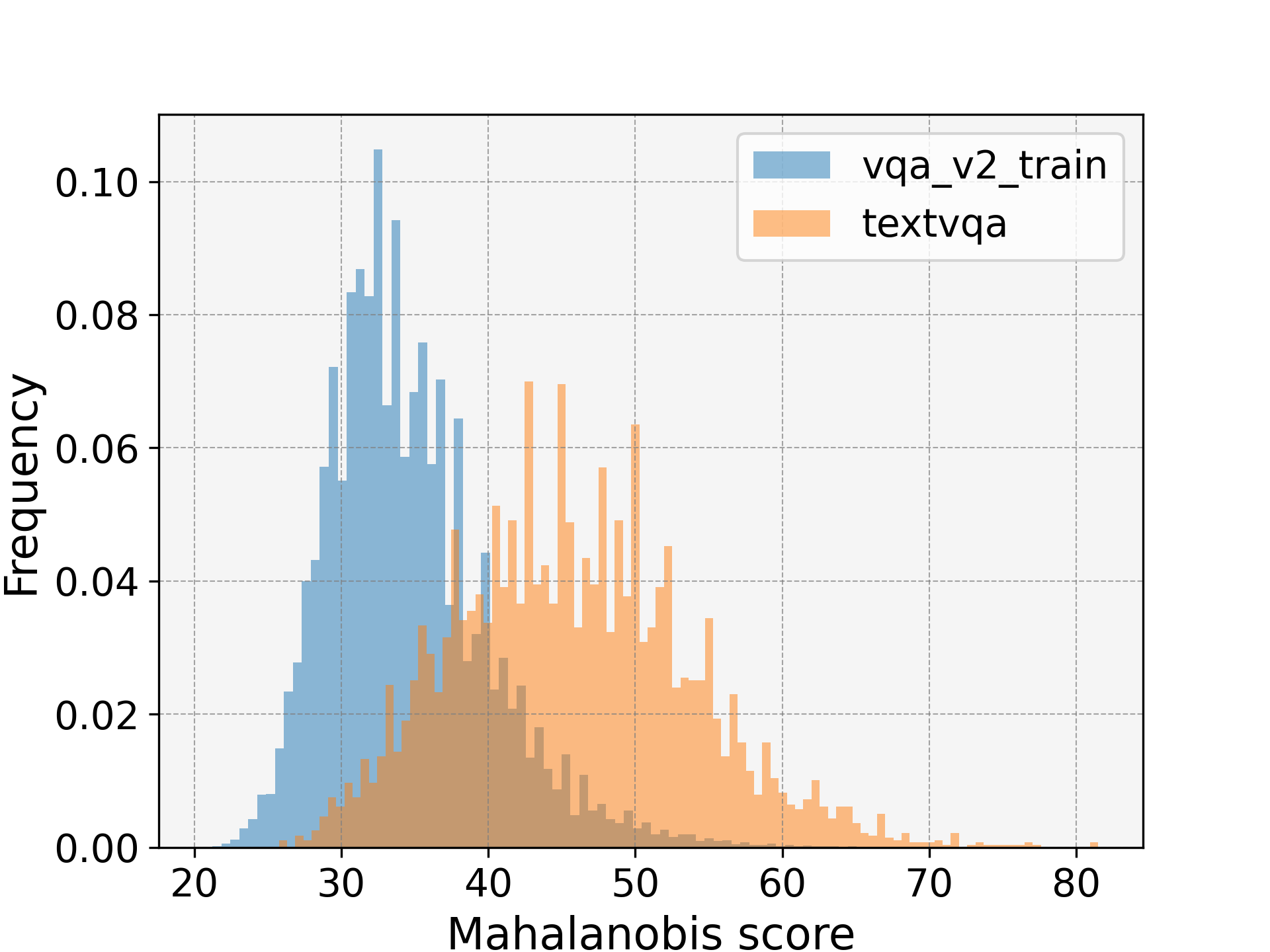}
        \caption{Text VQA}
        \label{fig:vqatextvqa}
    \end{subfigure}
    \hfill
    \begin{subfigure}[b]{0.3\linewidth}
        \centering
        \includegraphics[width=\linewidth]{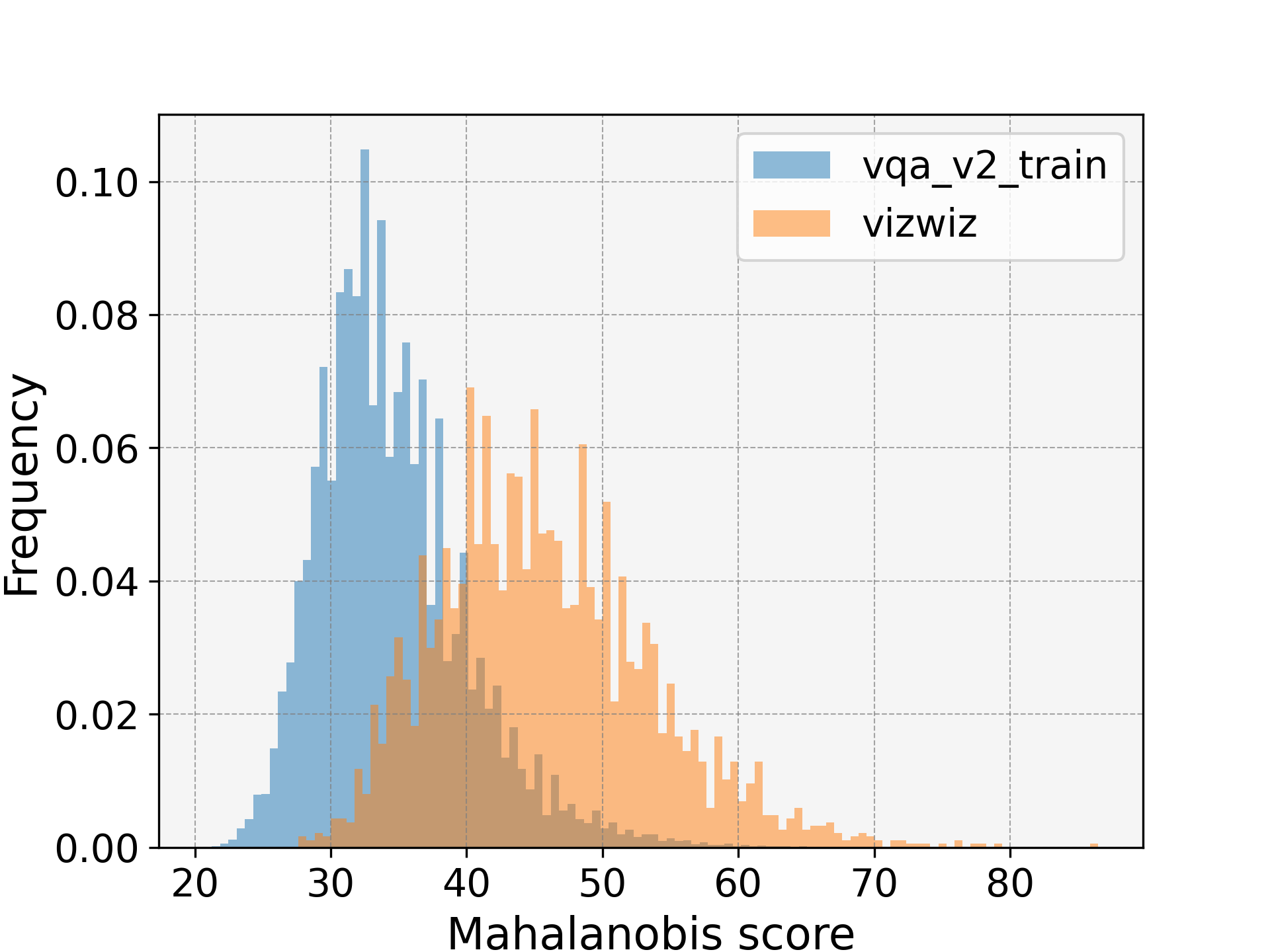}
        \caption{VizWiz}
        \label{fig:vqavizwiz}
    \end{subfigure}

    \vspace{0.5cm} %

    \begin{subfigure}[b]{0.3\linewidth}
        \centering
        \includegraphics[width=\linewidth]{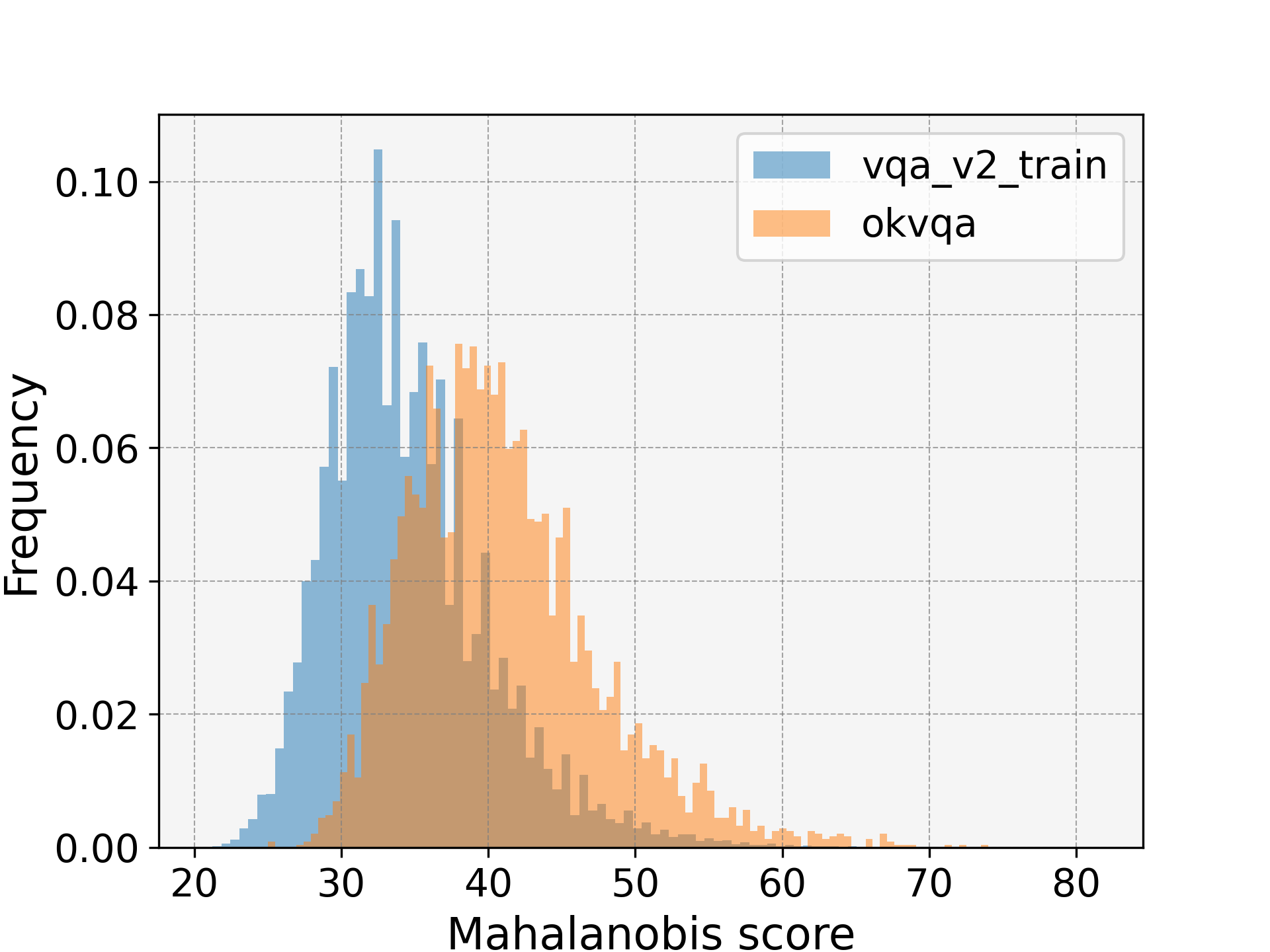}
        \caption{OKVQA}
        \label{fig:vqaokvqa}
    \end{subfigure}

    \caption{Histogram for Vanilla FT V+Q Shifts : We depict the \( S_{\text{Maha}} \) score on the V+Q shift for each sample in the VQAv2 train split in blue and the corresponding test samples in orange. For all test splits, V+Q shifts show a greater degree of shift compared to the corresponding visual and question shift.}
    \label{fig:vqa_histograms}
\end{figure*}

We display the histograms for the Mahalanobis score distribution between test datasets with the ID set.  Fig.~\ref{fig:v_histograms},~\ref{fig:vq_histograms} and~\ref{fig:vqa_histograms} are visual, question and joint shifts from vanilla FT repectively.

The histograms show that under Vanilla FT, visual shifts are minimal across most VQA datasets except for VizWiz, while question shifts are greater for further OOD datasets. Combined visual and question shifts exhibit the largest deviations across all test splits.

\section{Correlation between Shift \& Performance}
\label{sec:suppl_dist_perf}

Tab.~\ref{tab:corr_tab} shows the correlation between shift and performance for different embeddings under different fine-tuning methods. Overall, visual and joint shifts exhibit the strongest correlation with performance across all types of methods. The negative correlation supports the intuition that larger shifts in all modalities degrade VQA performance.

\begin{table}[!h]
    \centering
    \caption{Correlation between Shift Score vs. Performance for different embeddings under various fine-tuning methods. 
    }
    \label{tab:corr_tab}
    \small %
    \begin{tabular}{c|c|c|c}
        \toprule
        \textbf{Method} & \textbf{V} & \textbf{Q} & \textbf{Joint} \\
        \midrule
        Pre-Train~\cite{huggingface_paligemma_3b_pt_224} & -0.80 & -0.66  & -0.80 \\
        Vanilla FT$_\text{LoRA}$~\cite{hu_lora_2021} & -0.74 & -0.63 & -0.78 \\
        Linear Prob$_\text{LoRA}$ & -0.82 & -0.55 & -0.81 \\
        LP-FT$_\text{LoRA}$~\cite{kumar_fine-tuning_2022} & -0.75 & -0.58 & -0.80 \\
        FTP$_\text{LoRA}$~\cite{tian_fast_2023} & -0.86 & -0.63 & -0.75 \\
        SPD$_\text{LoRA}$~\cite{tian_rethinking_2024} & -0.52 & -0.61 & -0.79 \\
        \bottomrule
    \end{tabular}
\end{table}

\section{Correlation between Uni- \& Multi-Modal Shifts per Dataset}
\label{sec:suppl_xconcept}

\begin{figure}[!h]
    \centering
    \begin{subfigure}[t]{1\linewidth}
        \centering
        \includegraphics[width=\linewidth]{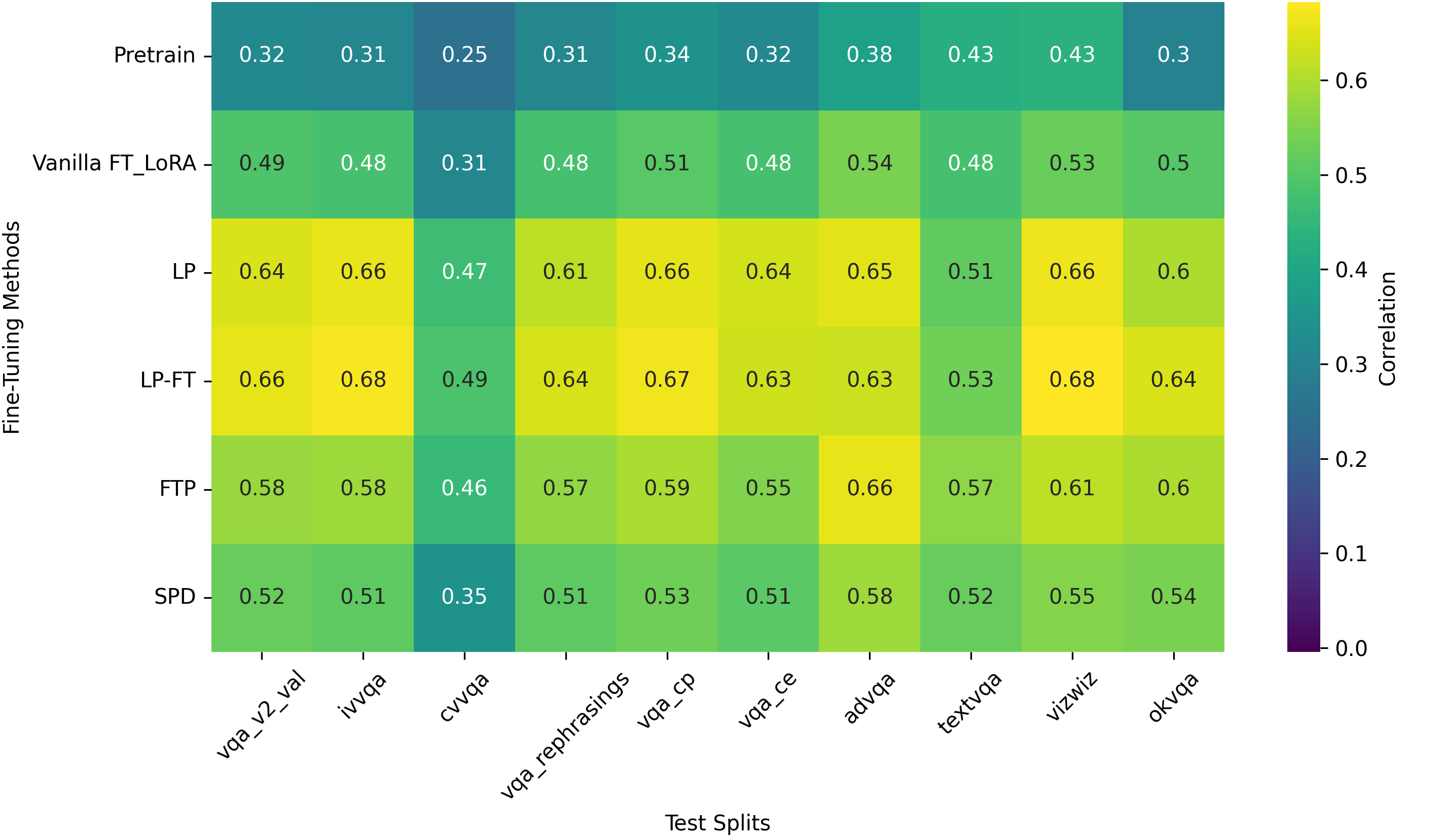}
        \caption{Question-Joint shift correlation heatmap}
        \label{fig:ques_ft_correlation_heatmap}
    \end{subfigure}
    \begin{subfigure}[t]{1\linewidth}
        \centering
        \includegraphics[width=\linewidth]{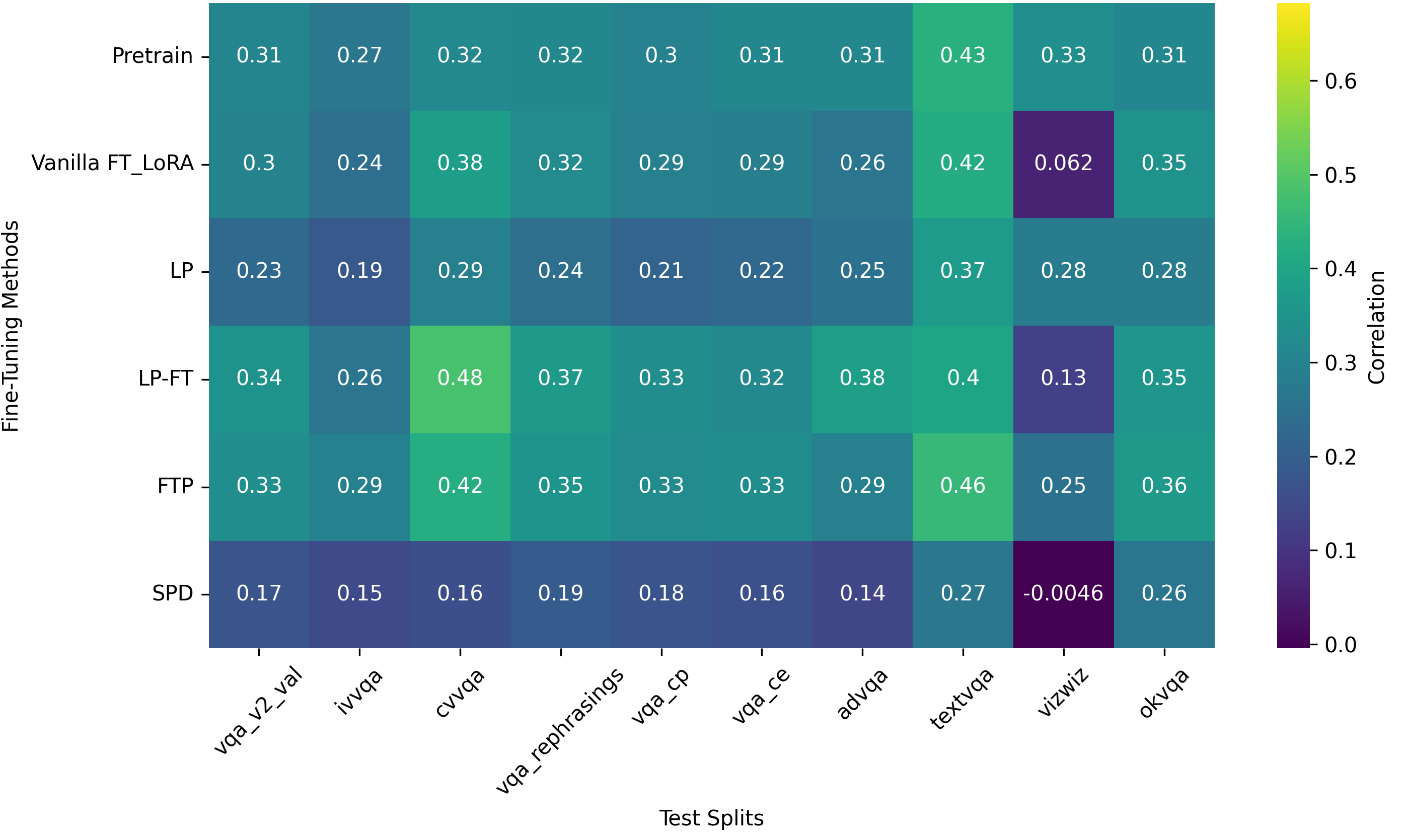}
        \caption{Image-Joint shift correlation heatmap}
        \label{fig:img_final_correlation_heatmap}
    \end{subfigure}
    \caption{Heatmap of correlation between uni-modal and multi-modal shifts per dataset.}
    \label{fig:crosscorrheatmap}
\end{figure}

Fig.~\ref{fig:crosscorrheatmap} shows the heatmap of the correlation between uni-modal and multi-modal shifts per dataset. Question-joint shift correlations are higher than image-joint shift correlations across all VQA datasets and fine-tuning methods. However, pre-train model maintains similar correlation between both modalities. Vanilla FT and SPD exhibits the lowest question-joint shift correlation shown by the darkest row color across all fine-tuning methods in Fig.~\ref{fig:ques_ft_correlation_heatmap}. Whilst, SPD shows the lowest image-joint shift correlation across the datasets in Fig.~\ref{fig:img_final_correlation_heatmap}.

\section{Modality Importance of different Datasets and Fine-Tuning Methods}
\label{sec:suppl_xttn}

\begin{figure*}[!h]
    \centering
    \begin{subfigure}{0.166\linewidth}
        \centering
        \includegraphics[width=\linewidth]{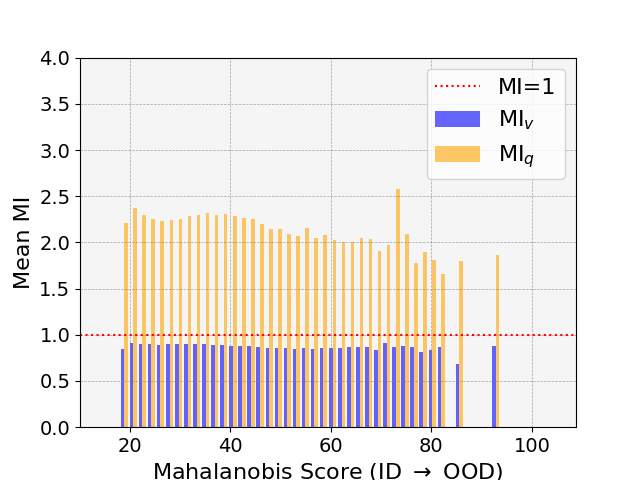}
        \caption{VQAv2 val, PT}
    \end{subfigure}%
    \begin{subfigure}{0.166\linewidth}
        \centering
        \includegraphics[width=\linewidth]{figures/new_attention_line_plot/gt_fft/lora/vqa_v2_val_joint_attn_ratio.png}
        \caption{VQAv2, FT}
    \end{subfigure}%
    \begin{subfigure}{0.166\linewidth}
        \centering
        \includegraphics[width=\linewidth]{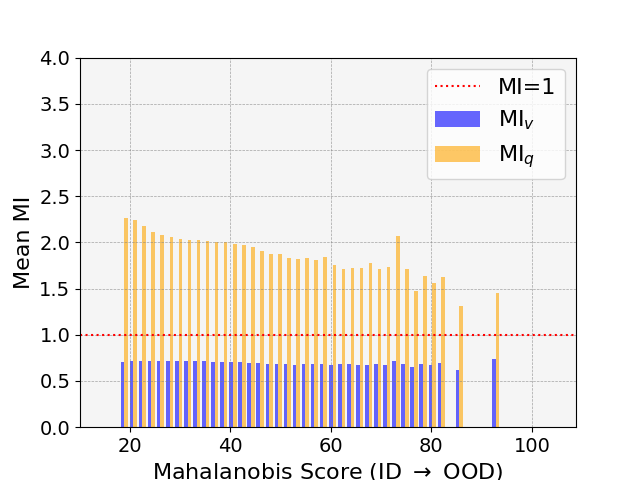}
        \caption{VQAv2 val, LP}
    \end{subfigure}%
    \begin{subfigure}{0.166\linewidth}
        \centering
        \includegraphics[width=\linewidth]{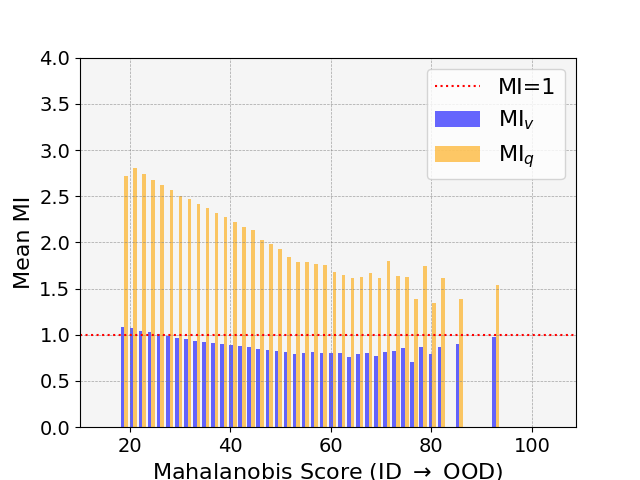}
        \caption{VQAv2 val, LP-FT}
    \end{subfigure}%
    \begin{subfigure}{0.166\linewidth}
        \centering
        \includegraphics[width=\linewidth]{figures/new_attention_line_plot/gt_fft/ftp/vqa_v2_val_joint_attn_ratio.png}
        \caption{VQAv2, FTP}
    \end{subfigure}%
    \begin{subfigure}{0.166\linewidth}
        \centering
        \includegraphics[width=\linewidth]{figures/new_attention_line_plot/gt_fft/spd/vqa_v2_val_joint_attn_ratio.png}
        \caption{VQAv2, SPD}
    \end{subfigure}

    \vskip\baselineskip
    \begin{subfigure}{0.166\linewidth}
        \centering
        \includegraphics[width=\linewidth]{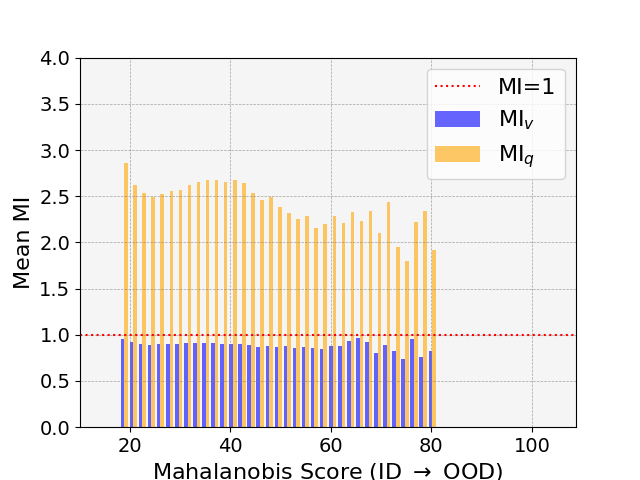}
        \caption{IV-VQA, PT}
    \end{subfigure}%
    \begin{subfigure}{0.166\linewidth}
        \centering
        \includegraphics[width=\linewidth]{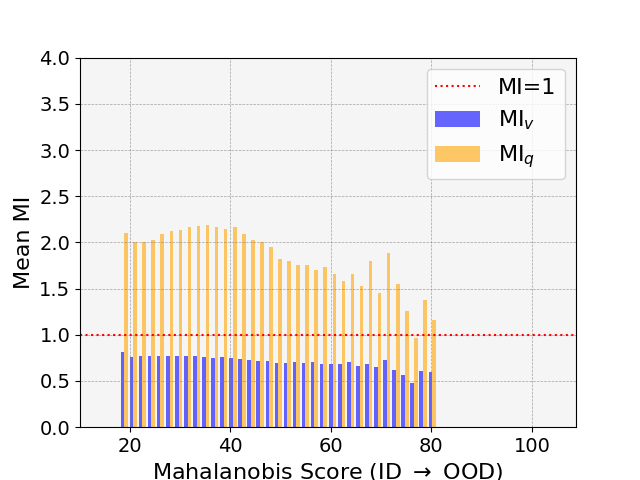}
        \caption{IV-VQA, FT}
    \end{subfigure}%
    \begin{subfigure}{0.166\linewidth}
        \centering
        \includegraphics[width=\linewidth]{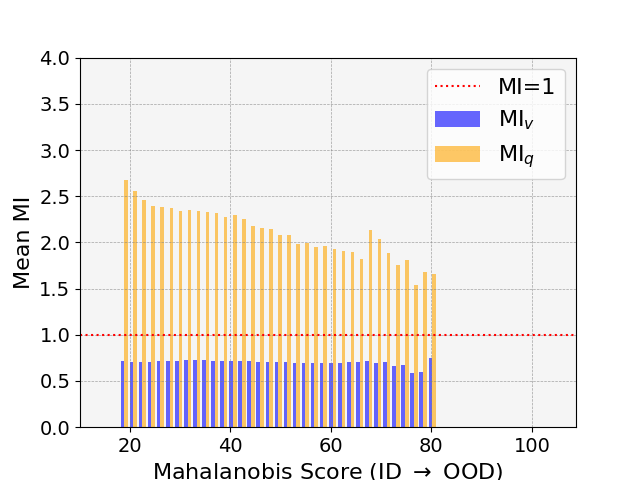}
        \caption{IV-VQA, LP}
    \end{subfigure}%
    \begin{subfigure}{0.166\linewidth}
        \centering
        \includegraphics[width=\linewidth]{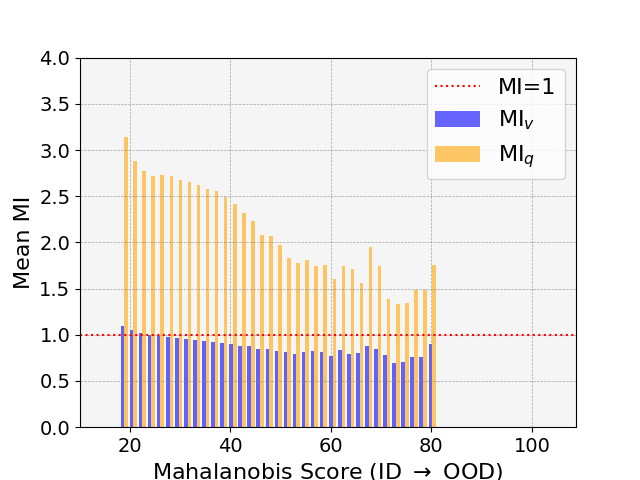}
        \caption{IV-VQA, LP-FT}
    \end{subfigure}%
    \begin{subfigure}{0.166\linewidth}
        \centering
        \includegraphics[width=\linewidth]{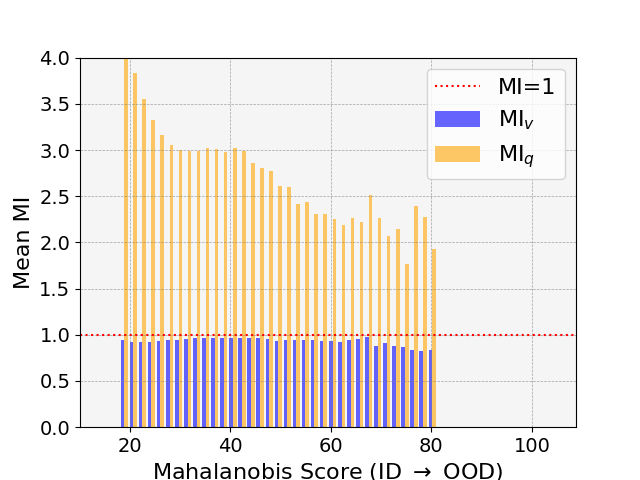}
        \caption{IV-VQA, FTP}
    \end{subfigure}%
    \begin{subfigure}{0.166\linewidth}
        \centering
        \includegraphics[width=\linewidth]{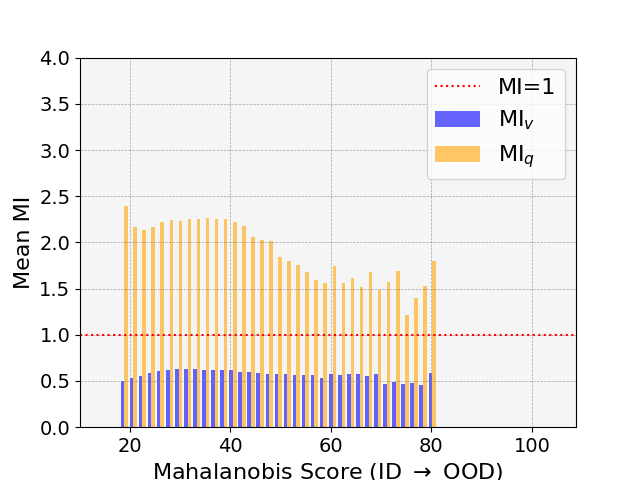}
        \caption{IV-VQA, SPD}
    \end{subfigure}

    \vskip\baselineskip
    \begin{subfigure}{0.166\linewidth}
        \centering
        \includegraphics[width=\linewidth]{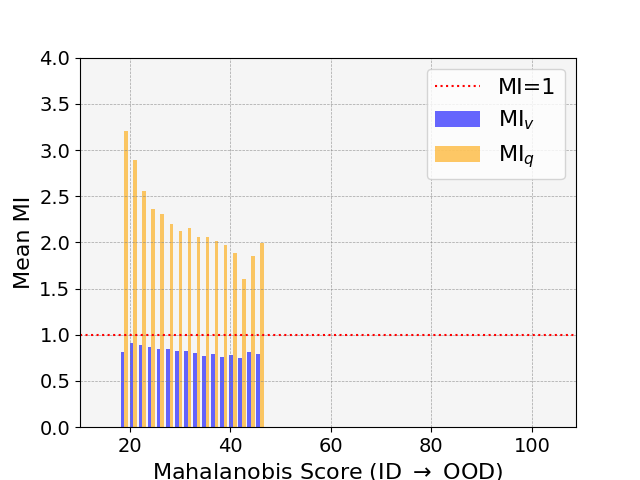}
        \caption{CV-VQA, PT}
    \end{subfigure}%
    \begin{subfigure}{0.166\linewidth}
        \centering
        \includegraphics[width=\linewidth]{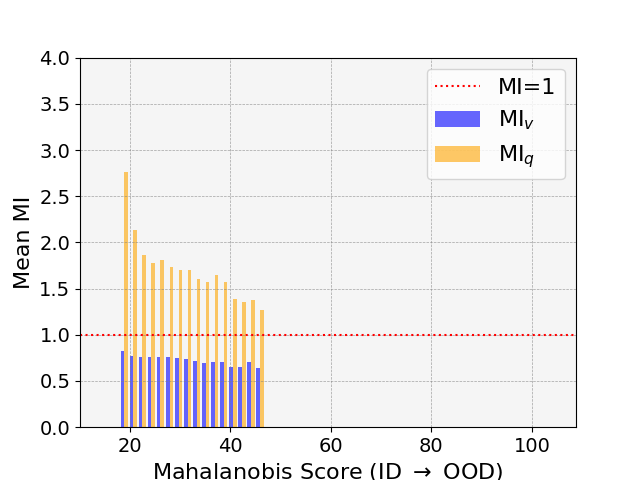}
        \caption{CV-VQA, FT}
    \end{subfigure}%
    \begin{subfigure}{0.166\linewidth}
        \centering
        \includegraphics[width=\linewidth]{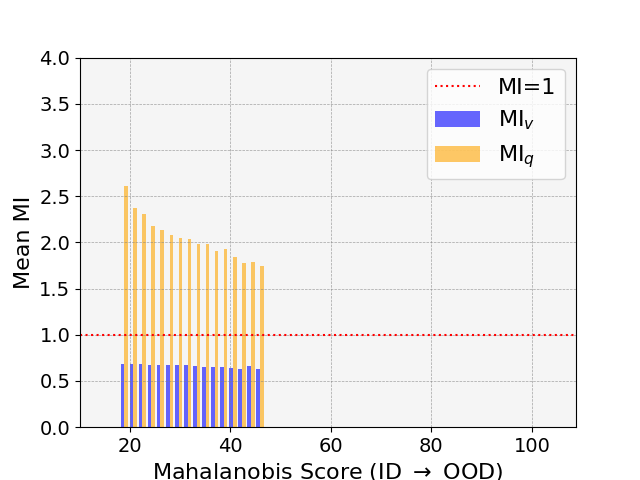}
        \caption{CV-VQA, LP}
    \end{subfigure}%
    \begin{subfigure}{0.166\linewidth}
        \centering
        \includegraphics[width=\linewidth]{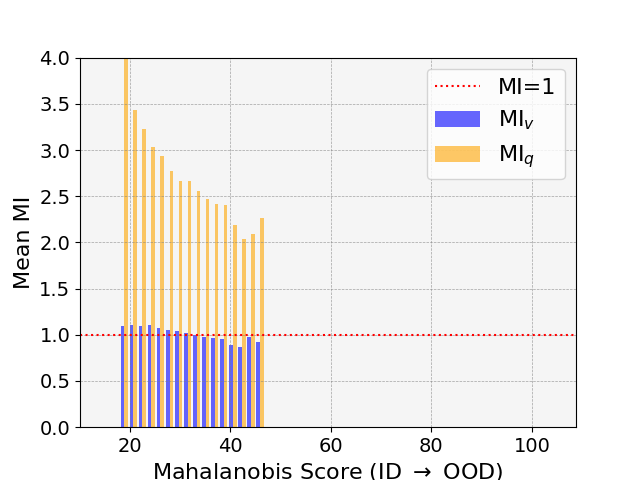}
        \caption{CV-VQA, LP-FT}
    \end{subfigure}%
    \begin{subfigure}{0.166\linewidth}
        \centering
        \includegraphics[width=\linewidth]{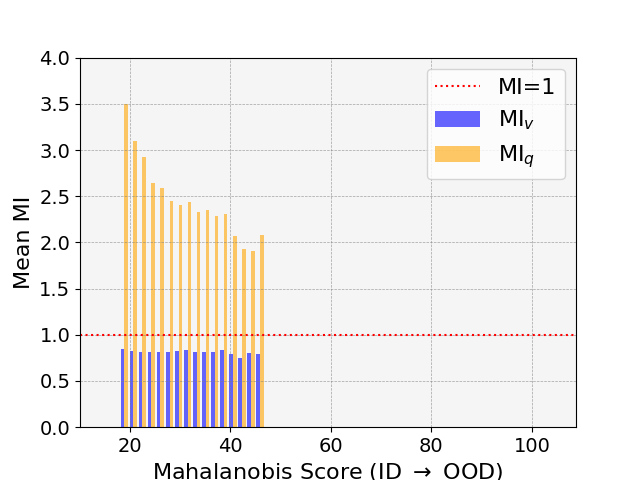}
        \caption{CV-VQA, FTP}
    \end{subfigure}%
    \begin{subfigure}{0.166\linewidth}
        \centering
        \includegraphics[width=\linewidth]{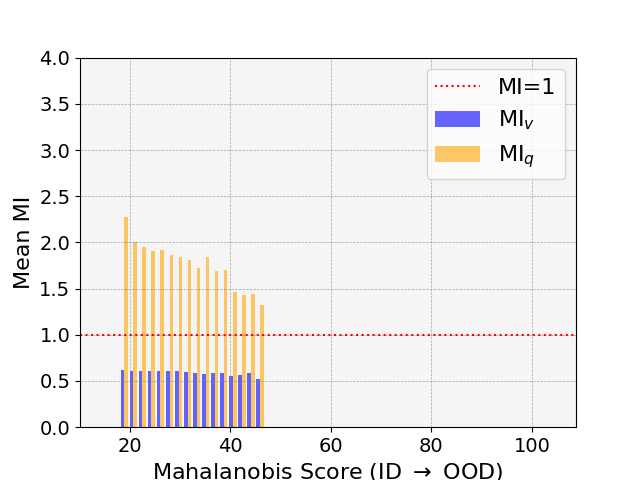}
        \caption{CV-VQA, SPD}
    \end{subfigure}

    \vskip\baselineskip
    \begin{subfigure}{0.166\linewidth}
        \centering
        \includegraphics[width=\linewidth]{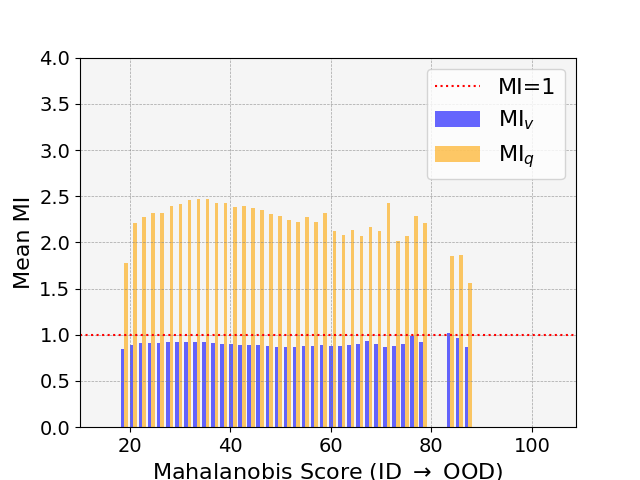}
        \caption{VQA-Rep., PT}
    \end{subfigure}%
    \begin{subfigure}{0.166\linewidth}
        \centering
        \includegraphics[width=\linewidth]{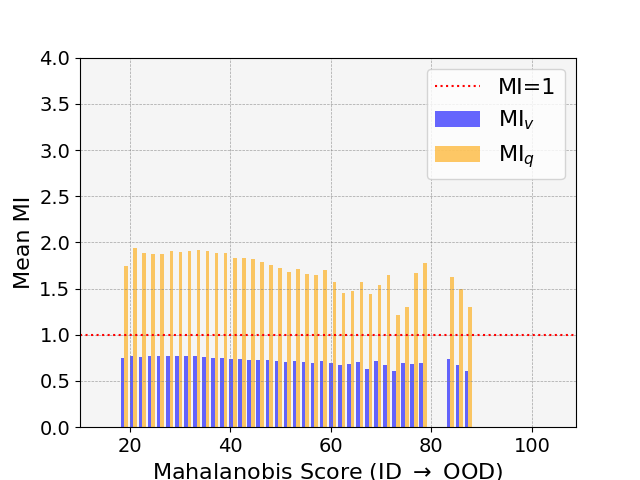}
        \caption{VQA-Rep., FT}
    \end{subfigure}%
    \begin{subfigure}{0.166\linewidth}
        \centering
        \includegraphics[width=\linewidth]{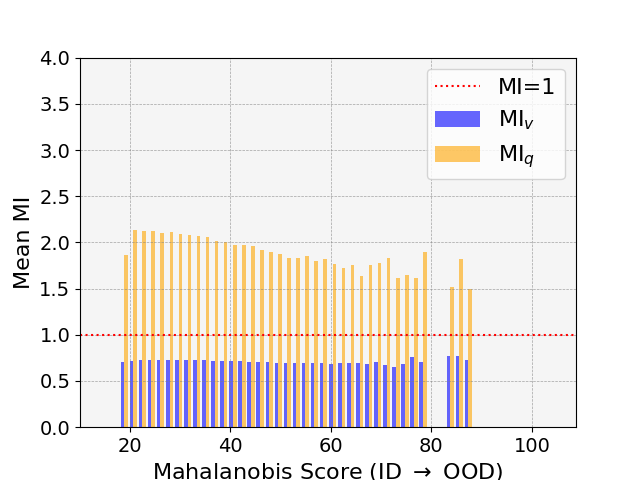}
        \caption{VQA-Rep., LP}
    \end{subfigure}%
    \begin{subfigure}{0.166\linewidth}
        \centering
        \includegraphics[width=\linewidth]{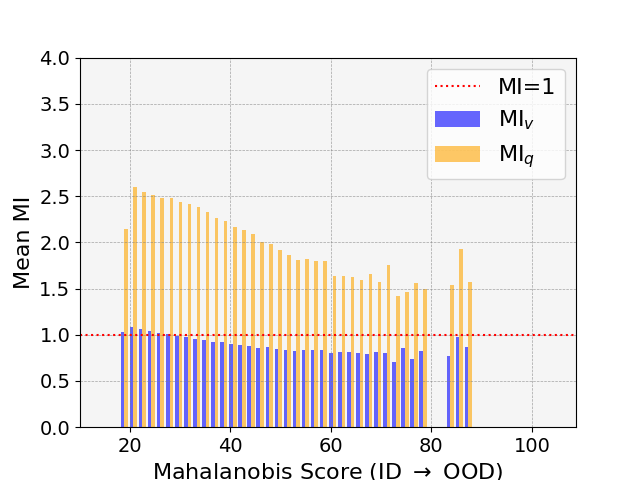}
        \caption{VQA-Rep., LP-FT}
    \end{subfigure}%
    \begin{subfigure}{0.166\linewidth}
        \centering
        \includegraphics[width=\linewidth]{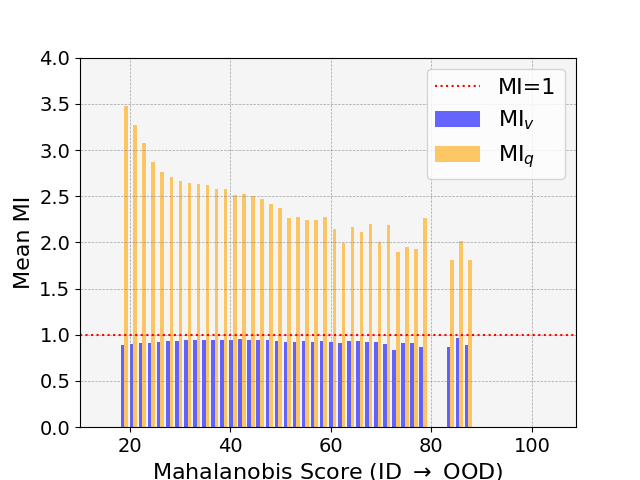}
        \caption{VQA-Rep., FTP}
    \end{subfigure}%
    \begin{subfigure}{0.166\linewidth}
        \centering
        \includegraphics[width=\linewidth]{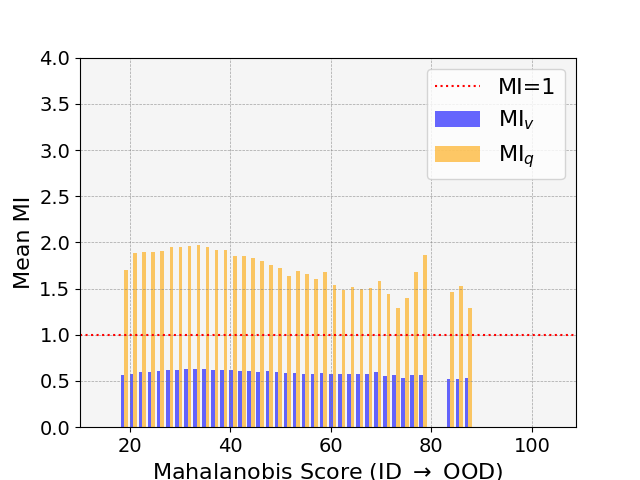}
        \caption{VQA-Rep., SPD}
    \end{subfigure}

    \vskip\baselineskip
    \begin{subfigure}{0.166\linewidth}
        \centering
        \includegraphics[width=\linewidth]{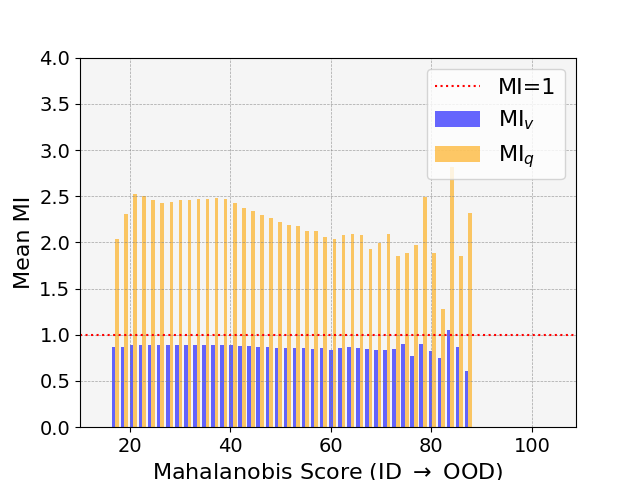}
        \caption{VQA-CP, PT}
    \end{subfigure}%
    \begin{subfigure}{0.166\linewidth}
        \centering
        \includegraphics[width=\linewidth]{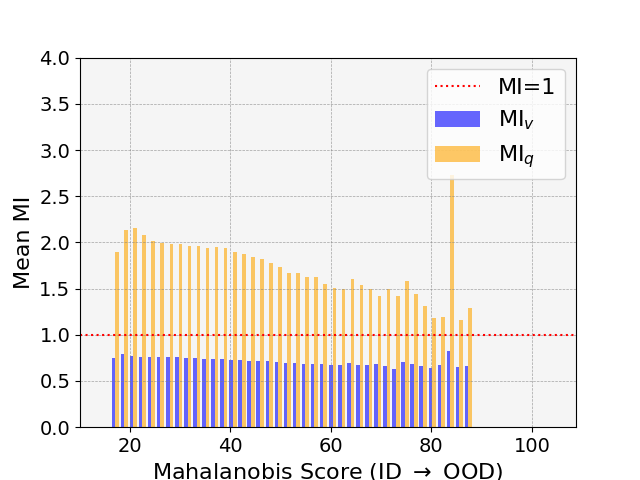}
        \caption{VQA-CP, FT}
    \end{subfigure}%
    \begin{subfigure}{0.166\linewidth}
        \centering
        \includegraphics[width=\linewidth]{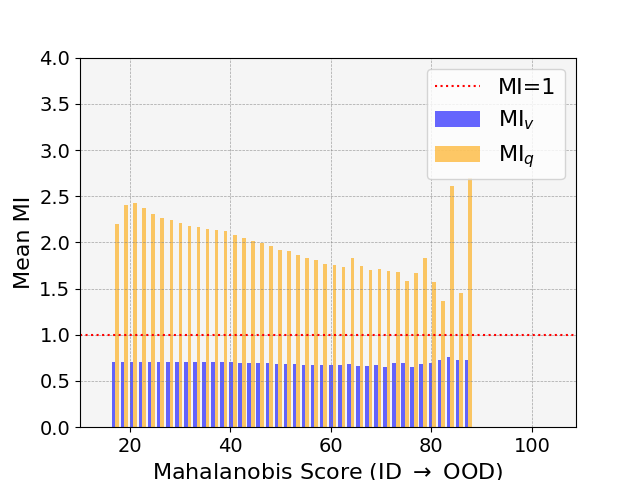}
        \caption{VQA-CP, LP}
    \end{subfigure}%
    \begin{subfigure}{0.166\linewidth}
        \centering
        \includegraphics[width=\linewidth]{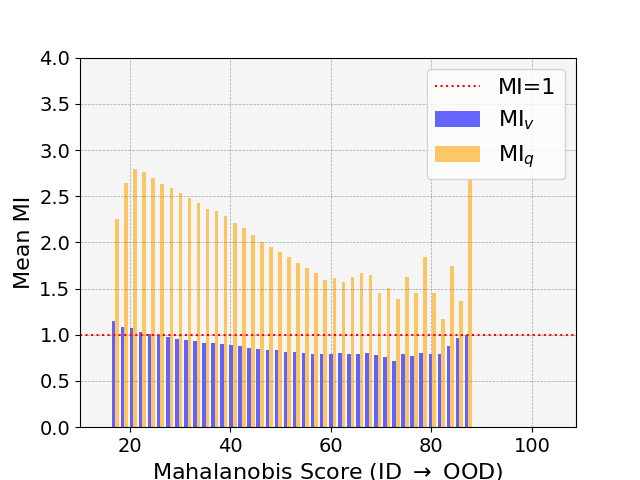}
        \caption{VQA-CP, LP-FT}
    \end{subfigure}%
    \begin{subfigure}{0.166\linewidth}
        \centering
        \includegraphics[width=\linewidth]{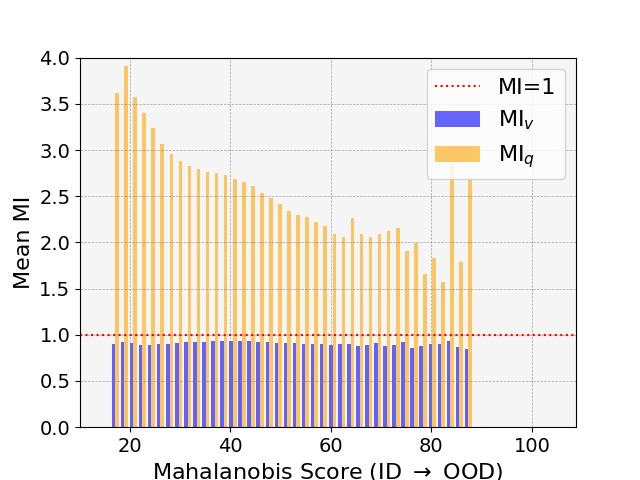}
        \caption{VQA-CP, FTP}
    \end{subfigure}%
    \begin{subfigure}{0.166\linewidth}
        \centering
        \includegraphics[width=\linewidth]{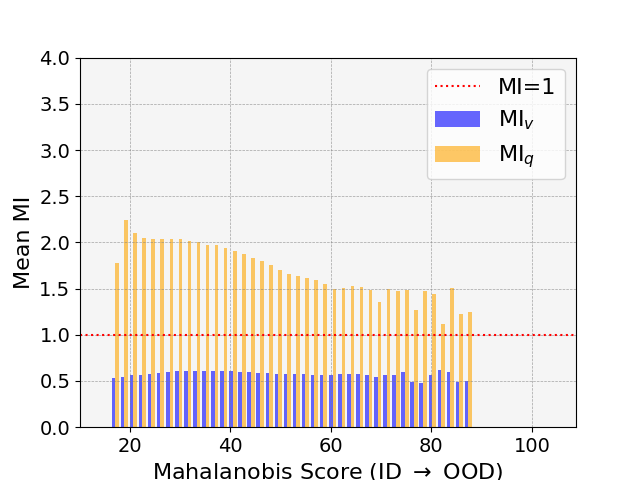}
        \caption{VQA-CP, SPD}
    \end{subfigure}

    \vskip\baselineskip
    \begin{subfigure}{0.166\linewidth}
        \centering
        \includegraphics[width=\linewidth]{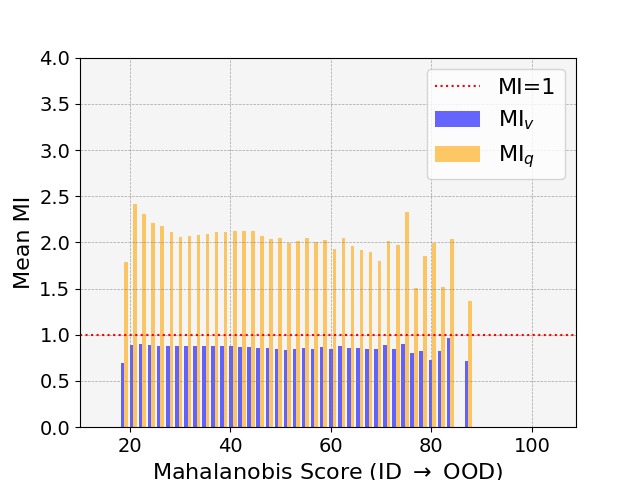}
        \caption{VQA-CE, PT}
    \end{subfigure}%
    \begin{subfigure}{0.166\linewidth}
        \centering
        \includegraphics[width=\linewidth]{figures/new_attention_line_plot/gt_fft/lora/vqa_ce_test_joint_attn_ratio.png}
        \caption{VQA-CE, FT}
    \end{subfigure}%
    \begin{subfigure}{0.166\linewidth}
        \centering
        \includegraphics[width=\linewidth]{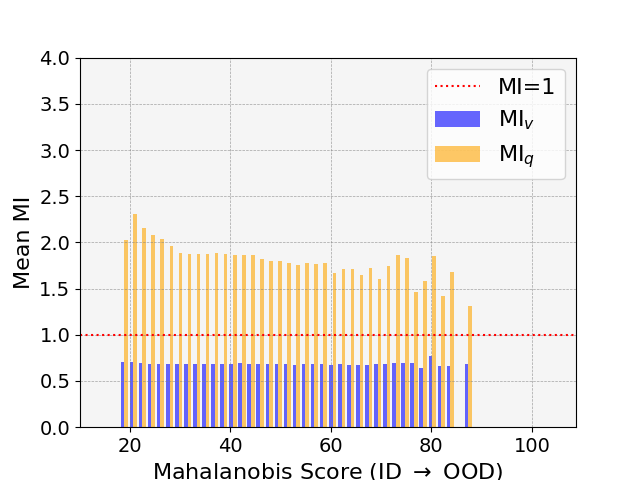}
        \caption{VQA-CE, LP}
    \end{subfigure}%
    \begin{subfigure}{0.166\linewidth}
        \centering
        \includegraphics[width=\linewidth]{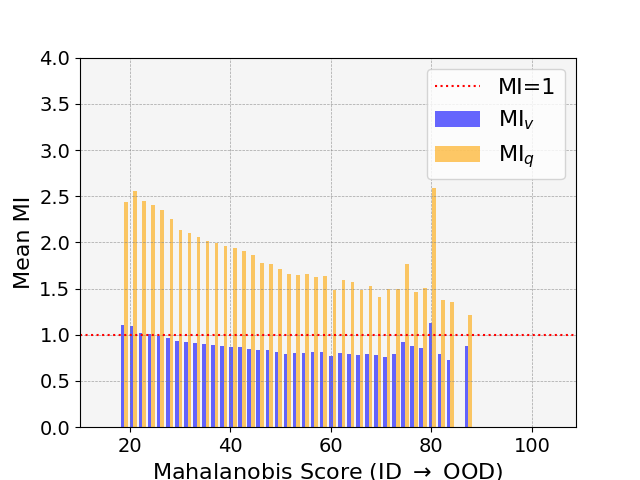}
        \caption{VQA-CE, LP-FT}
    \end{subfigure}%
    \begin{subfigure}{0.166\linewidth}
        \centering
        \includegraphics[width=\linewidth]{figures/new_attention_line_plot/gt_fft/ftp/vqa_ce_test_joint_attn_ratio.png}
        \caption{VQA-CE, FTP}
    \end{subfigure}%
    \begin{subfigure}{0.166\linewidth}
        \centering
        \includegraphics[width=\linewidth]{figures/new_attention_line_plot/gt_fft/spd/vqa_ce_test_joint_attn_ratio.png}
        \caption{VQA-CE, SPD}
    \end{subfigure}

    \vskip\baselineskip
    \begin{subfigure}{0.166\linewidth}
        \centering
        \includegraphics[width=\linewidth]{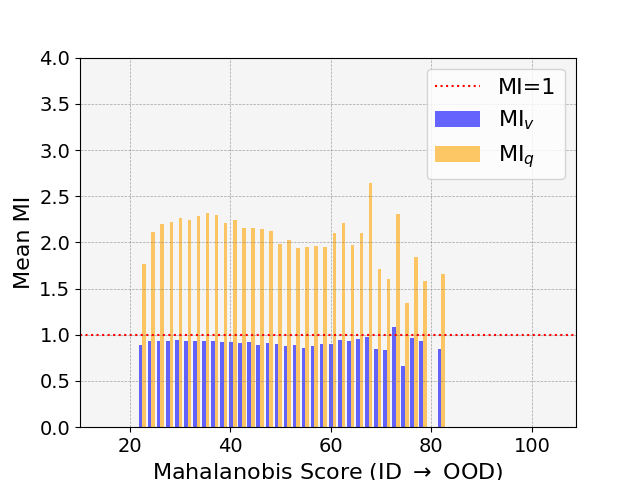}
        \caption{AdVQA, PT}
    \end{subfigure}%
    \begin{subfigure}{0.166\linewidth}
        \centering
        \includegraphics[width=\linewidth]{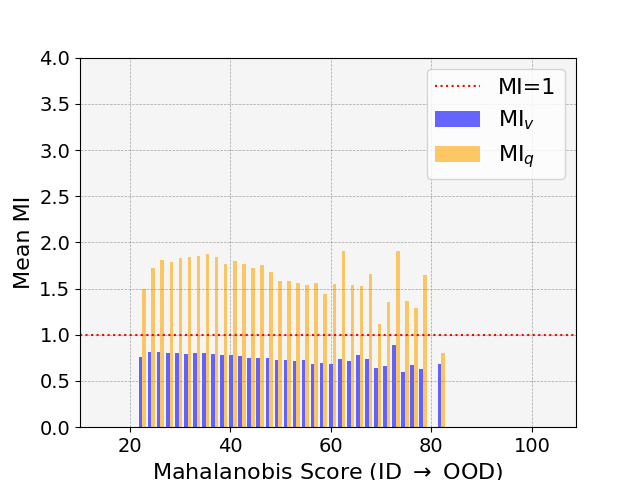}
        \caption{AdVQA, FT}
    \end{subfigure}%
    \begin{subfigure}{0.166\linewidth}
        \centering
        \includegraphics[width=\linewidth]{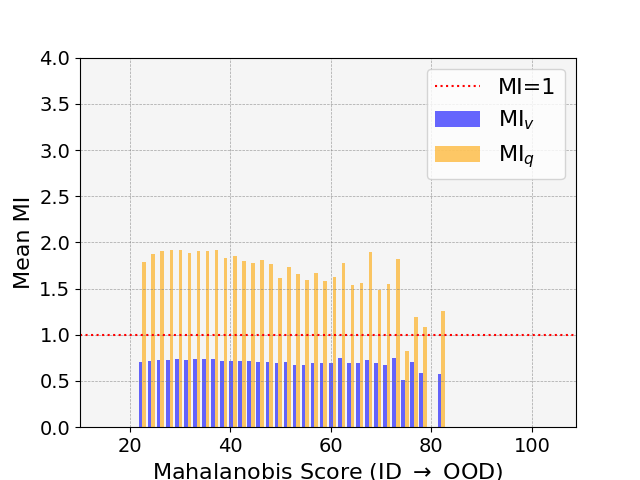}
        \caption{AdVQA, LP}
    \end{subfigure}%
    \begin{subfigure}{0.166\linewidth}
        \centering
        \includegraphics[width=\linewidth]{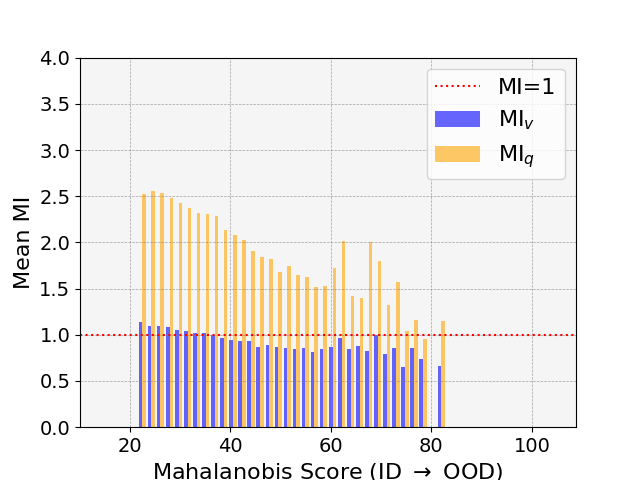}
        \caption{AdVQA, LP-FT}
    \end{subfigure}%
    \begin{subfigure}{0.166\linewidth}
        \centering
        \includegraphics[width=\linewidth]{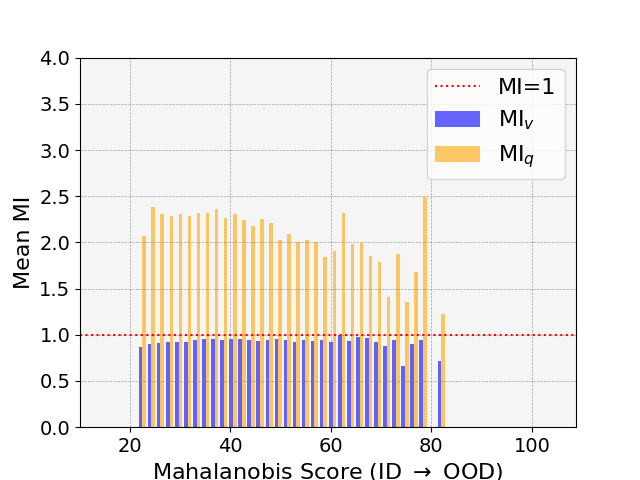}
        \caption{AdVQA, FTP}
    \end{subfigure}%
    \begin{subfigure}{0.166\linewidth}
        \centering
        \includegraphics[width=\linewidth]{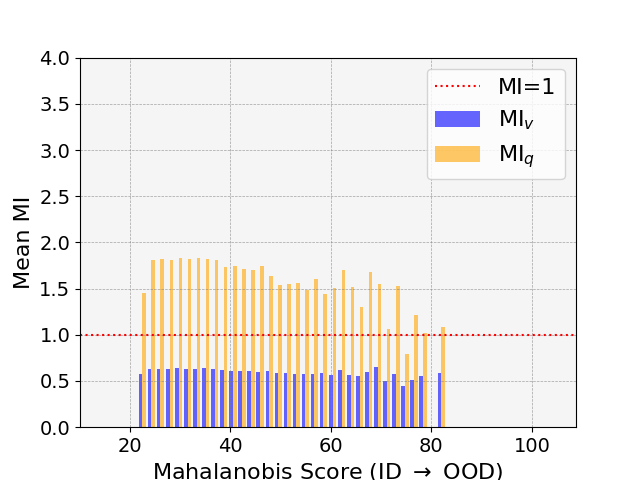}
        \caption{AdVQA, SPD}
    \end{subfigure}

    \caption{Variation of MI$_v$ and MI$_q$ w.r.t. shift score under PT, FT, LP, LP-FT, FTP and SPD across all ID and near OOD datasets. The blue and orange bars represent  MI$_v$ and MI$_q$ respectively. The red dotted line marks a reference MI of 1.
    }
    \label{fig:modality_importance_hist_id_nearood}
\end{figure*}

\begin{figure*}[!h]
    \centering
    \begin{subfigure}{0.166\linewidth}
        \centering
        \includegraphics[width=\linewidth]{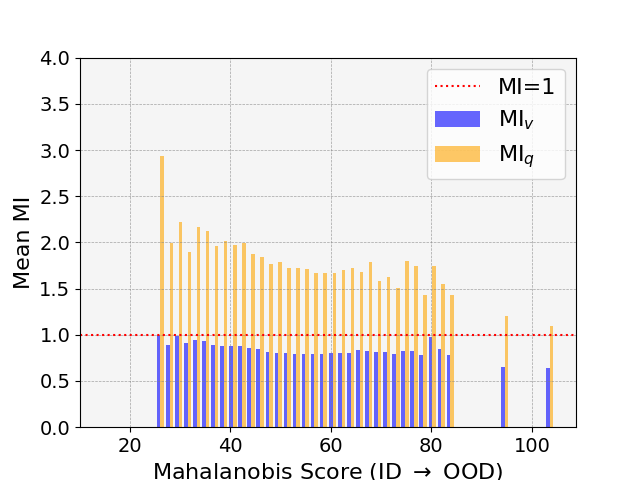}
        \caption{TextVQA, PT}
    \end{subfigure}%
    \begin{subfigure}{0.166\linewidth}
        \centering
        \includegraphics[width=\linewidth]{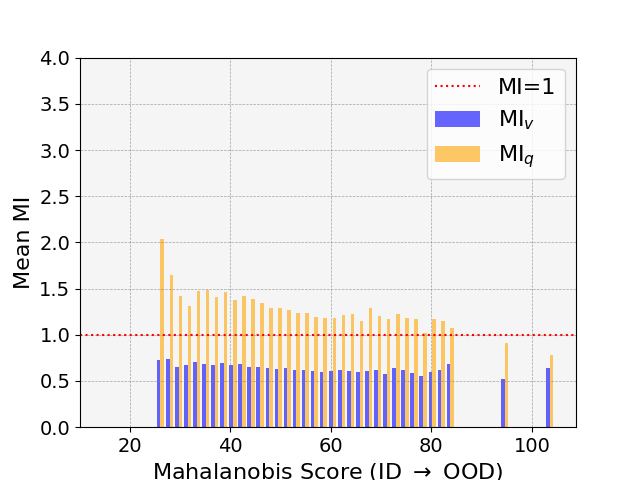}
        \caption{TextVQA, FT}
    \end{subfigure}%
    \begin{subfigure}{0.166\linewidth}
        \centering
        \includegraphics[width=\linewidth]{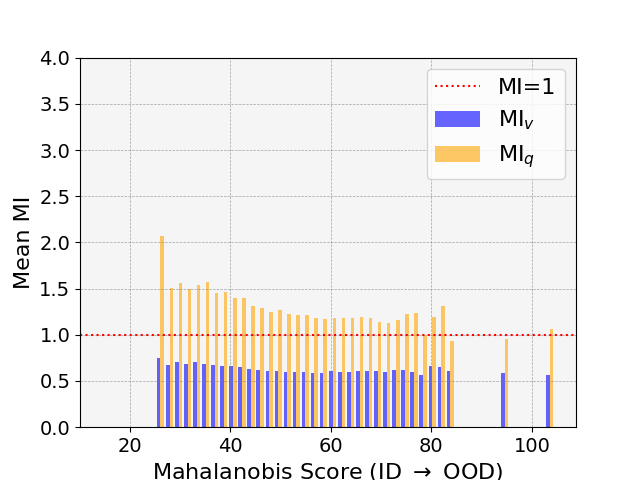}
        \caption{TextVQA, LP}
    \end{subfigure}%
    \begin{subfigure}{0.166\linewidth}
        \centering
        \includegraphics[width=\linewidth]{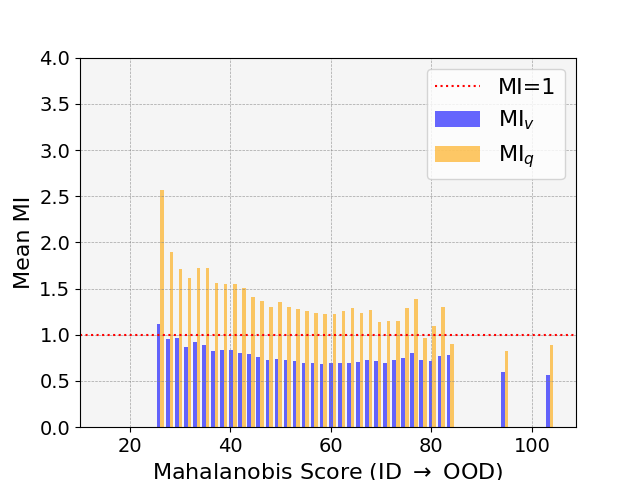}
        \caption{TextVQA, LP-FT}
    \end{subfigure}%
    \begin{subfigure}{0.166\linewidth}
        \centering
        \includegraphics[width=\linewidth]{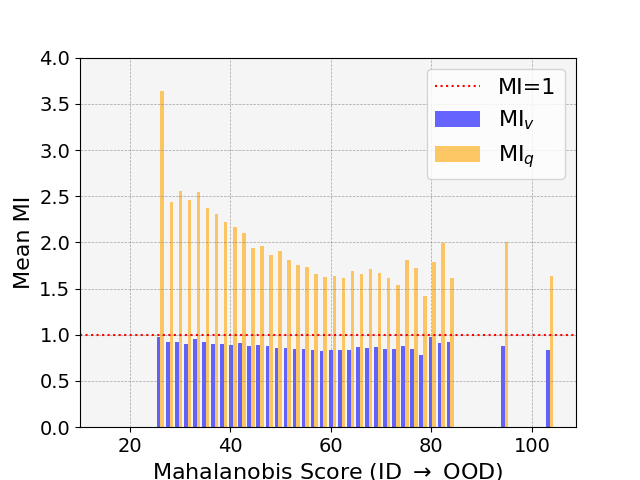}
        \caption{TextVQA, FTP}
    \end{subfigure}%
    \begin{subfigure}{0.166\linewidth}
        \centering
        \includegraphics[width=\linewidth]{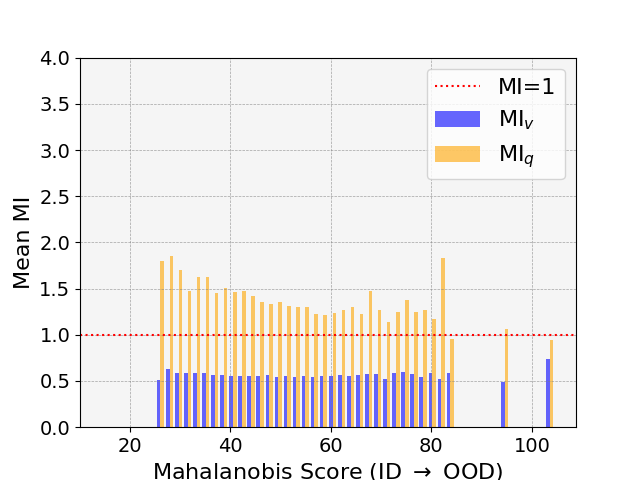}
        \caption{TextVQA, SPD}
    \end{subfigure}

    \vskip\baselineskip
    \begin{subfigure}{0.166\linewidth}
        \centering
        \includegraphics[width=\linewidth]{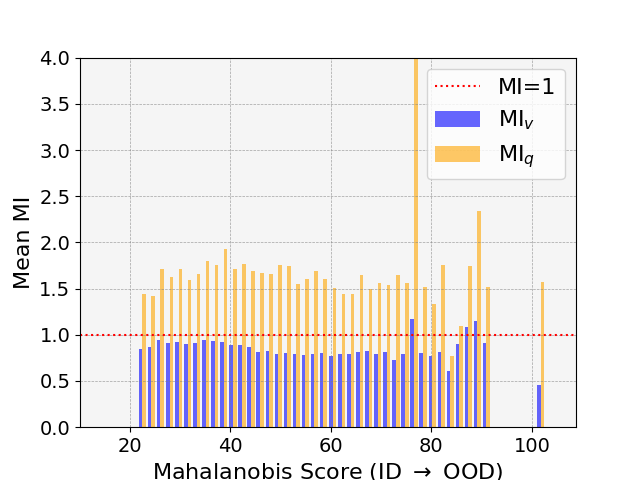}
        \caption{VizWiz, PT}
    \end{subfigure}%
    \begin{subfigure}{0.166\linewidth}
        \centering
        \includegraphics[width=\linewidth]{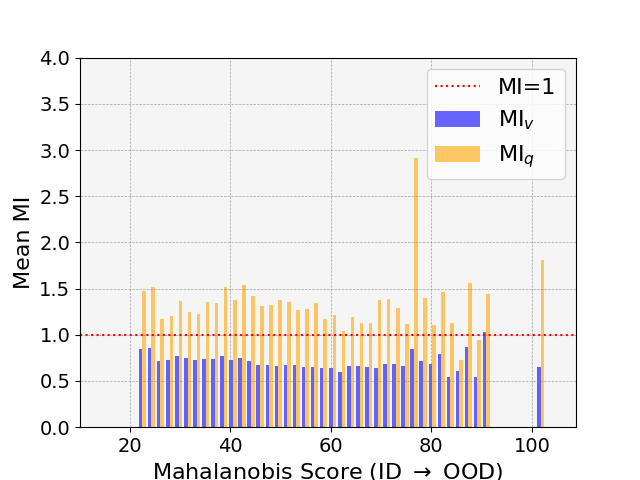}
        \caption{VizWiz, FT}
    \end{subfigure}%
    \begin{subfigure}{0.166\linewidth}
        \centering
        \includegraphics[width=\linewidth]{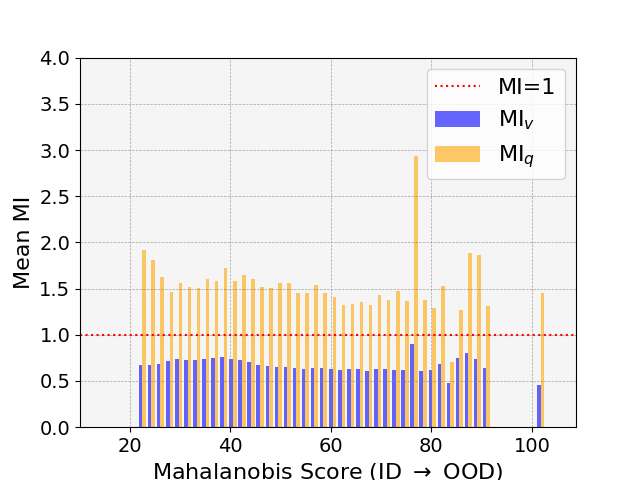}
        \caption{VizWiz, LP}
    \end{subfigure}%
    \begin{subfigure}{0.166\linewidth}
        \centering
        \includegraphics[width=\linewidth]{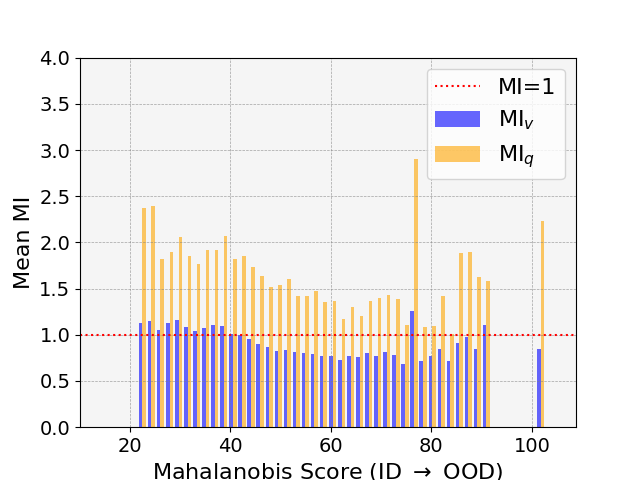}
        \caption{VizWiz, LP-FT}
    \end{subfigure}%
    \begin{subfigure}{0.166\linewidth}
        \centering
        \includegraphics[width=\linewidth]{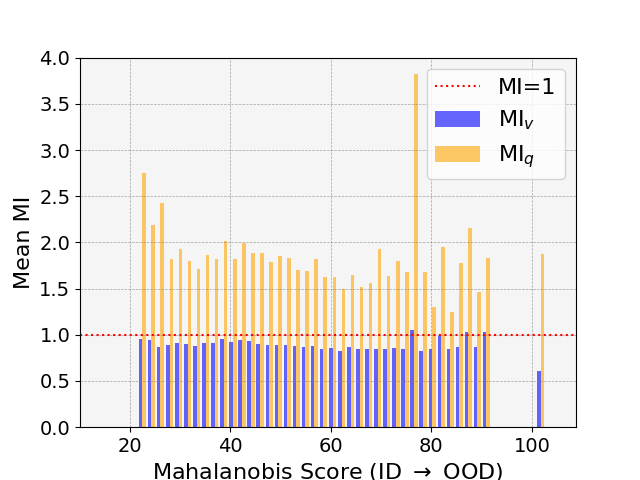}
        \caption{VizWiz, FTP}
    \end{subfigure}%
    \begin{subfigure}{0.166\linewidth}
        \centering
        \includegraphics[width=\linewidth]{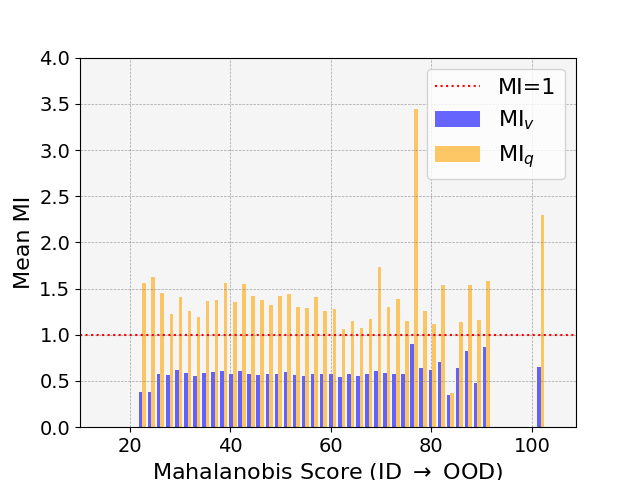}
        \caption{VizWiz, SPD}
    \end{subfigure}

    \vskip\baselineskip
    \begin{subfigure}{0.166\linewidth}
        \centering
        \includegraphics[width=\linewidth]{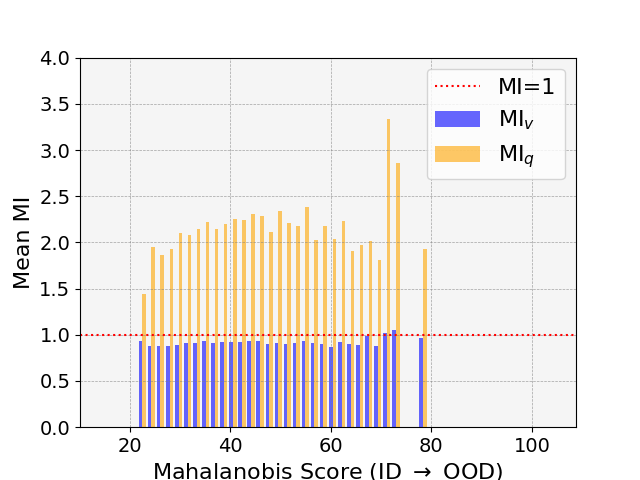}
        \caption{OK-VQA, PT}
    \end{subfigure}%
    \begin{subfigure}{0.166\linewidth}
        \centering
        \includegraphics[width=\linewidth]{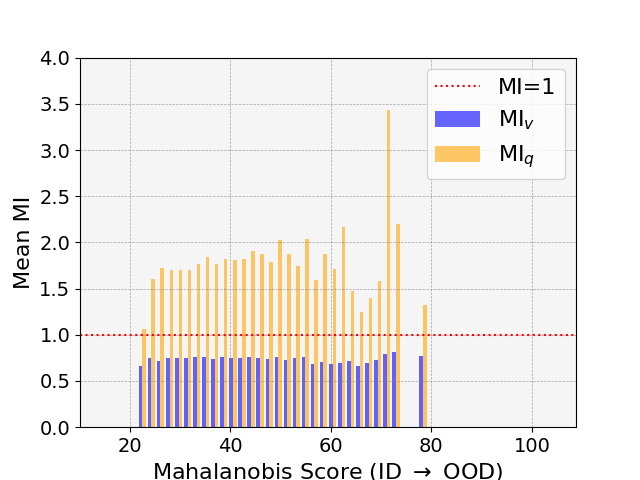}
        \caption{OK-VQA, FT}
    \end{subfigure}%
    \begin{subfigure}{0.166\linewidth}
        \centering
        \includegraphics[width=\linewidth]{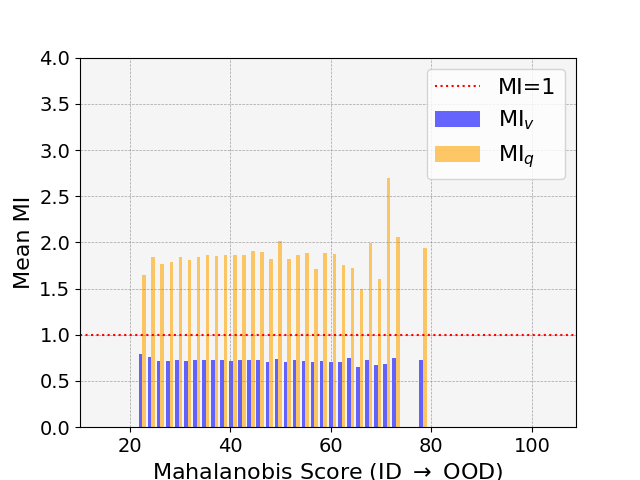}
        \caption{OK-VQA, LP}
    \end{subfigure}%
    \begin{subfigure}{0.166\linewidth}
        \centering
        \includegraphics[width=\linewidth]{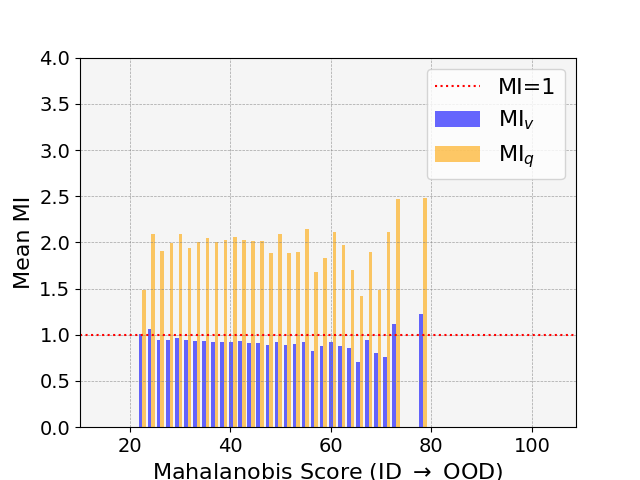}
        \caption{OK-VQA, LP-FT}
    \end{subfigure}%
    \begin{subfigure}{0.166\linewidth}
        \centering
        \includegraphics[width=\linewidth]{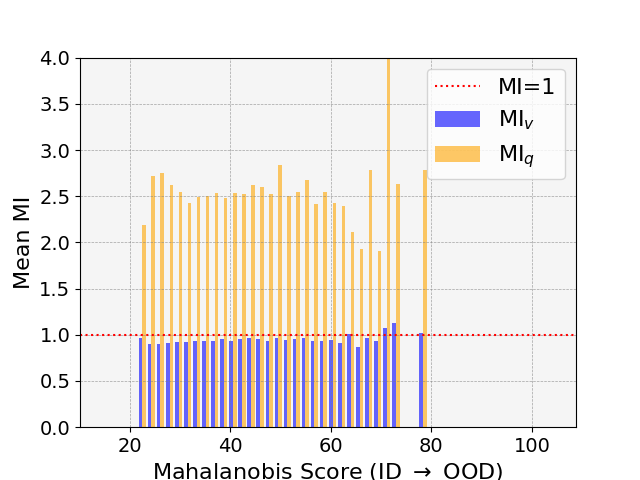}
        \caption{OK-VQA, FTP}
    \end{subfigure}%
    \begin{subfigure}{0.166\linewidth}
        \centering
        \includegraphics[width=\linewidth]{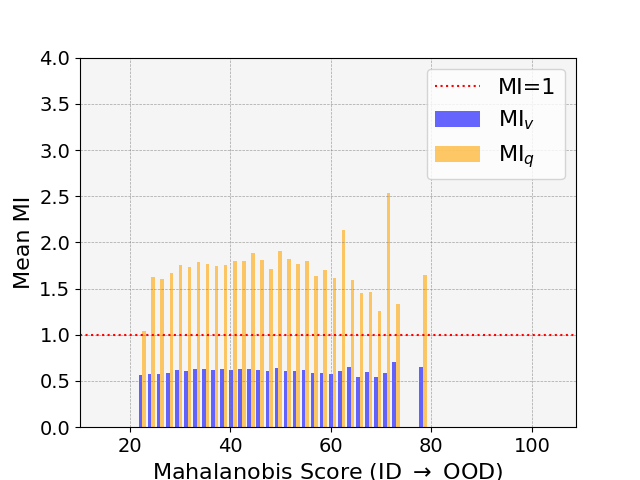}
        \caption{OK-VQA, SPD}
    \end{subfigure}
    
    \caption{Variation of MI$_v$ and MI$_q$ w.r.t. shift score under PT, FT, LP, LP-FT, FTP and SPD across all far OOD datasets. The blue and orange bars represent  MI$_v$ and MI$_q$ respectively. The red dotted line marks a reference MI of 1.
    }
    \label{fig:modality_importance_hist_farood}
\end{figure*}

Fig.~\ref{fig:modality_importance_hist_id_nearood} and~\ref{fig:modality_importance_hist_farood} show the variation of MI$_v$ and MI$_q$ w.r.t. shift score under all datasets and fine-tuning methods.
Overall, intra-modality attention is more dominant than inter-modality attention. There is a stronger intra-modality attention for text tokens than image tokens. In OOD samples, text tokens increasingly attend to image tokens. A more robust model tends to have higher MI$_q$ and lower MI$_v$.

\section{Additional Results using Full Fine-Tuning and LLaVA}
\label{sec:suppl_llava}

We present additional results in Tab.~\ref{tab:vqa_llava}, including LLaVA-7B~\cite{liu2023visualinstructiontuning} with LoRA and PaliGemma-3B with full fine-tuning. 
These results are consistent with PaliGemma with LoRA: FTP and SPD remain relatively robust models, which validates the credibility of our analysis.

\begin{table}[htb]
        \centering
          \caption{\textbf{LLaVA-7B (LoRA) and PaliGemma-3B (Full Fine-Tuning) Fine-Tuned on VQAv2.} 
          We sample 10\% of the VQAv2 training and validation set.
          \textbf{Bold}: best. \underline{Underline}: second best.}
          \label{tab:vqa_llava}
        \resizebox{\linewidth}{!}{
        \begin{tabular}{@{}c|c|c|c|c@{}}
        \toprule

        & VQAv2 val & Near OOD Avg. & Far OOD Avg. & OOD Avg. \\
        
        \midrule
        \rowcolor{gray! 20} \multicolumn{5}{c}{(a) LLaVA-7B with LoRA under 10\% of VQAv2 (train \& val)} \\
        \midrule
        Zero-Shot &
          \multicolumn{1}{c|}{3.27} &
          3.44 &
          \multicolumn{1}{c|}{0.68} &
          2.52
          \\ 
          
        Vanilla FT &
          \multicolumn{1}{c|}{\underline{72.49}} &
          \underline{60.07} &
          \multicolumn{1}{c|}{\underline{28.67}} &
          \underline{49.60}
          \\ 

        LP-FT &
          \multicolumn{1}{c|}{53.01}  &
          28.63 &
          \multicolumn{1}{c|}{7.64} &
          21.63
          \\

        WiSE-FT &
          \multicolumn{1}{c|}{60.47} &
          43.33 &
          \multicolumn{1}{c|}{9.07} &
          31.98
          \\

        FTP &
          \multicolumn{1}{c|}{67.95} &
          58.49 &
          \multicolumn{1}{c|}{26.21} &
          47.73
          \\

       SPD  &
          \multicolumn{1}{c|}{\textbf{73.59}} &
          \textbf{61.98} &
          \multicolumn{1}{c|}{\textbf{29.98}} &
          \textbf{51.31}
          \\

        \midrule
        \rowcolor{gray! 20} \multicolumn{5}{c}{(b) PaliGemma-3B with Full Fine-Tuning under 10\% of VQAv2 (train \& val)} \\
        \midrule
        Zero-Shot   & \multicolumn{1}{c|}{54.42} & \multicolumn{1}{c|}{45.70} & \multicolumn{1}{c|}{20.10} & 37.17 \\
        Vanilla FT  & \multicolumn{1}{c|}{\underline{95.80}}& \multicolumn{1}{c|}{60.73} & \multicolumn{1}{c|}{26.56} & 49.34 \\
        Linear Prob & \multicolumn{1}{c|}{86.80} & \multicolumn{1}{c|}{59.61} & \multicolumn{1}{c|}{24.17} & 47.80 \\
        LP-FT       & \multicolumn{1}{c|}{94.44} & \multicolumn{1}{c|}{57.13} & \multicolumn{1}{c|}{21.03} & 45.10 \\
        FTP         & \multicolumn{1}{c|}{95.40} & \multicolumn{1}{c|}{\textbf{64.33}} & \multicolumn{1}{c|}{\textbf{32.55}} & \textbf{53.74} \\
        SPD         & \multicolumn{1}{c|}{\textbf{95.84}} & \multicolumn{1}{c|}{\underline{63.92}} & \multicolumn{1}{c|}{\underline{32.46}} & \underline{53.43} \\

        \bottomrule
        \end{tabular}
        }
      \end{table}
\section{Fine-Tuning Results on GQA}
\label{sec:suppl_gqa}

We use VQAv2 as the ID dataset since most OOD VQA datasets, covering various shifts, are built on it. The only exception, GQA-OOD~\cite{kervadec2021rosesredvioletsblue} (based on GQA~\cite{hudson2019gqanewdatasetrealworld}), has only answer shifts. To further validate our findings, we fine-tune PaliGemma-3B on GQA as ID and evaluate it on GQA-OOD and VQAv2 variants (Tab.~\ref{tab:gqa_pali}). The results follow the same trend: FTP and SPD remain relatively robust, with SPD excelling on Near OOD (GQA-OOD) and FTP on Far OOD (VQAv2 and its variants). This consistency reinforces the generalizability of our analysis.

\begin{table}[htb]
\caption{\textbf{PaliGemma-3B Fine-tuned on GQA with LoRA and Evaluated on GQA-OOD, VQAv2 and its variants.} 
We sample 10\% of the GQA training set.
\textbf{Bold}: best. \underline{Underline}: second best.}
\label{tab:gqa_pali}
\resizebox{\linewidth}{!}{
\begin{tabular}{@{}c|c|c|ccc@{}}
\toprule
            & ID             & Near OOD & \multicolumn{3}{c}{Far OOD}                                                    \\ \cmidrule(l){2-6} 
            & GQA            & GQA-OOD        & VQAv2          & VQAv2 Near OOD Avg. & VQAv2 Far OOD Avg. \\ \midrule
Zero-Shot   &      41.44          & 29.33          & 54.42          & 45.70               & 20.10              \\
Vanilla FT  & \textbf{67.00} & {\underline{53.97}}    & 64.97          &                     57.08&                    23.42\\
Linear Prob & 61.70          & 50.32          & 54.27          &                     39.64&                    14.43\\
LP-FT       & 61.51          &                50.72& 55.72          &                     43.89&                    14.95\\
FTP         & 64.97          & 53.15          & \textbf{66.40} & \textbf{58.38}      & \textbf{25.26}     \\
SPD         & {\underline{66.80}}    & \textbf{54.04} & {\underline{65.27}}    & {\underline{57.53}}         & {\underline{24.5}5}        \\ \bottomrule
\end{tabular}}
\end{table}
\section{Quantifying Shifts using Maximum Mean Discrepancy}
\label{sec:suppl_mmd}

Mahalanobis distance is a dominant metric for measuring distribution shifts~\cite{miyai2024generalizedoutofdistributiondetectionvision}.
We further compare shifts using Maximum Mean Discrepancy (MMD)~\citep{gretton2008kernelmethodtwosampleproblem,dong2022neuralmeandiscrepancyefficient,ouyang2022maximummeandiscrepancygeneralization} with RBF kernel in Tab.~\ref{tab:mmd}. We observe similar trends as with Mahalanobis distance (i.e., higher scores indicate greater shifts), reinforcing the reliability of our shift scores. 

\begin{table}[htb]
\centering
\caption{\textbf{Maximum Mean Discrepancy metric with LoRA and Pretrained embeddings} on VQAv2 and its variants. We sample 1000 instances per dataset. Gamma=1.0, scale-up factor=$10^4$ }
\label{tab:mmd}
\resizebox{\linewidth}{!}{
\begin{tabular}{@{}c|c|c|c|c|c|c@{}}
\toprule
            & ID             & IVVQA          & VQA-REP        & VQA-CE          & TextVQA         & VizWiz          \\ \midrule
\( f_{\text{vanilla$\_$ft}}(q) \)   &    20.02          & 20.12          & 20.10         & 20.18          & 21.92           &          23.18       \\
\( f_{\text{vanilla$\_$ft}}(v) \)  & 20.17 & 20.20 & 20.13          &      20.28          &          21.98       &        23.07         \\
\( f_{\text{vanilla$\_$ft}}(v, q) \) & 20.10          & 20.12          & 20.12          &            20.20     &      22.44           &      23.02           \\
\( f_{\text{pt}}(q) \)       &  20.16         &       20.18         & 20.22      &     20.28            &      21.98           &    23.32             \\
\( f_{\text{pt}}(v) \)       &      20.15         &   20.18             &    20.20            &   20.25              &   22.47              &     23.75            \\
\( f_{\text{pt}}(v, q) \)         & 20.08         & 20.06          & 20.16 & 20.10  & 22.28  &      23.26           \\
             \bottomrule
\end{tabular}}
\end{table}

\section {Qualitative Analysis: Inspect via Sampling}

\begin{figure}[htb]
    \centering 
    \includegraphics[width=\linewidth]{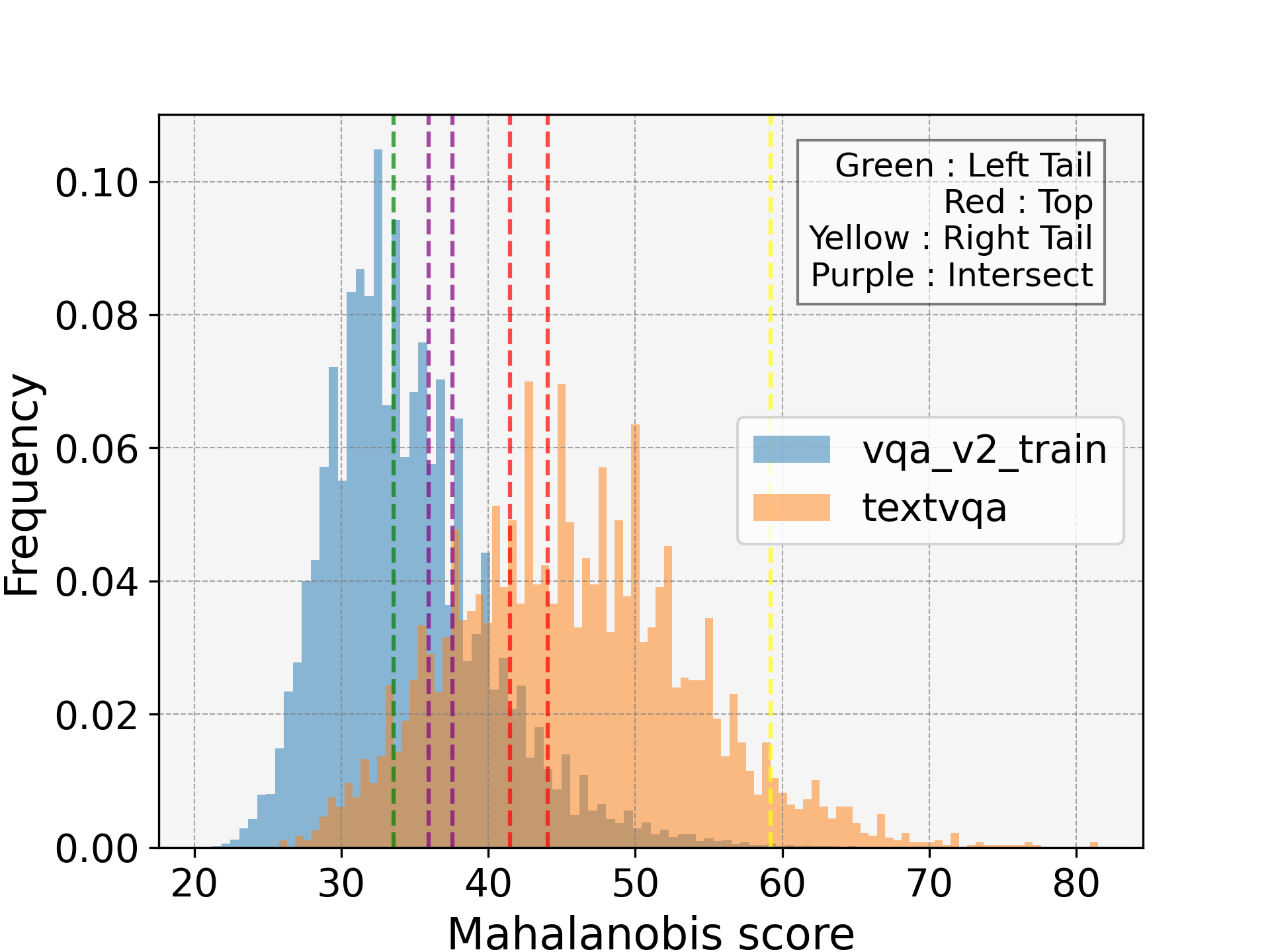} 
    \caption{Sampling region in histogram}
    \label{fig:sample_region}
\end{figure}

In order to investigate the types of ID and OOD samples under different modalities, we perform sampling on the various regions of the histogram to inspect how the model represents ID/OOD embeddings. This also serves as a verification of the reliability in quantifying shifts via feature-based representations. We select the vanilla fine-tuned (with LoRA) model as our model of choice and consider various regions in the distribution between both train and test splits. 

Under V, Q and V+Q, we sample 50 instances from 4 regions as shown in Fig.~\ref{fig:sample_region} including:

\begin{itemize}
    \item \textbf{Left tail (top \%5):} ID samples.
    \item \textbf{Top region:} top occurring samples in the test set. 
    \item \textbf{Intersect region:} similar samples between train \& test split.
    \item \textbf{Right tail (bottom \%5):} clear outlier concepts and exhibit uncommon \& hard instances (e.g. an image object that barely appears in left tail/peak region).
\end{itemize}

\subsection{Image Shift $f_\text{ft}(v)$}
We observe that the ID images involve commonly occuring objects shown in Tab.~\ref{tab:objects}. 
Under image shifts, we have the following observations and potential hypothesis. 

\begin{table}[htb]
    \centering
    \caption{Commonly Occurring Objects}
    \label{tab:objects}
    \begin{tabular}{c|c} %
        \toprule
        \textbf{Category} & \textbf{Examples} \\
        \midrule
        Animals & cats, dogs, giraffe \\
        Sports & baseball, skateboard, frisbee \\
        Objects & kite, fire hydrant, pizza \\
        Vehicles & bus, car, airplane \\
        \bottomrule
    \end{tabular}
\end{table}

\begin{figure*}[htb]
    \centering
    \begin{subfigure}[t]{0.45\linewidth} %
        \centering
        \includegraphics[width=0.9\linewidth]{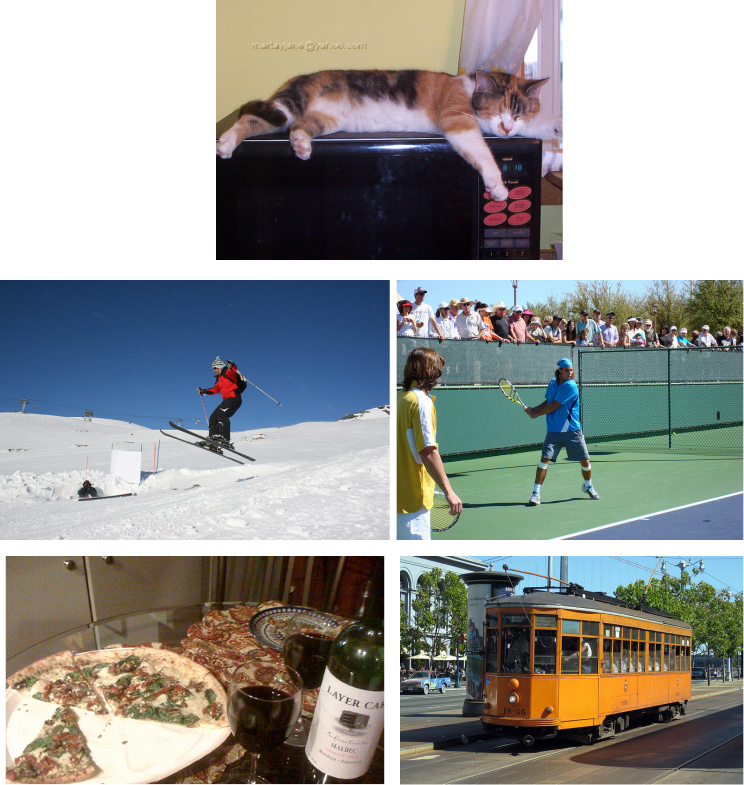} %
        \caption{ID images}
        \label{fig:id_images}
    \end{subfigure}
    \hfill %
    \begin{subfigure}[t]{0.45\linewidth} %
        \centering
        \includegraphics[width=0.9\linewidth]{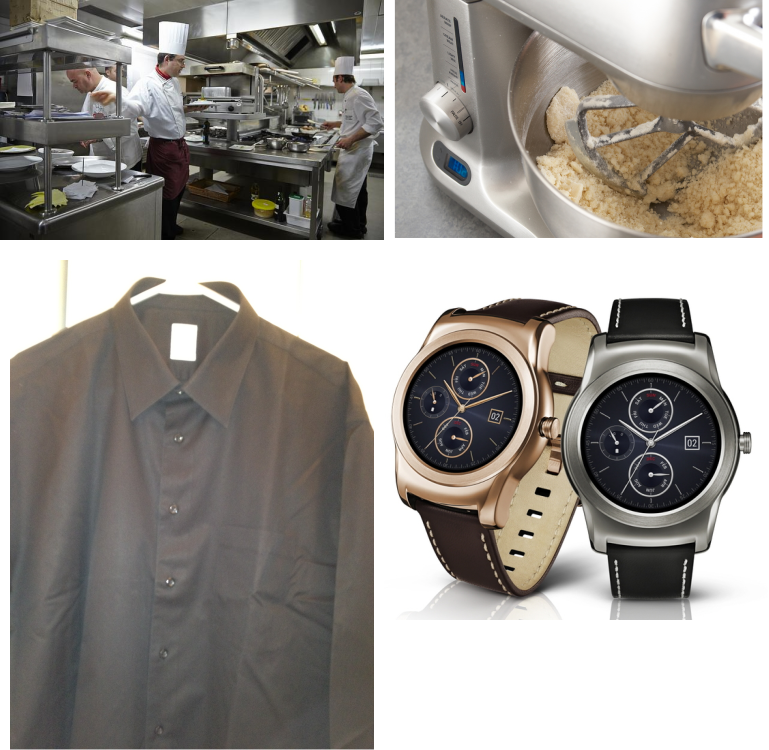} %
        \caption{OOD images}
        \label{fig:ood_images}
    \end{subfigure}
    \caption{Comparison set of ID and OOD images in terms of object categories and compositions (e.g., \# objects)}
    \label{fig:id_ood_comparison}
\end{figure*}

\begin{figure*}[htb]
    \centering
    \begin{subfigure}[t]{0.48\linewidth}
        \centering
        \includegraphics[width=0.9\linewidth,height=6cm,keepaspectratio]{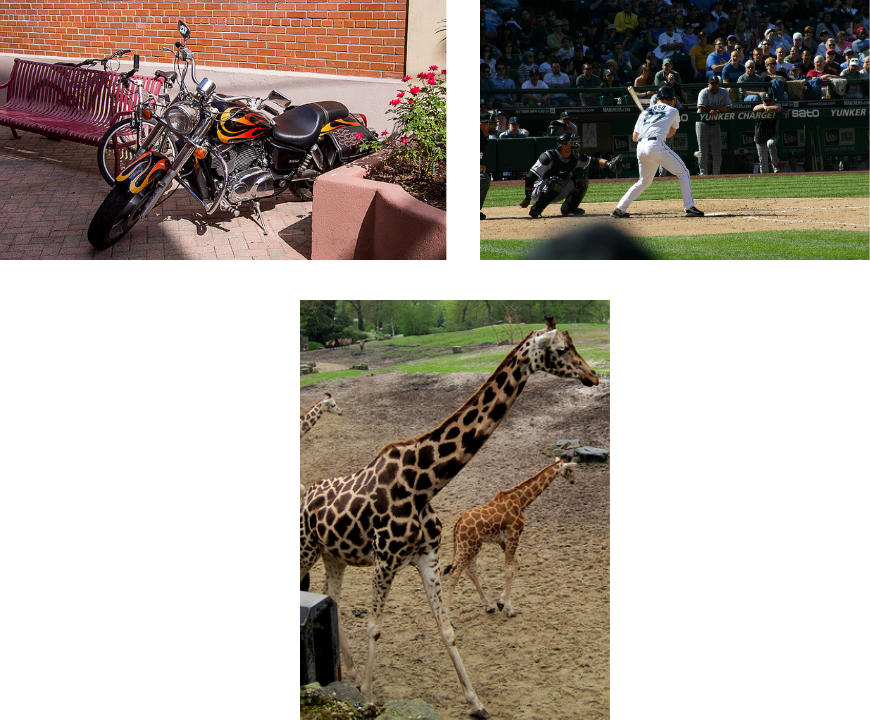}
        \caption{ID images}
        \label{fig:rare_id_images}
    \end{subfigure}
    \hfill %
    \begin{subfigure}[t]{0.48\linewidth}
        \centering
        \includegraphics[width=0.9\linewidth,height=6cm,keepaspectratio]{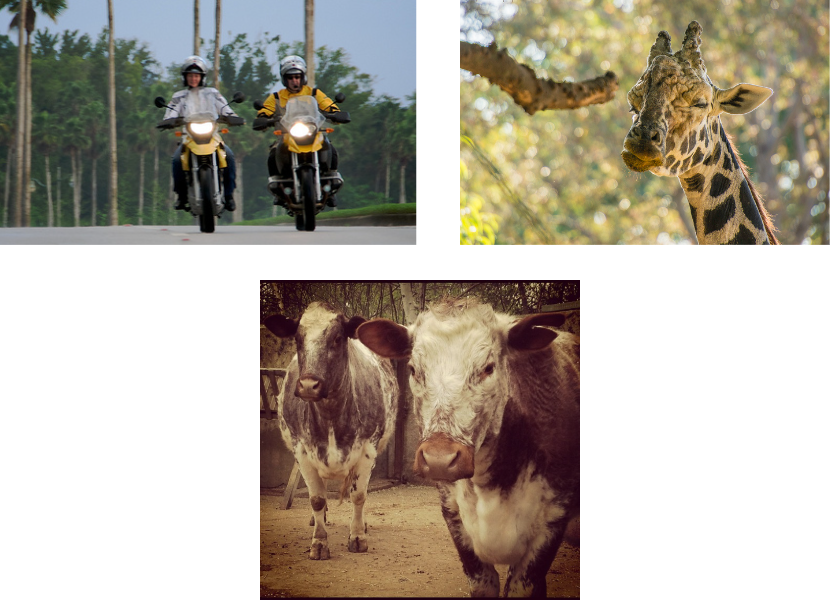}
        \caption{OOD images}
        \label{fig:rare_ood_images}
    \end{subfigure}
    \caption{Comparison of ID and OOD images but with similar objects.}
    \label{fig:rare_id_ood_comparison}
\end{figure*}

We observe that 1) There are distinct differences between left tail/peak and right tail (OOD) samples in terms of object categories and compositions (e.g., \# objects). Some examples are shown in Fig.~\ref{fig:id_ood_comparison}.
2) Intersecting regions have similar type concepts.
3) Tail samples seem to have less number of objects and contain more close up images.
4) There are still some instances where similar objects appear in significantly different regions, i.e., a commonly occuring object appears in the right region (OOD tail). These instances can be shown in Fig.~\ref{fig:rare_id_ood_comparison}.

We hypothesize that
1) There are significant visible shifts under the image domain with barely overlapping image objects between ID and OOD regions. However, some odd samples depicted in Fig.~\ref{fig:rare_id_ood_comparison} suggests some weakness of the fine-tuned model in robustly representing image embeddings. It fails to represent closer embeddings between images with similar objects.
2) Weight updates under joint inputs causes the two image embeddings of similar objects to steer in different directions. The image embeddings are indirectly conditioned on different questions and answers.

\subsection {Question Shift $f_\text{ft}(q)$}

Tab.~\ref{tab:quesex} details the ID and OOD question examples. The ID questions are much more straightforward where it involves simple identification of colours, activity, yes/no questions etc. OOD questions tend to incorporate more outside knowledge, which explains the drastic question shift in OKVQA, and more complex visual grounding such as OCR (Optical Character Recognition) tasks. 

\subsection {Joint Shift $f_\text{ft}(v,q)$}
Under the joint shift, there is an added complexity to this since we must consider the possible combinations of shifts under mixed modalities. 

\begin{table}[htb]
    \centering
    \caption{Region Samples}
    \label{tab:region_sample}
    \begin{tabular}{c|l} %
        \toprule
        \textbf{Region} & \textbf{Examples} \\
        \midrule
        Left Tail & \multirow{3}{*}{ID Object + ID Question} \\
        Peak &  \\
        Intersect & \\ \midrule
        \multirow{3}{*}{Right Tail} & ID Object + OOD Question \\
        & OOD Object + ID Question\\
        & OOD Object + OOD Question\\
        \bottomrule
    \end{tabular}
\end{table}

Tab.~\ref{tab:region_sample} outlines the types of VQA samples we expect to see under different regions. 
Intuitively, we would expect samples with OOD Question + OOD Object to be found in the right tail region. Whilst finding OOD Question + ID Object may indicate that the specific dataset has more samples with prominent OOD questions with ID images or that the joint shifts are more heavily influenced by text modality, causing samples with OOD questions to have a significant push to the joint embedding towards the right tail region.

Similarly, for ID Question + OOD Object, if those samples are found in the right tail, then it may indicate dataset having more OOD samples with ID Question + OOD Object or the visual modality has a greater influenced in steering the joint embedding.

\begin{table*}[htb]
    \centering
    \caption{Common questions with examples for ID and OOD cases.}
    \label{tab:quesex}
    \resizebox{\linewidth}{!}{ %
    \begin{tabular}{c|c|c}
        \toprule
        \textbf{Category} & \textbf{ID Examples} & \textbf{OOD Examples} \\
        \midrule
        Color & What color is the cat? & What color is on the inside of the speaker? \\
        \midrule
        Activity & What is this man doing? &  What is the person washing?\\
        \midrule
        Counting & How many \{ID objects\}? & How many are chocolate-covered donuts? \\ 
        \midrule
        \multirow{3}{*}{Outside knowledge} & \multirow{3}{*}{-} & What type of feed does this breed of horse need? \\
                          & & What is the name of the knot used on this tie? \\
                          & & If you add the two visible numbers, on the jerseys, what is the total sum? \\ 
        \midrule
        Text-in-image & What is the number in lights on the bus? & What is the name the package delivery company?\\
        \midrule
        \multirow{2}{*}{Brand/Species} 
        &  What brand of bike is this? & What brand of watch is shown? \\
        &  What species of animal are we looking at?  & What species of fish is being served?
        \\ 
        \bottomrule
    \end{tabular}}
\end{table*}

\begin{figure*}[htb]
    \centering
    \begin{subfigure}[t]{0.48\linewidth}
        \centering
        \includegraphics[width=\linewidth,height=6cm,keepaspectratio]{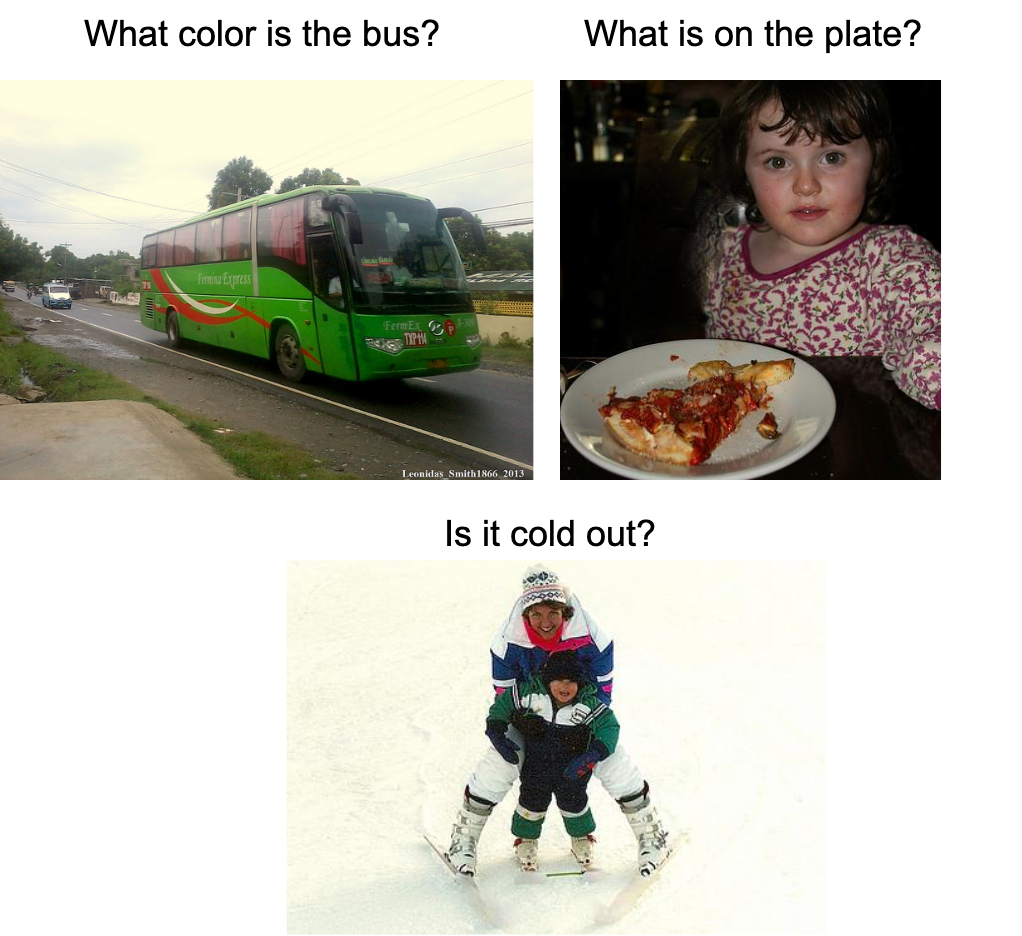}
        \caption{ID samples}
        \label{fig:id_joint}
    \end{subfigure}
    \hfill %
    \begin{subfigure}[t]{0.48\linewidth}
        \centering
        \includegraphics[width=\linewidth,height=6cm,keepaspectratio]{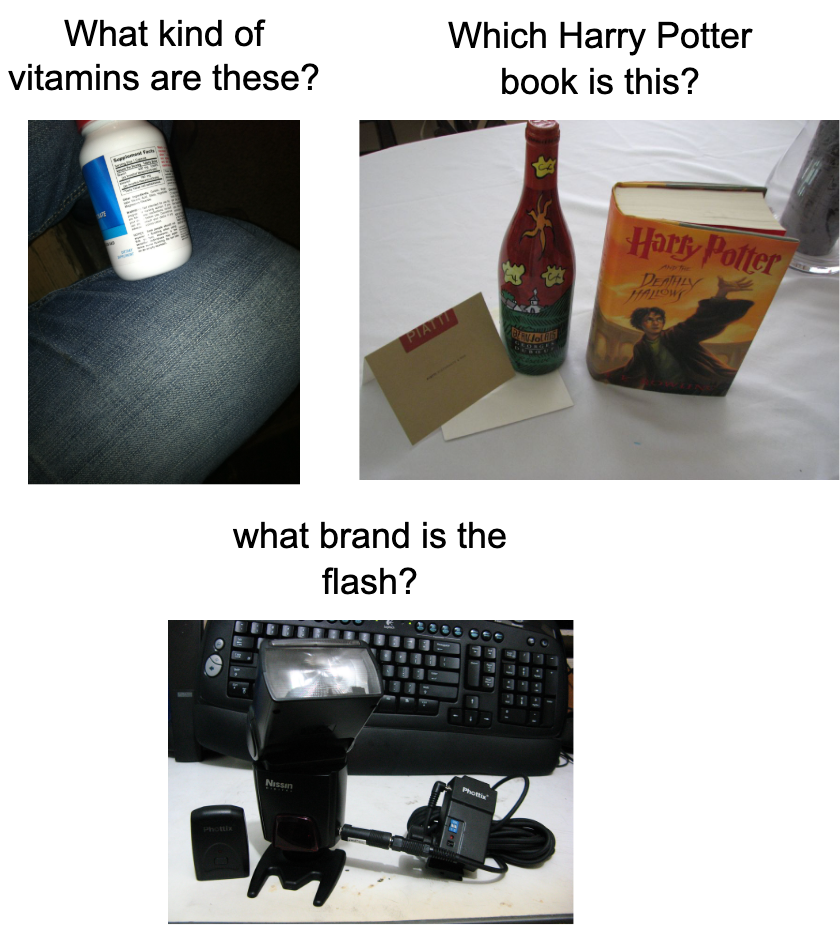}
        \caption{OOD samples}
        \label{fig:ood_joint}
    \end{subfigure}
    \caption{Comparison set of ID and OOD joint samples.}
    \label{fig:joint_samples}
\end{figure*}

ID and OOD joint samples can be viewed in Fig.~\ref{fig:joint_samples}.
Most of the ID samples have objective questions and images with easier to view objects. On the other hand, the right tail samples seem to involve harder questions that are more subjective, require outside knowledge, and more reasoning including VQA that involves reading text from an image. 
Interestingly, most of the right tail samples attribute to OOD questions regardless of whether its image pair is ID/OOD, i.e., the majority of inspected OOD samples are either OOD Question + OOD Object or OOD Question + ID Object. This suggests that amongst the 10 VQA datasets there is a much higher question shift than there is to images as well as a higher sensitivity to question shifts, since farther OODs are OOD Question + ID Object $>$ ID Question + OOD Object.

We also measure the composition of right tail samples under the joint modality by filtering samples that fall under each respective category and using an OOD threshold cutoff for each modality. 
The percentage of samples that make up the right tail region are shown in 
Tab.~\ref{tab:ood_port}. This aligns with our qualitative analysis that OOD questions make up a significant portion of Joint OOD regardless of whether they are ID/OOD images.

\begin{table}[h]
\centering
\caption{Percentage of each OOD type in Joint OOD. We use a threshold cutoff of 45 for visual OOD, 50 for question OOD, and 60 for joint OOD.}
\label{tab:ood_port}
\resizebox{\linewidth}{!}{
\begin{tabular}{@{}c|c|c|c@{}}
\toprule
 & OOD V + ID Q & ID V + OOD Q & OOD V + OOD Q \\ \midrule
\% composition in Joint OOD & 9.68 & 45.83 & 14.76 \\
\bottomrule
\end{tabular}}
\end{table}


\end{document}